\documentclass[10pt,twocolumn,letterpaper]{article}

\usepackage{cvpr}              
\usepackage{graphicx}
\usepackage{amssymb}

\usepackage{amsmath,amsfonts,bm,bbm}

\usepackage[usenames,dvipsnames]{xcolor}
\definecolor{snsgray}{RGB}{179, 179, 179}
\definecolor{snsorange}{RGB}{252, 141, 98}
\definecolor{snsblue}{RGB}{141, 160, 203}

\definecolor{coolgrey}{RGB}{157,157,157}
\definecolor{lightgrey}{RGB}{235,238,238}
\definecolor{lightteal}{RGB}{198,211,222}
\definecolor{cyan}{RGB}{136, 204, 238}
\definecolor{teal}{RGB}{68, 170, 153}
\definecolor{sand}{RGB}{221, 204, 119}
\definecolor{rose}{RGB}{204, 102, 119}
\definecolor{red}{RGB}{250, 94, 91}
\definecolor{orange}{RGB}{255, 200, 63}
\definecolor{yellow}{RGB}{254, 239, 109}

\definecolor{darkgreen}{rgb}{0.09, 0.45, 0.27}

\usepackage{xfrac}

\def\eqref#1{equation~\ref{#1}}

\def\1{\bm{1}}

\DeclareMathAlphabet{\mathsfit}{\encodingdefault}{\sfdefault}{m}{sl}
\SetMathAlphabet{\mathsfit}{bold}{\encodingdefault}{\sfdefault}{bx}{n}

\usepackage{booktabs}
\usepackage{marvosym}
\usepackage{xcolor}
\definecolor{cvprblue}{rgb}{0.21,0.49,0.74}
\usepackage[pagebackref,breaklinks,colorlinks,citecolor=cvprblue]{hyperref}
\usepackage{algorithm}
\usepackage{algorithmic}
\usepackage{booktabs}
\usepackage{graphicx}
\usepackage{subcaption}
\usepackage[T1]{fontenc}
\usepackage{multirow}
\usepackage[capitalize]{cleveref}
\usepackage{tcolorbox}
\tcbuselibrary{skins, breakable}
\usepackage{cuted}
\usepackage{multicol}
\usepackage{listings}
\usepackage[framemethod=TikZ]{mdframed}
\usepackage{xcolor, soul}
\usepackage{xcolor}
\usepackage{hyperref}
\hypersetup{
  colorlinks=true,
  linkcolor=blue,
  linktoc=none,
}

\definecolor{codegreen}{rgb}{0,0.6,0}
\definecolor{codegray}{rgb}{0.5,0.5,0.5}
\definecolor{codepink}{RGB}{252, 142, 172}
\definecolor{codepurple}{rgb}{0.58,0,0.82}

\definecolor{backcolour}{RGB}{245,245,245}
\lstdefinestyle{mystyle}{
    backgroundcolor=\color{backcolour},   
    commentstyle=\color{magenta},
    keywordstyle=\color{blue},
    numberstyle=\tiny\color{codegray},
    stringstyle=\color{codepurple},
    basicstyle=\fontfamily{\ttdefault}\footnotesize,
    breakatwhitespace=false,         
    breaklines=true,                 
    keepspaces=true,    
    frame=single,
    numbersep=5pt,                  
    showspaces=false,                
    showstringspaces=false,
    showtabs=false,                  
    tabsize=2,
    classoffset=1, 
    keywordstyle=\color{violet},
    classoffset=0,
}
\lstdefinelanguage{JavaScript}{
  keywords={typeof, new, true, false, catch, function, return, null, catch, switch, var, if, in, while, do, else, case, break},
  keywordstyle=\color{blue}\bfseries,
  ndkeywords={class, export, boolean, throw, implements, import, this},
  ndkeywordstyle=\color{darkgray}\bfseries,
  identifierstyle=\color{black},
  sensitive=false,
  comment=[l]{//},
  morecomment=[s]{/*}{*/},
  commentstyle=\color{purple}\ttfamily,
  stringstyle=\color{red}\ttfamily,
  morestring=[b]',
  morestring=[b]"
}
\tcbuselibrary{skins, breakable} 
\definecolor{redhl}{HTML}{FF0000}
\definecolor{greenhl}{HTML}{00FF00}
\definecolor{bluehl}{HTML}{00FFFF}
\definecolor{greyhl}{HTML}{C0C0C0}
\crefname{section}{Sec.}{Secs.}
\Crefname{section}{Section}{Sections}
\Crefname{table}{Table}{Tables}
\crefname{table}{Tab.}{Tabs.}

\usepackage{xcolor}
\usepackage{mdframed}
\definecolor{mylightblue}{RGB}{100,149,237} 

\tcbset{
  enhanced, 
  colback=white!200!black, 
  colframe=mylightblue, 
  colbacktitle=mylightblue, 
  title filled, 
  coltitle=white, 
  fonttitle=\bfseries, 
  arc=3mm, 
  outer arc=3mm, 
  boxrule=0.5mm, 
  toprule=0.5mm, 
  bottomrule=0.5mm, 
  titlerule=0.5mm, 
  drop fuzzy shadow, 
}

\definecolor{mycustomcolor}{HTML}{F5F5FB}

\makeatletter
\def\blfootnote{\xdef\@thefnmark{}\@footnotetext}
\makeatother

\begin{document}

\title{
\includegraphics[width=0.035\textwidth]{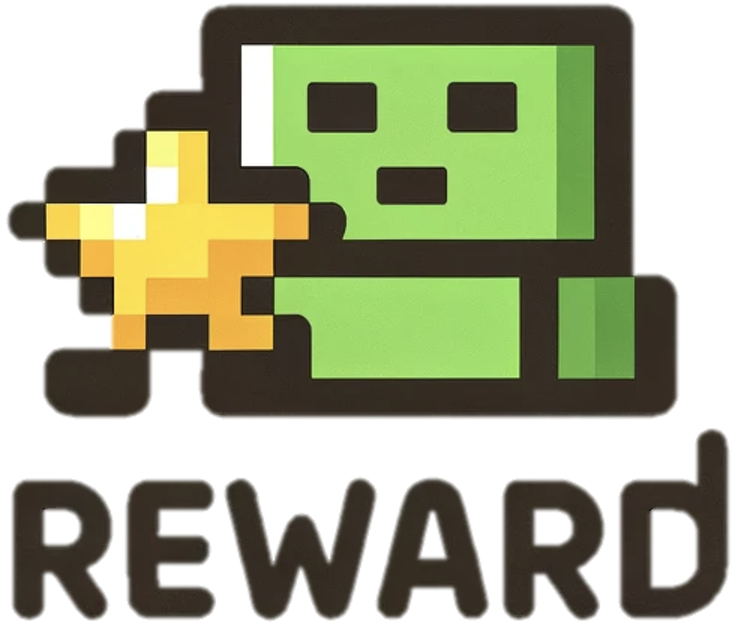}
Auto MC-Reward: Automated Dense Reward Design with \\ Large Language Models for Minecraft
\vspace{-1.2em}}

\author{
\textbf{Hao Li}\textsuperscript{1,2$*$}, \
\textbf{Xue Yang}\textsuperscript{2$*$}, \
\textbf{Zhaokai Wang}\textsuperscript{2,3$*$}, \
\textbf{Xizhou Zhu}\textsuperscript{4,5}, \\
\textbf{Jie Zhou}\textsuperscript{4}, \
\textbf{Yu Qiao}\textsuperscript{2}, \
\textbf{Xiaogang Wang}\textsuperscript{1,5}, \
\textbf{Hongsheng Li}\textsuperscript{1}, \
\textbf{Lewei Lu}\textsuperscript{5}, \
\textbf{Jifeng Dai}\textsuperscript{4,2\Letter}\\
\textsuperscript{1}CUHK-SenseTime Joint Laboratory, The Chinese University of Hong Kong \\
\textsuperscript{2}OpenGVLab, Shanghai AI Laboratory \\
\textsuperscript{3}Shanghai Jiao Tong University \quad
\textsuperscript{4}Tsinghua University \quad
\textsuperscript{5}SenseTime Research
 \\ 
\normalsize{\url{https://yangxue0827.github.io/auto_mc-reward.html}
\vspace{-4pt}
}
}

\twocolumn[{
    \maketitle
    \renewcommand\twocolumn[1][]{#1}
    \begin{center}
    \centering
    \vspace{-5mm}
    \includegraphics[width=0.93\textwidth]{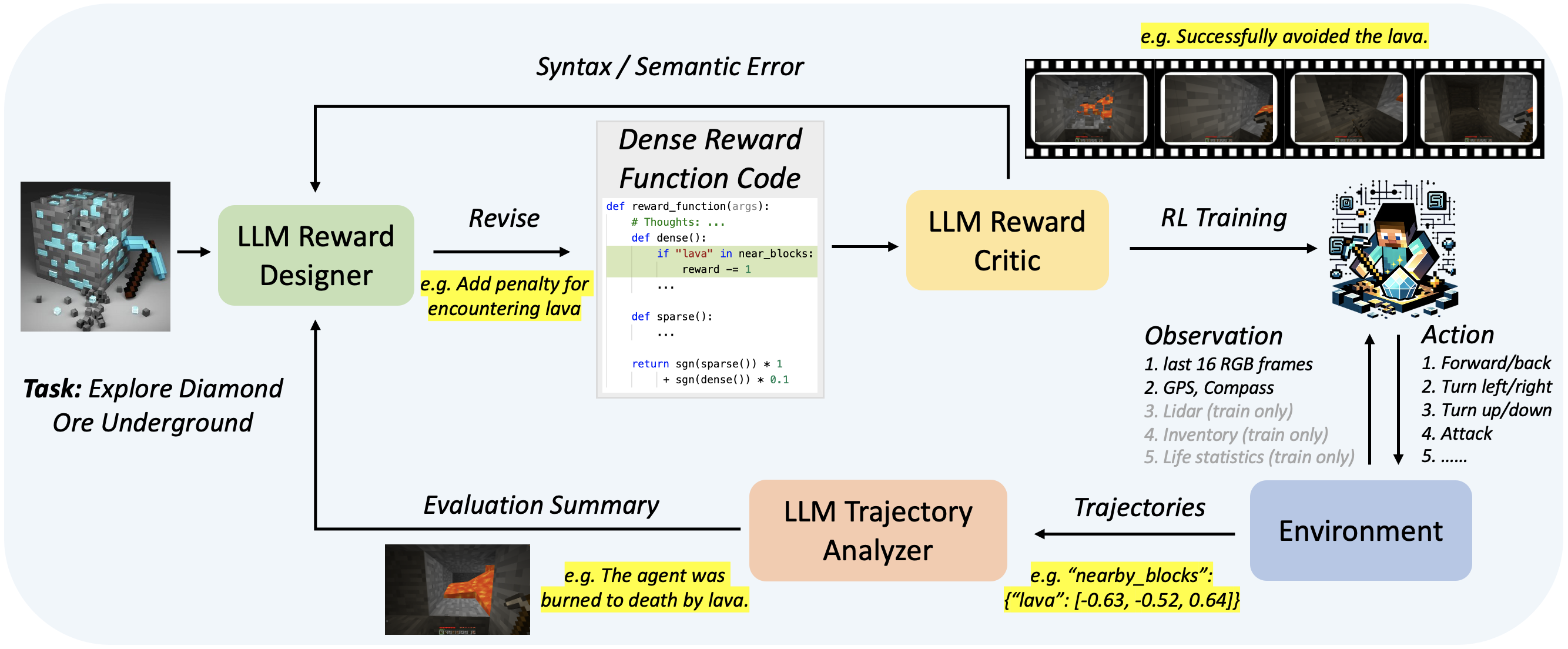}
    \vspace{-2mm}
    \captionof{figure}{Overview of our Auto MC-Reward. Auto MC-Reward consists of three key LLM-based components: Reward Designer, Reward Critic, and Trajectory Analyzer. A suitable dense reward function is iterated through the continuous interaction between the agent and the environment for reinforcement learning training of specific tasks, so that the model can better complete the task. An example of exploring diamond ore is shown in the figure: i) Trajectory Analyzer finds that the agent dies from lava in the failed trajectory, and then gives suggestion for punishment when encountering lava; ii) Reward Designer adopts the suggestion and updates the reward function; iii) The revised reward function passes the review of Reward Critic, and finally the agent avoids the lava by turning left.}
    \label{fig:pipeline}
    \end{center}
    
}]

\blfootnote{\noindent$^{*}$Equal contribution. This work was completed by Hao Li and Zhaokai Wang during their internship at Shanghai Artificial Intelligence Laboratory. ~\textsuperscript{\Letter}Corresponding author: Jifeng Dai <daijifeng@tsinghua.edu.cn>.}

\begin{abstract}

Many reinforcement learning environments (e.g., Minecraft) provide only sparse rewards that indicate task completion or failure with binary values. The challenge in exploration efficiency in such environments makes it difficult for reinforcement-learning-based agents to learn complex tasks. To address this, this paper introduces an advanced learning system, named Auto MC-Reward, that leverages Large Language Models (LLMs) to automatically design dense reward functions, thereby enhancing the learning efficiency. Auto MC-Reward consists of three important components: Reward Designer, Reward Critic, and Trajectory Analyzer. Given the environment information and task descriptions, the Reward Designer first design the reward function by coding an executable Python function with predefined observation inputs. Then, our Reward Critic will be responsible for verifying the code, checking whether the code is self-consistent and free of syntax and semantic errors. Further, the Trajectory Analyzer summarizes possible failure causes and provides refinement suggestions according to collected trajectories. In the next round, Reward Designer will further refine and iterate the dense reward function based on feedback.
Experiments demonstrate a significant improvement in the success rate and learning efficiency of our agents in complex tasks in Minecraft, such as obtaining diamond with the efficient ability to avoid lava, and efficiently explore trees and animals that are sparse in the plains biome.

\end{abstract}
    
\section{Introduction}
\label{sec:intro}

Minecraft, as the world's best-selling game, offers a range of tasks from exploration, survival to creating. It has become an important environment for researching efficient Reinforcement Learning (RL)~\cite{johnson2016malmo,guss2019minerl}. In particular, its extreme sparsity of rewards and huge complexity of the decision space pose significant challenges for RL. Currently, the most effective learning strategy involves pre-training through behavior cloning~\cite{baker2022video}, using learned behavioral priors to narrow the decision space. Nevertheless, it still requires billions of environmental interactions for effective learning due to the sparse reward nature of Minecraft.

On the other hand, previous researchers have proposed a variety of dense reward signals to enable efficient learning for specific sparse reward tasks~\cite{pathak2017curiosity,li2020random,oh2018self,andrychowicz2017hindsight,levy2017learning}. However, their applicability on the complex and long-horizon tasks in Minecraft remains an open question. To deeply reveal the challenges in Minecraft, we examine on several representative challenging tasks, \eg exploring underground for diamonds. We find that even after behavior cloning, most of these methods still fail to make significant progress on these tasks, further highlighting the difficulty of Minecraft and the limitations of existing dense reward methods.

It is noteworthy that for human players, Minecraft is a relatively simple casual game~\cite{duncan2011minecraft}. The advantage of human lies in their ability to summarize based on practice. For example, an accidental burning death from lava can teach human to avoid getting too close to it. Such summaries, based on life experience and practice, are key to human intelligence~\cite{tierney2009brain,seo2021impact}. Most existing RL methods overlook this ability. 
On the other hand, Large Language Models (LLMs) have recently demonstrated human-like common sense and reasoning capabilities~\cite{openai2023gpt,zhao2023survey,huang2022towards}. We find that leveraging LLMs can help RL agents simulate the practice summarization abilities of human. Based on the historical action trajectories and success-failure signals of the agents, LLMs can automatically design and refine corresponding auxiliary rewards, effectively overcoming the sparse reward challenge in Minecraft.

According to above analysis, we propose an automated method named Auto MC-Reward, to design and improve auxiliary reward functions according to task descriptions and historical action trajectories.
This method utilizes the task understanding and experience summarization abilities of LLMs, providing detailed and immediate rewards for learning guidance. Specifically
We first use LLMs to design task-related dense reward functions based on basic descriptions of the environment and tasks, named as Reward Designer. These reward functions are used to train agents after self-verification, \emph{i.e.} Reward Critic. To address potential biases or oversights in LLM's understanding, we also propose a LLM-based Trajectory Analyzer to analyze and summarize collected trajectories from the trained agent, and help Reward Designer to improve the reward functions. 

We verify the effectiveness of Auto MC-Reward on a series of representative benchmarks, including horizontal exploration for diamonds underground and approaching trees and animals in the plains biome. Experiments show that Auto MC-Reward achieves significantly better results on these tasks compared to original sparse reward and existing dense reward methods, showing its advanced ability of empowering efficient learning on sparse reward tasks. By iteratively refining the design of rewards functions, Auto MC-Reward enables the agent to efficiently learn new behaviors that is beneficial to the corresponding tasks, \emph{e.g.} avoiding lava, which greatly improves the success rate. Moreover, Auto MC-Reward achieves a high diamond obtaining success rate (36.5\%) with only raw information, demonstrating its ability of solving long-horizon tasks.

\section{Related Work}
\label{sec:related}

\noindent\textbf{Minecraft Agents} are intelligent agents designed to accomplish various tasks while playing the game Minecraft. Most of previous works adopt reinforcement learning for agent training. Due to the extremely sparse rewards and complex decision space of Minecraft tasks, early attempts have tried hierarchical RL~\cite{tessler2017deep,mao2022seihai,lin2022juewu,skrynnik2021hierarchical}, curriculum learning~\cite{kanitscheider2021multi}, and imitation learning~\cite{amiranashvili2020scaling} to empower more efficient RL training. To narrow the decision space, recent work~\cite{baker2022video} build a foundation model by performing imitation on YouTube videos. DreamerV3~\cite{hafner2023mastering} instead learns a world model that explores the environment efficiently. As the LLMs demonstrate their general planning ability, a series of research~\cite{huang2022inner,wang2023describe,yuan2023plan4mc,zhu2023ghost,wang2023voyager} leverage LLMs as high-level planners that decompose long-term complex tasks as basic skills and implement the skills with RL agents or handcrafted scripts.

Auto MC-Reward aims to design dense rewards for Minecraft tasks automatically using LLMs, which is orthogonal to previous works on Minecraft agents that mainly focus on RL learning algorithms or high-level planning.

\vspace{0.5em}\noindent\textbf{Efficient Learning in Sparse Reward Tasks} is a long-standing challenge in RL due to the lack of effective learning signals~\cite{ladosz2022exploration}. A common solution is to handcraft dense reward functions that provide intermediate reward signals based on human expertise, which requires time-consuming trial-and-error for each environment and task. Another line of previous research focus on proposing general-purpose dense auxiliary reward functions, such as curiosity-driven exploration~\cite{jaderberg2016reinforcement,pathak2017curiosity,burda2018large,li2020random}, self-imitation learning~\cite{oh2018self}, and goal-conditioned reinforcement learning~\cite{andrychowicz2017hindsight,levy2017learning,kulkarni2016hierarchical,vezhnevets2017feudal}. Despite the success on certain specific tasks, the applicability of these methods in the complex environment of Minecraft remains uncertain. 
Recent works~\cite{fan2022minedojo,ma2022vip,ma2023liv,du2023vision,du2023guiding} also propose to use pre-trained models to assign reward to intermediate states of completing tasks. However, these approaches produce reward values in a black-box manner, which cannot be interpreted and improved based on the experience of the agents, and the generalizability of these models on new tasks is not guaranteed.

In contrast, Auto MC-Reward automatically produces explainable reward functions according to the task descriptions. Moreover, the reward functions can be improved to be more precise based on the experience of the agent.

\vspace{0.5em}\noindent\textbf{Automated Reward Function Design} aims to find an optimal reward function that drives efficient reinforcement learning for interested tasks. Previous works~\cite{chiang2019learning,faust2019evolving} employ evolutionary algorithm for searching optimal reward functions for specific tasks. Most of these attempts have a highly constrained search space that only adjusts parameters of task-specific handcrafted reward templates. Recently, a series of research~\cite{du2023guiding,kwon2023reward,lin2022inferring,carta2022eager} employs LLMs for integrating human preference into open-domain tasks without clear completion criteria by directly prompting LLMs with environment trajectories and natural language task descriptions. The reward values are generated on-the-fly by LLMs, which is black-box and has heavy computational cost due to the nature of LLMs. In contrast, Auto MC-Reward employs LLMs to generate white-box code-form reward functions.

Concurrent works~\cite{yu2023language,xie2023text2reward,ma2023eureka} also propose to use LLMs as a coder to generate reward functions for robotics control tasks. Specifically, L2R~\cite{yu2023language} needs to prepare reward function templates and cannot cope with unexpected situations in open worlds. Text2Reward~\cite{xie2023text2reward} and EUREKA~\cite{ma2023eureka} require complete environment code or description and rely on human feedback, which are not available in open worlds. Different from these methods, Auto MC-Reward considers more complex Minecraft environments that has diverse scenarios and high uncertainty, requiring more precise and thorough reward designing.

\section{Method}
\label{sec:method}

Auto MC-Reward consists of three components: Reward Designer, Reward Critic, and Trajectory Analyzer. Given the environment information and task descriptions, the Reward Designer proposes the reward function by coding an executable Python function with pre-defined observation inputs. The Reward Critic verifies if the proposed reward function is self-consistent and if it meets the format requirements. The designed reward function which passes the Reward Critic is used to train agents in the environment. To improve the designed reward function according to empirical experience, the Trajectory Analyzer is proposed to summarize possible failure causes and provide refinement suggestions on the reward function based on the inference trajectories of the trained agents. Then the Reward Designer modifies the reward function based on the feedback from Trajectory Analyzer. Figure~\ref{fig:pipeline} shows the overview of the Auto MC-Reward.

\subsection{Reward Designer}
\label{sec:designer}

We utilize a Reward Designer to generate the reward function code to provide intermediate instructive learning signals to the agent. It takes as input task descriptions, game information, and reward function requirements, generating reward functions in executable code form. When updating reward function, we also provide analysis of the agent's performance when interacting with the game environment. The input prompt is introduced in Section~\ref{sec:imple_detail}.

The generated reward function uses a pre-defined observation format as input. This includes the nearest distance of each block type within the visible range in the current and previous steps, changes in inventory between adjacent steps, health points, and the agent's location in each past step. These parameters can provide information on the agent's current and historical states, assisting the reward function in various situations.

\vspace{0.5em}\noindent\textbf{Multi-Step Memory}. Long-term tasks require the transfer of information across multiple steps. 
Thus, we introduce a multi-step memory mechanism. It is provided to Reward Designer as a empty dictionary at the beginning, and the reward function can save necessary data into the memory to be used in future steps.
In the actual reward function of the explore-tree task, we observed that the agent records the distance to a tree at each step, thereby encouraging getting closer with the tree than the previous step.

\vspace{0.5em}\noindent\textbf{Chain of Thought}. We require the LLM to first describe its design thoughts, such as considering potential failure reasons and the details of the reward function design. These thoughts are to be written as comments at the beginning of the code. This is a mechanism similar to Chain of Thought (CoT)~\cite{wei2022chain}, where the thought process precedes the coding implementation. In the specific code implementation, necessary comments will also be generated every few lines (\eg, ``Check if lava is in the field of view in the previous step''). This approach not only allows Reward Designer to refer to the text-form thoughts during reward function initialization, but also assists subsequent Reward Critic in assessing the code's rationality, and helps Reward Designer to understand the current reward function's purpose when updating the reward function.

\vspace{0.5em}\noindent\textbf{Scale Constraints}. We impose a specific scale constraint for the reward function, where the LLM generates two sub-functions: dense and sparse. \emph{Sparse} denotes rewards for achieving the final goal or heavy penalties (like death), while \emph{dense} represents dense intermediate signals during the task completion process. We preset their numerical values and only allow the LLM to determine their positivity or negativity, limiting \emph{sparse} to $\{1, 0, -1\}$, and \emph{dense} to $\{0.1, 0, -0.1\}$. They are then added together for the final reward. Therefore, the final reward values can be one value of $\{\pm1.1, \pm1.0, \pm 0.9, \pm0.1, 0 \}$. The final reward is calculated as $R = \operatorname{sgn}(sparse) * 1 + \operatorname{sgn}(dense) * 0.1$, where $\operatorname{sgn}$ denotes the sign function.
This is to keep the reward within a reasonable range, allowing the LLM to focus on various scenarios that need to be considered in the reward function, rather than trivial tasks like adjusting the reward value.

\begin{figure}[!tb]
    \centering
    \includegraphics[width=0.43\textwidth]{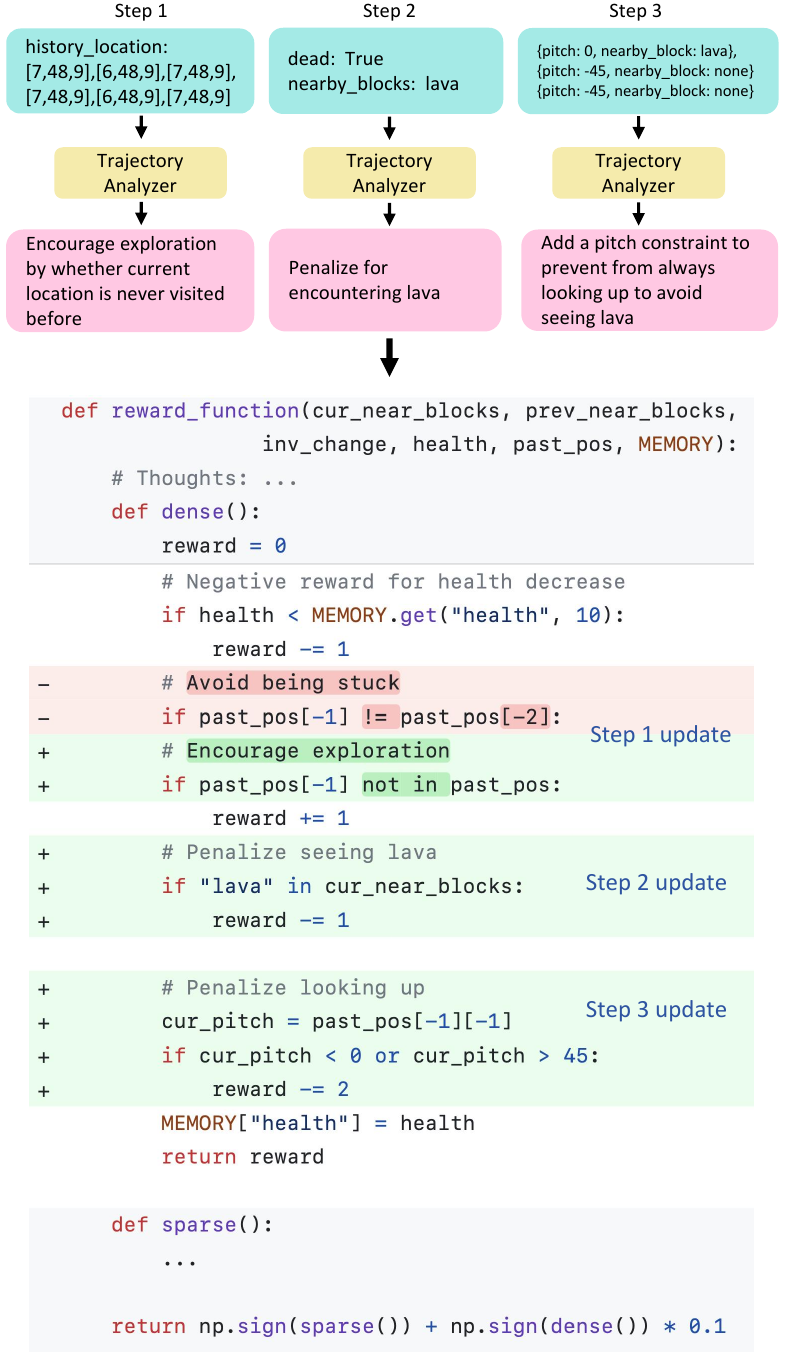}
    \vspace{-5pt}
    \caption{Example of updating the reward function. Trajectory Analyzer provides analysis for three scenarios at different steps, and then Reward Designer update the reward function based on the suggestions. We only display part of the trajectory data for brevity. \textbf{Step 1}: rewrite the code of encouraging exploration to avoid going back and forth. \textbf{Step 2}: add lava penalty to avoid falling into lava. \textbf{Step 3}: add pitch constraint to avoid constantly looking up to avoid lava.}
    \label{fig:code}
    \vspace{-10pt}
\end{figure}

\subsection{Reward Critic}
In practice, it is difficult for LLM to generate a relatively complete reward function in the beginning. There may be errors in understanding parameter formats and data types (\textbf{syntax errors}), failure to consider game-specific information, or misunderstanding of tasks (\textbf{semantic errors}), etc. 

In order to eliminate above errors that are not easy to find, we design a LLM based Reward Critic to automatically review the designed reward function. In addition to checking for syntax errors, Reward Critic is also asked to check the quality of the reward function to further eliminate semantic errors. Specifically, we require Reward Critic to check whether the current code implementation matches its thoughts, whether it meets the reward function design requirements, and whether it takes game information into account. If the review fails, the Critic will provide a critique, and the Reward Designer will then modify the reward function based on the criterion and submit it for review again. The above process is repeated up to 3 times.

If an error occurs during the execution of the reward function in the process of interacting with the environment, the Python traceback of the error message will be fed back to Reward Designer for modification. These errors may include misunderstandings of input parameters, list index out of range, uninitialized keys in dictionaries, and other such issues. Some runtime errors only appear during the actual execution of the code.

\subsection{Trajectory Analyzer}\label{sec:trajectory_analyzer}
LLMs have the ability to understand environmental information and task instructions through in-context prompts to generate dense rewards. However, this zero-shot approach completely relies on LLM’s understanding of the task and imagination of the problems it may face, and it is difficult to ensure the effectiveness of the designed reward. Take the the yellow highlighted part in Figure~\ref{fig:pipeline} as an example, in the initially designed reward function, Reward Designer does not consider the situation where the agent would encounter lava and be burned to death. Thus, in order to introduce empirical improvements to the designed dense reward, we propose to use LLMs, named as Trajectory Analyzer, to summarize the historical information of the interaction between the trained agent and the environment and use it to guide the revision of the reward function. The division of labor of Reward Designer and Trajectory Analyzer allows for independent operations of data analysis and reward function updates. Trajectory Analyzer does not need to know the details of the reward function, and Reward Designer does not need to process complex trajectory data.

Specifically, the current trained model is used to interact with the environment and obtain $K$ trajectories.
Then, we truncate these trajectories and use a LLM to summarize the observations of the last consecutive $L$ frames of each failed trajectory to automatically infer its possible failure reasons. Based on the analysis of the reasons for the failure, the LLM Trajectory Analyzer is asked to propose key points that Reward Designer needs to consider in the next round of reward function revision. For instance, failure scenarios where punishment is not considered, misalignment of dense reward and sparse reward causes the agent's behavior to deviate from the final goal, etc.

Figure~\ref{fig:code} shows an example of multiple rounds of improving the reward function during the search for diamonds. In the first step, through analysis of the trajectory, Trajectory Analyzer finds that the agent would opportunistically find a shortcut to increase the reward, that is, move back and forth to deceive the reward function into thinking that the agent is moving actively. Therefore, the Reward Designer modifies the code snippet that encourages the agent to move, \emph{i.e.} encourage the agent to appear in unvisited locations as much as possible. Although the initially designed reward function has taken into account the penalty for the loss of the agent's health, the agent still cannot effectively learn to avoid lava.
When modifying the reward function in the second round, Trajectory Analyzer discovers through the failed trajectory that the agent may die from lava, so it is suggested that Reward Designer increase the penalty for encountering lava, as shown in the step 2 update in Figure \ref{fig:code}.
According to the interactive experience, Reward Designer explicitly punishes the continuous appearance of lava in the field of view. However, the excessive punishment of lava caused the agent to choose to turn its perspective upward or downward to avoid the appearance of lava in the visible range, making it impossible for the agent to continue effective exploration, which deviates from the ultimate goal. To this end, Reward Designer further constrain the agent's perspective in step 3, so that the lava disappeared from the agent's perspective by turning left/right while continuing to search for diamonds, which is the desired strategy. Figure~\ref{fig:lava_cow}(a) shows the successful trajectory of avoiding lava: 
The agent sees the lava after breaking the stone ahead using iron pickaxe, and then turn left to avoid the lava through the mining tunnel.

\begin{algorithm}[!tb]
\footnotesize
    \caption{\small{Auto MC-Reward Training Pseudo Code}}
    \label{alg}
    \begin{algorithmic}
        \REQUIRE Task ($T$), Inital Agent ($A_0$), Environment (Env), Max  number of Critic reviews ($N_\text{Critic}$)
        \ENSURE Final Agent ($A_N$), Final Reward ($R_N$)
        \STATE Summary = None
        \STATE Critique = None
        \STATE $R_0$ = None
        \FOR{$i = 1, \dots, N$}
            \STATE R$_i$ = RewardDesigner(Summary, Critique, $T$, $R_{i-1}$)
            \FOR{$j = 1, \dots, N_\text{Critic}$}
                \STATE Critique, Done = RewardCritic($R_i$)
                \IF{Done}
                    \STATE break
                \ELSE
                    \STATE $R_i$ = RewardDesigner(Summary, Critique, $T$, $R_{i}$)
                \ENDIF
            \ENDFOR
            \STATE $A_i$ = TrainAgent($A_0$, $R_{i}$, Env, $T$)
            \STATE Traj$_i$, Stat$_i$ = Eval(Env, $A_i$)
            \STATE Summary = TrajectoryAnalyzer(Traj$_i$, Stat$_i$)
            \STATE Critique = None
        \ENDFOR
    \end{algorithmic}
    \label{alg:pseudo}
\end{algorithm}

\section{Experiment}
\label{sec:exp}

\subsection{Environment Setup}
We mainly use the harvest mode in the MineDojo~\cite{fan2022minedojo} environment to verify the model's ability to play Minecraft. 
The training pseudo code of Auto MC-Reward is shown in Algorithm~\ref{alg:pseudo}. 
We set up the following challenging tasks for model performance comparison and ablation study:

\begin{itemize}
    \item \textbf{Exploring diamond ore \includegraphics[width=0.02\textwidth]{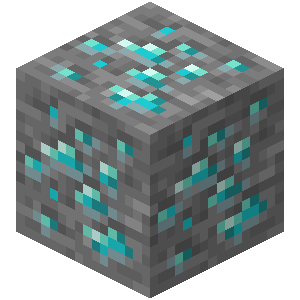} on the 11-th floor underground}: Initially, the agent~\includegraphics[width=0.02\textwidth]{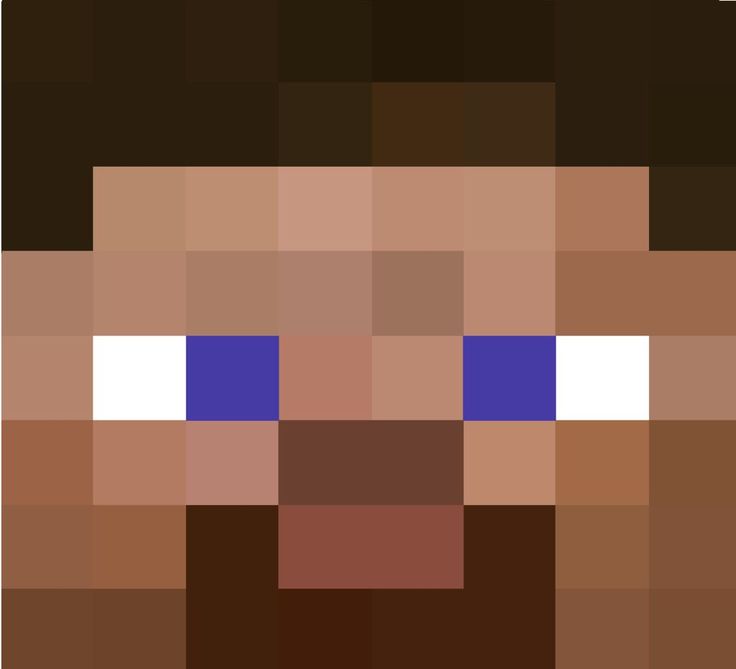}  is equipped with an iron pickaxe on the 11-th floor underground. When the diamond ore is within the visible range and the distance is less than 2 distance units, the task is deemed completed. The difficulty of the task lies in the fact that diamonds are very rare, lava frequently appears during exploration, leading to death, and the maximum number of steps is limited to 60,000. When steps exceed the limit, the trajectory is considered failed. Long-term exploration can demonstrate the advantages of dense rewards.
    \item \textbf{Approaching tree \includegraphics[width=0.02\textwidth]{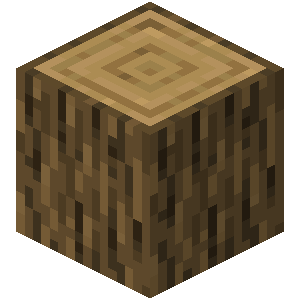} in plains biome \includegraphics[width=0.018\textwidth]{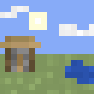}}: The task is considered successful if the tree is within the agent's~\includegraphics[width=0.02\textwidth]{figs/icon/steve2.jpg} visible range and the distance is less than 1 distance unit. The difficulty of the task lies in the fact that the trees are very sparse on plains, which is extremely detrimental to sparse reward functions. The maximum number of steps is limited to 2,000 steps.
    \item \textbf{Approaching specific animal (\emph{e.g.} cow \includegraphics[width=0.02\textwidth]{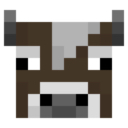}, sheep \includegraphics[width=0.02\textwidth]{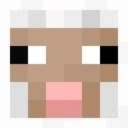}) in plains biome \includegraphics[width=0.018\textwidth]{figs/icon/plains.png}}: The task is considered successful if the animal is within the agent's~\includegraphics[width=0.02\textwidth]{figs/icon/steve2.jpg} visible range and the distance is less than 2 distance unit. The difficulty of the task is that the animals are constantly moving. The maximum number of steps is limited to 2,000 steps.
    \item \textbf{Obtaining diamond \includegraphics[width=0.023\textwidth]{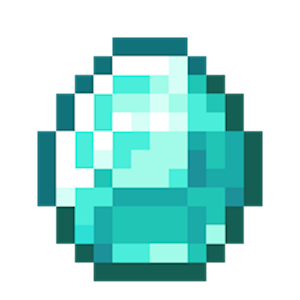}}: The agent~\includegraphics[width=0.02\textwidth]{figs/icon/steve2.jpg} needs to complete the whole process of mining diamonds, including key behaviors such as finding and obtaining materials on the surface, crafting, digging down, going back to the ground, and mining stone/iron ore/diamond ore. The tech tree is shown in Figure~\ref{fig:mine_diamond}.
\end{itemize}

\subsection{Implementation Details}
\label{sec:imple_detail}
\noindent\textbf{LLM Prompt.} The components of the input prompts for Trajectory Analyzer include task description, game information, statistical metrics, and information on failed trajectories. Components of the input prompts for Reward Designer and Reward Critic includes task description, game information, input parameters, and reward function requirements and format. We use GPT-4~\cite{openai2023gpt} for all the LLM components, and set temperature to 0.3. Since the LLMs are only used once for each whole agent training instead of each action, their computation overhead is negligible.

\begin{itemize}
    \item \textbf{Instruction}: Instructions on initializing, updating and handling execution error of reward function for Reward Designer, reviewing function for Reward Critic, and analyzing trajectory for Trajectory Analyzer.
    \item \textbf{Task description}: The objective, initial conditions, success criteria, and task flow. For example, for the explore diamond task, the objective is ``to find and approach a diamond, achieving a high success rate while avoiding death.'' The initial condition is ``agent at y level 11 with an iron pickaxe.'' The success criteria is ``being less than 1 meter from the nearest diamond block'', and the task flow is ``horizontally explore to find a diamond, face it, and approach it''. In the task description, we do not provide prior game strategy information (task challenges, DFS exploration strategies, or avoiding lava, etc.) to ensure the method's versatility.
    \item \textbf{Game information}: Game version, block names, field of view, action space, and units of measurement. Game information provides knowledge about the game's simulation environment, not game strategy.
    \item \textbf{Statistical metrics and information on failed trajectories}: success rates, and actions sequences, reward sequences, final inventory and nearby blocks of $K=10$ failed trajectories. If a trajectory exceeds $L=32$ steps, it is truncated to the last 32 steps.
     \item \textbf{Input parameters}: The nearest distance of each block type within the visible range in the current and previous steps, changes in inventory between adjacent steps, health points, and the agent's location in each past step. The memory is also provided as an input parameter for storing information to monitor changes across different steps. We provide explanation and examples of the parameters in the input prompt.
     \item \textbf{Reward function requirements and format}: We require the Designer to write a dense function and a sparse function, and consider only the sign of the two functions' return values, not the magnitude. The detail of the scale constraints is in Section~\ref{sec:designer}.
\end{itemize}

\begin{figure}[!tb]
    \centering
    \includegraphics[width=0.48\textwidth]{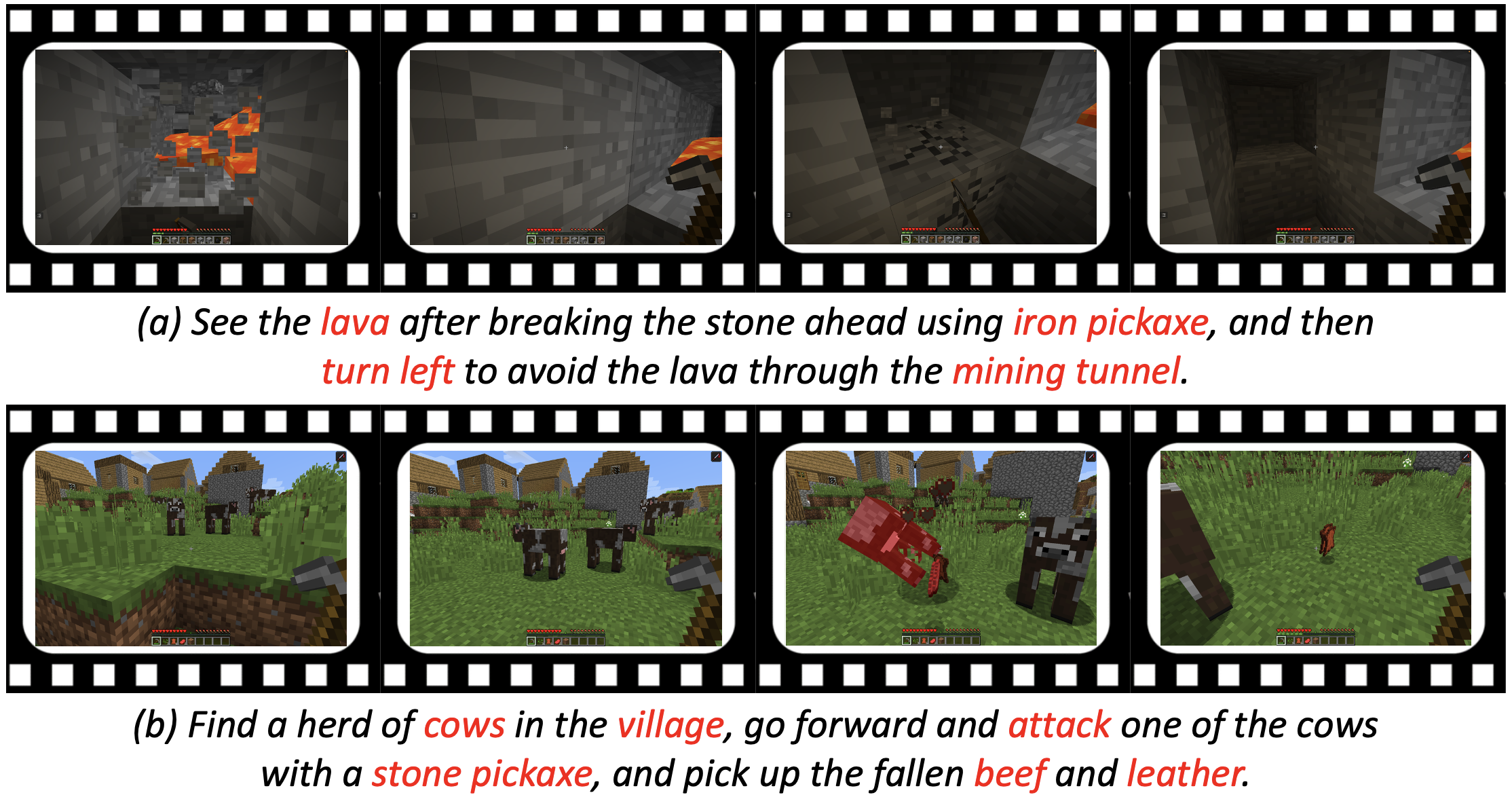}
    \caption{The trajectories of the new behaviors. (a) Avoid lava \includegraphics[width=0.015\textwidth]{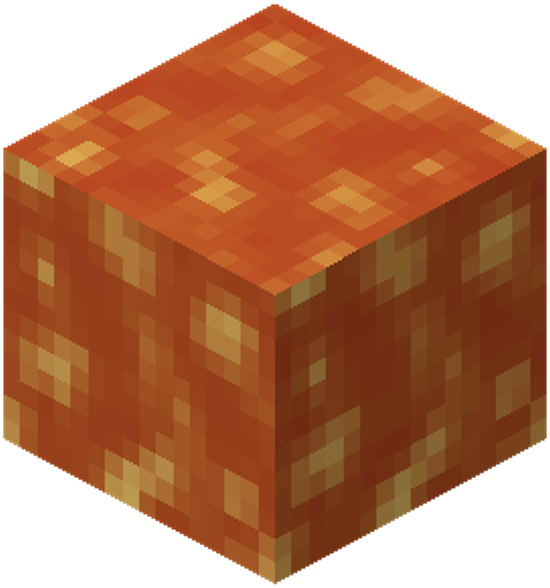} when exploring for diamond ore \includegraphics[width=0.015\textwidth]{figs/icon/diamond_ore.png}. (b) Attack cow \includegraphics[width=0.015\textwidth]{figs/icon/cow.jpg} in plains.}
    \label{fig:lava_cow}
    \vspace{-10pt}
\end{figure}

\begin{table*}[tb!]
    \centering
    \caption{Comparison with other reward methods on three Minecraft tasks. Max steps for exploring tree \includegraphics[width=0.018\textwidth]{figs/icon/oak_log.png} and cow \includegraphics[width=0.018\textwidth]{figs/icon/cow.jpg} are set to 2000. $^{\dag}$Sparse reward receives a low death rate because it is often stuck in the same place or move in a small area without encountering lava \includegraphics[width=0.015\textwidth]{figs/icon/lava.png}.}
    \vspace{-0.5em}
    \fontsize{8.0pt}{10.0pt}\selectfont
    \setlength{\tabcolsep}{1.5mm} 
    \setlength{\abovecaptionskip}{2mm} 
        \renewcommand\arraystretch{1.2}
    \resizebox{0.98\textwidth}{!}{
        \begin{tabular}{cc|cccc|cc|cc}
         \toprule
        \multirow{2}{*}{\textbf{Method}} &\multirow{2}{*}{\textbf{Reward}} & \multicolumn{4}{c|}{\textbf{Explore Diamond Ore~\includegraphics[width=0.018\textwidth]{figs/icon/diamond_ore.png} Underground}} & \multicolumn{2}{c|}{\textbf{Approach Tree~\includegraphics[width=0.018\textwidth]{figs/icon/oak_log.png} on Plains}} & \multicolumn{2}{c}{\textbf{Approach Cow~\includegraphics[width=0.018\textwidth]{figs/icon/cow.jpg} on Plains}} \\ 
        \cline{3-10}
        & & avg. dist. $\uparrow$ & death (\%) $\downarrow$ & lava escape (\%) $\uparrow$ & succ. (\%) $\uparrow$ & avg. step $\downarrow$ & succ. (\%) $\uparrow$ & avg. step $\downarrow$ & succ. (\%) $\uparrow$  \\ \hline
         \textbf{Naive Handcraft} & - & 85.7 & 74.3 & 1.5 & 18.6 & 1993 & 2.1 & 1956 & 10.8 \\
         \textbf{Imitation Learning} & - & 102.2 & 55.6 & 46.8 & 38.9 & 1988 & 2.5 & 1772 & 22.4 \\
         \textbf{RL} & Sparse & 16.8 & 1.5$^{\dag}$ & 0 & 0.5 & 1936 & 4.3 & 1854 & 12.6 \\
         \textbf{RL} & Dense (Curiosity) & 102.6 & 55.1 & 46.0 & 39.3& 1672 & 45.8& 1477 & 13.7 \\
         \textbf{RL} & Dense (Self-Imitation) & 104.0 & 54.8 & 47.2 & 39.7 & 1532 & 42.5 & 1280 & 23.5 \\
         \textbf{RL} & Dense (MineCLIP) & 105.9 & 54.0 & 47.8 & 40.5 & 1022 & 65.6 & 1206 & 44.9 \\
         \textbf{Ours} & Dense (LLM) & \textbf{142.8} & \textbf{45.2} & \textbf{70.0} & \textbf{45.2} & \textbf{972} & \textbf{73.4} & \textbf{1134} & \textbf{56.3} \\
         \bottomrule
\end{tabular}}
    \label{table:ablation_il_rl}
    \vspace{-8pt}
\end{table*}

\begin{table}[!tb]
\caption{Comparison with previous methods on success rates of obtaining diamond \includegraphics[width=0.018\textwidth]{figs/icon/diamond.png}. We list observations that are used in the inference phase. Auto MC-Reward achieves a remarkable success rate without exploiting unfair information (\emph{i.e.} Lidar and Voxel) during inference.}
\vspace{-0.5em}

\centering
\resizebox{0.45\textwidth}{!}{

\begin{tabular}{@{}cll|c@{}}
\toprule
\textbf{Method} & \textbf{Controller} & \textbf{Observation} & \multicolumn{1}{c}{\begin{tabular}[c]{@{}c@{}}\textbf{Diamond}\\ \textbf{Succ. (\%)}\end{tabular}} \\ 
\midrule
Human~\cite{baker2022video} & - & - & 50.0 \\ \midrule
DreamerV3~\cite{hafner2023mastering} & RL & RGB, Status & 0.01 \\
DEPS~\cite{wang2023describe} & IL & RGB, Status, Voxel & 0.6 \\
VPT~\cite{baker2022video} & IL + RL & RGB & 20.0 \\
GITM~\cite{zhu2023ghost} & Handcraft & Lidar, Voxel, Status & 55.0 \\
Ours & IL (GITM-guided) & RGB, GPS & 28.8 \\
Ours & IL + RL & RGB, GPS & 36.5 \\
\bottomrule
\end{tabular} }
\label{tab:comparison}
\end{table}

\begin{table}[t]
    \centering
    \caption{Ablations on Reward Critic and Trajectory Analyzer for explore diamond ore \includegraphics[width=0.015\textwidth]{figs/icon/diamond_ore.png} task. The first row corresponds to using the sparse reward from the original environment. $^{\dag}$Sparse reward receives a low death rate because it is often stuck in the same place or move in a small area without encountering lava \includegraphics[width=0.015\textwidth]{figs/icon/lava.png}.}
    \vspace{-0.5em}
    \fontsize{8.5pt}{12.0pt}\selectfont
    \setlength{\tabcolsep}{1.8mm}
    \setlength{\abovecaptionskip}{2.3mm}
    \renewcommand\arraystretch{1}
    \resizebox{0.48\textwidth}{!}{
        \begin{tabular}{c c c | c c c c}
        \toprule
        \textbf{Designer} &
        \textbf{Critic} & \textbf{Analyzer} & \textbf{Avg. Dist. $\uparrow$} & \textbf{Death $\downarrow$} & \textbf{Lava Esc. $\uparrow$} & \textbf{Succ. $\uparrow$} \\ \hline
         & & & 16.8 & 1.5$^{\dag}$ & 0 & 0.5 \\
        $\checkmark$ & & & 75.8 & 58.2 & 30.4 & 35.1 \\
        $\checkmark$ & $\checkmark$ & & 95.2 & 49.3 & 40.7 & 40.5 \\
        $\checkmark$ & & $\checkmark$ & 130.6 & 47.8 & 64.8 & 43.1 \\
        $\checkmark$ & $\checkmark$ & $\checkmark$ & \textbf{142.8} & \textbf{45.2} & \textbf{70.0} & \textbf{45.2} \\
        \bottomrule
    \end{tabular}}
    \label{table:critic_analyzer}
    \vspace{-10pt}
\end{table}

\noindent\textbf{Imitation Learning Details.}\label{sec:il}
When large labeled datasets do not exist, the canonical strategy for training capable agents is RL, which is inefficient and expensive to sample for hard-exploration problems~\cite{berner2019dota,jaderberg2019human,baker2022video}, 
\emph{e.g.} mining diamond in Minecraft. Therefore, in order to more efficiently explore the effectiveness of the LLM-based reward function design mechanism proposed in this paper, we pre-trained some foundation models through imitation learning as done by VPT~\cite{baker2022video}. Specifically, we use GITM~\cite{zhu2023ghost} to continuously perform Diamond Mining task and record important observation data of each frame, such as RGB, action, inventory, GPS, compass, structured actions, etc. In the end, we collect about 11 million image data, totaling about 153 hours (the control frequency is 20 Hz) of game videos. Subsequently, we train these data through fully supervised learning by using Impala CNN~\cite{espeholt2018impala} and Transformer~\cite{vaswani2017attention} as backbone, and obtained several foundation models. The main differences between the foundation models are different biomes (forest and plains), temporal frames (16 and 128), and whether goal embedding is used. 
In subsequent experiments, these foundation models were used in two different purposes:
\begin{itemize}
    \item Give the RL model preliminary basic Minecraft gameplay capabilities, \emph{e.g.} forward/back, turn left/right, attack, etc. For some tasks that have not been learned (\emph{e.g.} approaching cows in Figure~\ref{fig:lava_cow}(b)) or not learned well (\emph{e.g.} avoid lava in Figure~\ref{fig:lava_cow}(a), explore tree on plains) in imitation learning, 
    RL algorithms can be studied more efficiently. 
    \item In the Diamond Mining task, the diamond collection success rate, lava escape rate, death rate, etc. between the RL model and the imitation learning model are compared to demonstrate the superiority of the proposed method.
\end{itemize}

\noindent\textbf{RL Training Details.}
We use proximal policy optimization (PPO) algorithm~\cite{schulman2017proximal} with generalized advantage estimation (GAE)~\cite{schulman2015high} to train our RL model. We use $\gamma=0.99$ and $\lambda=0.95$ for all of our experiments, and the total training frames is 256,000.
To prevent catastrophically forgetting or overly aggressive policy update during RL training, we follow VPT~\cite{baker2022video} to apply an auxiliary Kullback-Leibler (KL) divergence loss between the RL model and the frozen pre-trained policy. We also normalize the reward based on the trajectory returns to constrain the gradient scales of different tasks. See Appendix for details. 

\subsection{Main Results}

\noindent\textbf{Baselines.} We compare our Auto MC-Reward against the following methods:

\begin{itemize}
    \item\textbf{Naive Handcraft}: The agent keeps moving (and mining for diamond exploring task) in one direction with a small probability of turning left/right.
    \item\textbf{Imitation Learning}: Our foundation model pre-trained with GITM-generated data, as introduced in Section~\ref{sec:imple_detail}.
    \item\textbf{RL with Sparse Reward}: Use only the reward from the original environment, \emph{i.e.} only receives a reward when the success criteria is completed.
    \item\textbf{RL with Curiosity Dense Reward}~\cite{pathak2017curiosity}: Encourage the agent to discover and learn about parts of the environment that it has not encountered before.
    \item\textbf{RL with Self-Imitation Dense Reward}~\cite{oh2018self}: Encourage the agent to replicate its past actions that led to high rewards.
    \item\textbf{RL with MineCLIP~\cite{fan2022minedojo} Dense Reward}: Use MineCLIP to calculate the dense reward based on the similarity between RGB frames and task objectives.
\end{itemize}

\noindent\textbf{Results on Diamond Ore \includegraphics[width=0.02\textwidth]{figs/icon/diamond_ore.png} Exploring Task.}
For the plain imitation learning model, fitting the training data makes it lack the awareness of avoiding lava, so it often dies in lava during the search for diamonds, and only has 38.9\% success rate under the limit of 60,000 steps, as shown in Table~\ref{table:ablation_il_rl}. In contrast, our Auto MC-Reward makes the agent realize the importance of avoiding lava by continuously improving the dense reward function, and the final success rate has increased to 45.2\% with 70\% lava escape success rate. Figure~\ref{fig:lava_cow}(a) demonstrates good awareness of avoiding lava. Based on the same reinforcement learning algorithm, the disadvantages of sparse reward functions in long-horizon tasks are undoubtedly revealed. By watching the videos of collected trajectories, we find that using sparse functions often leads to irreversible behavior, such as being unable to break the surrounding ores to move due to maintaining a head-up posture. Although a low death rate of 1.5\% is achieved, the actual average moving distance is only 16.8, and the success rate is only 0.5\%. Due to the similar scenes underground, MineCLIP cannot give differentiated rewards, so its performance is close to the initial imitation learning model. Other baselines, like curiosity and self-imitation dense reward, also have mediocre performance and the success rate has not been significantly improved. 

\noindent\textbf{Results on Tree \includegraphics[width=0.02\textwidth]{figs/icon/oak_log.png} Approaching Task.}
Since trees are extremely sparse on the plain, the imitation learning model and the RL model with sparse reward cannot perform well, with only 2.5\% and 4.3\% success rates respectively, and their average action steps are close to the maximum limit.
MineCLIP dense reward receives a success rate of 65.6\% since it can provide positive reward when tree is visible. Curiosity and self-imitation methods also achieve better results than imitation learning.
For Auto MC-Reward, Reward Designer uses a strategy of giving positive rewards for getting closer and deducting rewards for going away, so that the agent learns to slowly approach the target, ultimately achieving 73.4\% success rate with only 972 average steps.

\begin{figure}[!tb]
    \centering
    \includegraphics[width=0.42\textwidth]{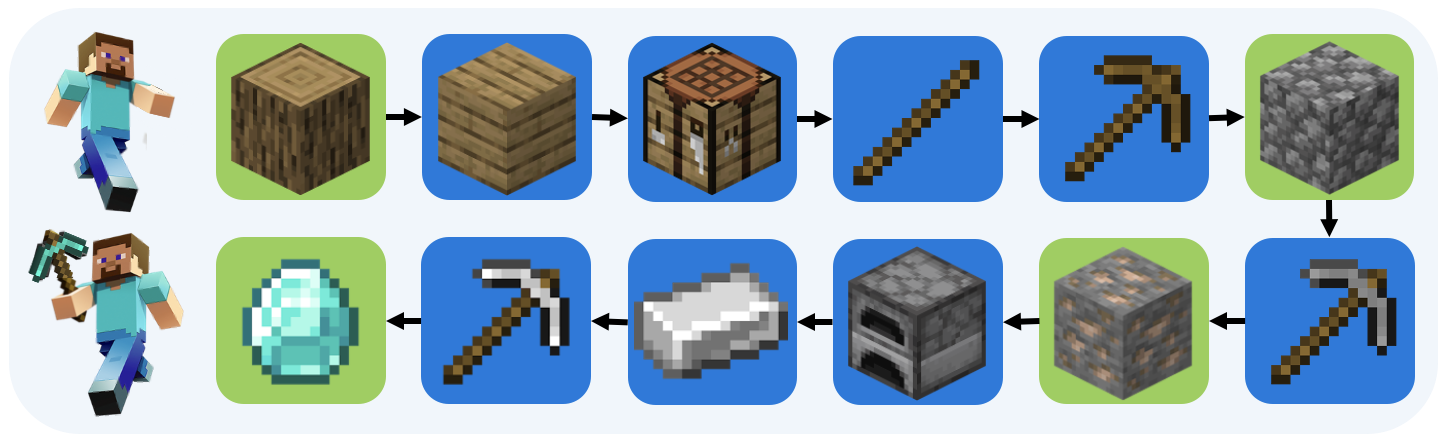}
    \vspace{-2mm}
    \caption{The tech tree of obtaining diamond. The green squares are tasks to be optimized with Auto MC-Reward, \emph{i.e.} obtaining log \includegraphics[width=0.015\textwidth]{figs/icon/oak_log.png}, cobblestone \includegraphics[width=0.015\textwidth]{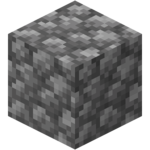}, iron ore \includegraphics[width=0.015\textwidth]{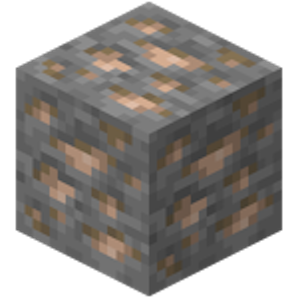} and diamond \includegraphics[width=0.018\textwidth]{figs/icon/diamond.png}.}
    \vspace{-15pt}
    \label{fig:mine_diamond}
    
\end{figure}

\noindent\textbf{Results on Cow \includegraphics[width=0.02\textwidth]{figs/icon/cow.jpg} Approaching Task.}
The task of exploring for cows does not appear in the training data of imitation learning, so the zero-shot ability on this task is not ideal, with about 22.4\% success rate and average steps close to the maximum limit. By checking the videos, we find most of the successful cases are due to good luck without intention to actively approach the target. The same experimental conclusion is also obtained in the experiment of sparse reward function.
Similar to the Tree Approaching Task, the superior dense reward function design mechanism makes our agent 43.7\% (56.3\% vs. 12.6\%) higher than sparse reward, as listed in Table~\ref{table:ablation_il_rl}. Another dense reward MineCLIP also shows strong performance in this task, but due to the need to calculate the similarity of images and texts at all times during training, the efficiency is unacceptable.

\begin{figure}[!tb]
    \centering
    \includegraphics[width=0.46\textwidth]{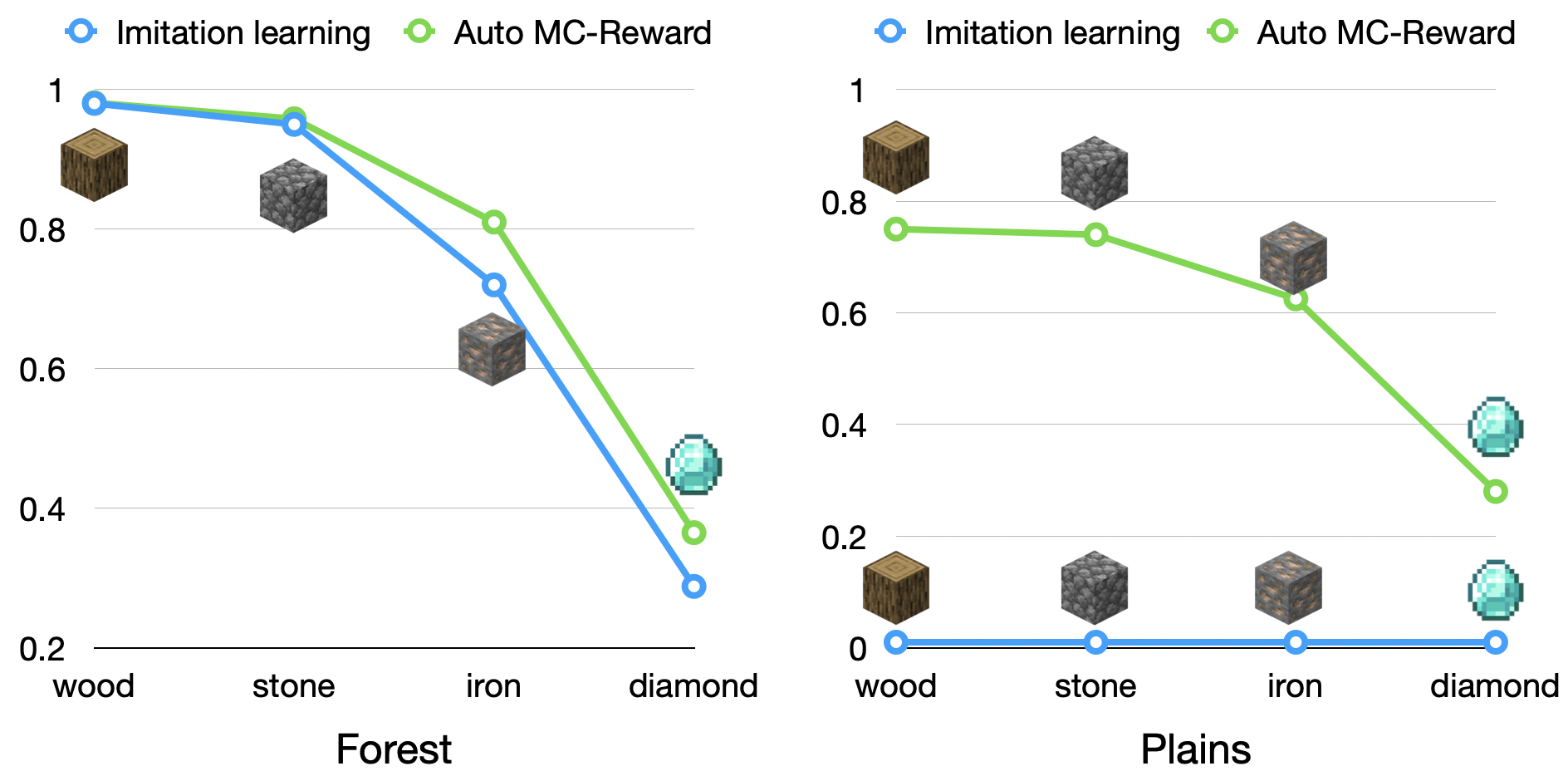}
    \caption{Cumulative success rates for 4 key items of obtaining diamond on forest \includegraphics[width=0.015\textwidth]{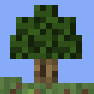} and plains \includegraphics[width=0.015\textwidth]{figs/icon/plains.png}. In terms of diamond, the performance comparison between imitation learning and Auto MC-Reward in two biomes are: 28.8\% vs. 36.5\%, and 0\% vs. 28.1\%.}
    \label{fig:linechart}
    \vspace{-10pt}
\end{figure}

\noindent\textbf{Results on Obtaining Diamond \includegraphics[width=0.023\textwidth]{figs/icon/diamond.png}.}
We verify the proposed method on a more difficult task, that is, the tech tree of collecting diamonds, as shown in Figure~\ref{fig:mine_diamond}. As mentioned before, our foundation imitation learning model already has a certain ability from birth to diamond mining. We use the proposed method to optimize several key tasks in the process to increase the success rate of final diamond acquisition. The green parts in Figure~\ref{fig:mine_diamond} are the tasks that need to be optimized, \emph{i.e.} obtaining log, cobblestone, iron ore and diamond. We conduct experiments in two biomes in Minecraft, and the cumulative success rate is shown in Figure~\ref{fig:linechart}. Specifically, the lower death rate allows our agent to have a higher success rate in mining iron ore and diamond, and ultimately achieves 36.5\% success rate on forest biome, which is 7.7\% higher than the imitation learning model. As for plains, the difficulty of obtaining log makes the imitation learning model unable to complete any tasks. Auto MC-Reward overcomes the difficulty of obtaining log, thus achieving a 28.1\% success rate in obtaining diamonds.
Table~\ref{tab:comparison} provides a rough comparison of several different methods on the task of mining diamonds. We achieve a high success rate without exploiting unfair information (\emph{i.e.} Lidar and Voxel) during the inference phase.

\subsection{Ablation studies}

\noindent\textbf{Effectiveness of Reward Designer.}
The first row of Table~\ref{table:critic_analyzer} is an RL experiment with a sparse reward function. As mentioned before, it cannot explore diamonds normally. After adding Reward Designer, it regained the ability to explore under a dense reward function.

\noindent\textbf{Effectiveness of Reward Critic.}
As listed in Table~\ref{table:critic_analyzer}, the success rate of exploring diamonds has increased from 35.1\% to 40.5\% by adding Reward Critic, because it can reduce the syntax and semantic errors in the code,
making the training process more effective and sufficient.
For example, the Trajectory Analyzer concludes that the agent died in lava and asks the Reward Designer to add relevant penalties. However, without being checked by Critic for semantic errors, it is possible that the added code snippet uses the word ``magma'' instead of the correct one ``lava''. This will result in insufficient learning of lava avoidance, which is reflected in a 2.1\% (43.1\% vs. 45.2\%) success rate difference.

\noindent\textbf{Effectiveness of Trajectory Analyzer.}
As observed in Table~\ref{table:critic_analyzer}, Trajectory Analyzer is the key to improve the success rate of completing tasks. It summarizes the reasons for failure to be fed into Reward Designer, allowing it to iteratively modify an appropriate dense reward function to guide the agent to overcome difficulties. In terms of Diamond Exploring Task, Trajectory Analyzer provides timely feedback on the potential risks of lava, which greatly improves the survival rate and moving distance, ultimately improving the success rate from 40.5\% to 45.2\%.

\vspace{-1mm}
\section{Conclusion}
\label{sec:conclusion}
\vspace{-2mm}

We proposed Auto MC-Reward, an automated dense reward design framework for addressing challenges caused by sparse reward and complex environment of Minecraft. It addresses the issue of sparse rewards by leveraging LLMs to automatically generate dense reward functions, enhancing learning efficiency. The system consists of three key components: Reward Designer, Reward Critic, and Trajectory Analyzer, which are used for the design, verification and analysis of the reward function respectively. Its capabilities are validated through experiments, demonstrating a remarkable improvement in complex tasks in Minecraft. Future work may deal with the limited trajectory length for analysis (last 32 frames) due to the context length of LLMs, which hinders the analysis of long-term failures (e.g., not exploring new areas, circling around lava). Auto MC-Reward humbly contributes to more effective learning in complex tasks through its automated dense reward function design.
We hope it can pave the way for further research in reinforcement learning and its real-world applications.

\vspace{-1mm}
\section*{Acknowledgement}
\vspace{-2mm}
This work is supported by the National Key R\&D Program of China (NO. 2022ZD0161300, NO. 2022ZD0160100), by the National Natural Science Foundation of China (62376134).

{
    \small
    \bibliographystyle{ieeenat_fullname}
    \bibliography{main}
}

\newpage

\appendix

\renewcommand{\thetable}{\Roman{table}} 
\renewcommand{\thefigure}{\Roman{figure}}
\renewcommand{\thealgorithm}{\Roman{algorithm}} 

\setcounter{figure}{0}
\setcounter{table}{0}
\setcounter{algorithm}{0}

\begin{onecolumn}

\section{Experiments on More Minecraft Tasks}
\subsection{Environment Setup}
This section conducts experiments on more Minecraft tasks to verify the effectiveness of the proposed reward function design mechanism. The experiments mainly includes the approaching task of three animals including pig \includegraphics[width=0.02\textwidth]{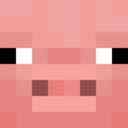}, sheep \includegraphics[width=0.02\textwidth]{figs/icon/sheep.jpg} and chicken \includegraphics[width=0.02\textwidth]{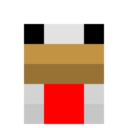}, as well as the attacking \includegraphics[width=0.02\textwidth]{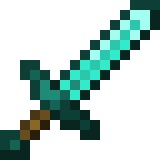} task of cow \includegraphics[width=0.02\textwidth]{figs/icon/cow.jpg}.
\begin{itemize}
    \item \textbf{Approaching animals in plains biome.} The task is considered successful if the animal is within the agent's \includegraphics[width=0.02\textwidth]{figs/icon/steve2.jpg} visible range and the distance is less than 2 distance unit. The difficulty of the task is that the animals are constantly moving and small (\emph{e.g.} chicken \includegraphics[width=0.02\textwidth]{figs/icon/chicken.png}). The maximum number of steps is limited to 2,000 steps.
    \item \textbf{Attacking cow in plains biome.} The task is considered successful if the agent \includegraphics[width=0.02\textwidth]{figs/icon/steve2.jpg} successfully kills \includegraphics[width=0.02\textwidth]{figs/icon/diamond_sword.png} a cow \includegraphics[width=0.02\textwidth]{figs/icon/cow.jpg} and obtains beef \includegraphics[width=0.02\textwidth]{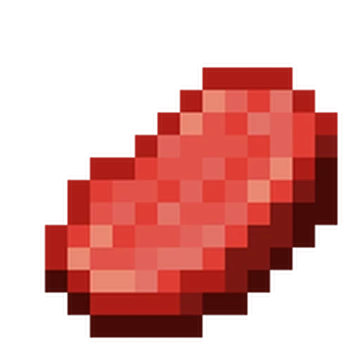} or leather \includegraphics[width=0.02\textwidth]{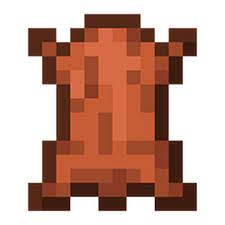}. The maximum number of steps is limited to 4,000 steps.
\end{itemize}

\subsection{Results}
Table~\ref{table:ablation_il_rl_} compares the performance between imitation learning model and ours. As the task of approaching for animals does not appear in the training data of imitation learning, our method achieves a significant improvement in success rate in fewer steps compared to imitation learning model's zero shot. In the more difficult task of attacking cow, our method also achieves clear advantages. 
By further checking the videos, we find most of the successful cases are due to good luck without intention to actively approach the target. In the supplementary materials, we have prepared several demo videos, in which we can clearly see that the agent trained by our method has a clear awareness of completing the corresponding tasks.

\begin{table*}[h]
    \centering
    \caption{Compare imitation learning with our method on four Minecraft tasks.}
    \fontsize{8.0pt}{10.0pt}\selectfont
    \setlength{\tabcolsep}{1.5mm} 
    \setlength{\abovecaptionskip}{2mm} 
        \renewcommand\arraystretch{1.2}
    \resizebox{0.98\textwidth}{!}{
        \begin{tabular}{c|cc|cc|cc|cc}
         \toprule
        \multirow{2}{*}{\textbf{Method}} & \multicolumn{2}{c|}{\textbf{Approach Pig \includegraphics[width=0.015\textwidth]{figs/icon/pig.png} on Plains}} & \multicolumn{2}{c|}{\textbf{Approach Sheep \includegraphics[width=0.015\textwidth]{figs/icon/sheep.jpg} on Plains}} & \multicolumn{2}{c|}{\textbf{Approach Chicken \includegraphics[width=0.015\textwidth]{figs/icon/chicken.png} on Plains}} & \multicolumn{2}{c}{\textbf{Attack Cow \includegraphics[width=0.015\textwidth]{figs/icon/diamond_sword.png} \includegraphics[width=0.015\textwidth]{figs/icon/cow.jpg} on Plains}} \\ 
        \cline{2-9}
        & avg. step $\downarrow$ & succ. (\%) $\uparrow$ & avg. step $\downarrow$ & succ. (\%) $\uparrow$ & avg. step $\downarrow$ & succ. (\%) $\uparrow$ & avg. step $\downarrow$ & succ. (\%) $\uparrow$  \\ \hline
        \textbf{Imitation Learning} & 1664 & 25.0 & 1734 & 23.5 & 1496 & 36.4 & 3522 & 13.3 \\
        \textbf{Ours} & \textbf{830} & \textbf{67.3} & \textbf{1024} & \textbf{60.6} & \textbf{1391} & \textbf{54.5} & \textbf{3133} & \textbf{40.8} \\
        \bottomrule
\end{tabular}}
    \label{table:ablation_il_rl_}
\end{table*}

\section{Experiments of scalability on MiniGrid Environment}

To verify the scalability and effectiveness of Auto MC-reward when applied on new environments other than Minecraft, we conduct preliminary experiments on the cross-lava task of the MiniGrid environment~\cite{MinigridMiniworld23}, where the agent is required to reach the goal point on the other corner of the room while avoiding rivers of deadly lava. We choose two environment settings of different difficulty levels:  \texttt{LavaCrossingS9N1} with map size $9\times9$ and $1$ lava stream, and  \texttt{LavaCrossingS9N3} with map size $9\times9$ and $3$ lava streams. We modify the prompts of game information, task description, and input parameters to adapt to the MiniGrid environment, but other parts (\eg, reward function requirements and format) remain unchanged. The example environment visualization and results are shown in Figure~\ref{fig:minigrid}. Auto MC-Reward shows its superiority to original sparse reward, indicating its effectiveness in new environments. 

\begin{figure}[htbp]
  \centering
  \begin{subfigure}[b]{0.2\linewidth}
    \includegraphics[width=\linewidth]{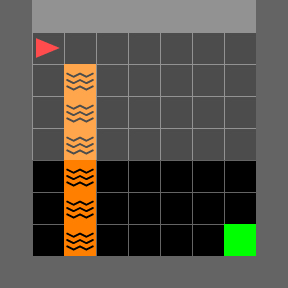}
    \label{fig:sub1}
  \end{subfigure}
  \hspace{5mm}
  \begin{subfigure}[b]{0.45\linewidth}
    \includegraphics[width=\linewidth]{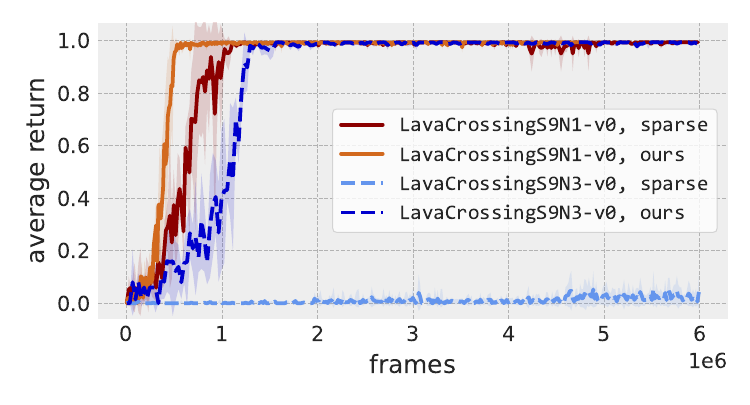}
    \label{fig:sub2}
  \end{subfigure}
\caption{(Left) Example environment visualization of MiniGrid \texttt{LavaCrossingS9N1} environment. (Right) Performance comparison on MiniGrid.}
  \label{fig:minigrid}
\end{figure}

\section{Experiments with Different LLMs}
Our framework leverages different LLM capabilities: code comprehension and generation for Designer and Critic, and JSON-like data interpretation for Analyzer. 
In Fig.~\ref{fig:llm}~(left), we investigate different LLMs. We use the case of avoiding lava from Sec.~\ref{sec:sup-interaction} to test whether each LLM can analyze that lava is the failure reason, and whether it can add the penalty for lava into the reward function. Notably, small LLMs like Mistral Medium show better performance than GPT-4. We hope that future work could provide more comprehensive evaluation to guide the selection of LLMs.

\begin{figure}[h]
\centering
\includegraphics[width=0.4\linewidth]{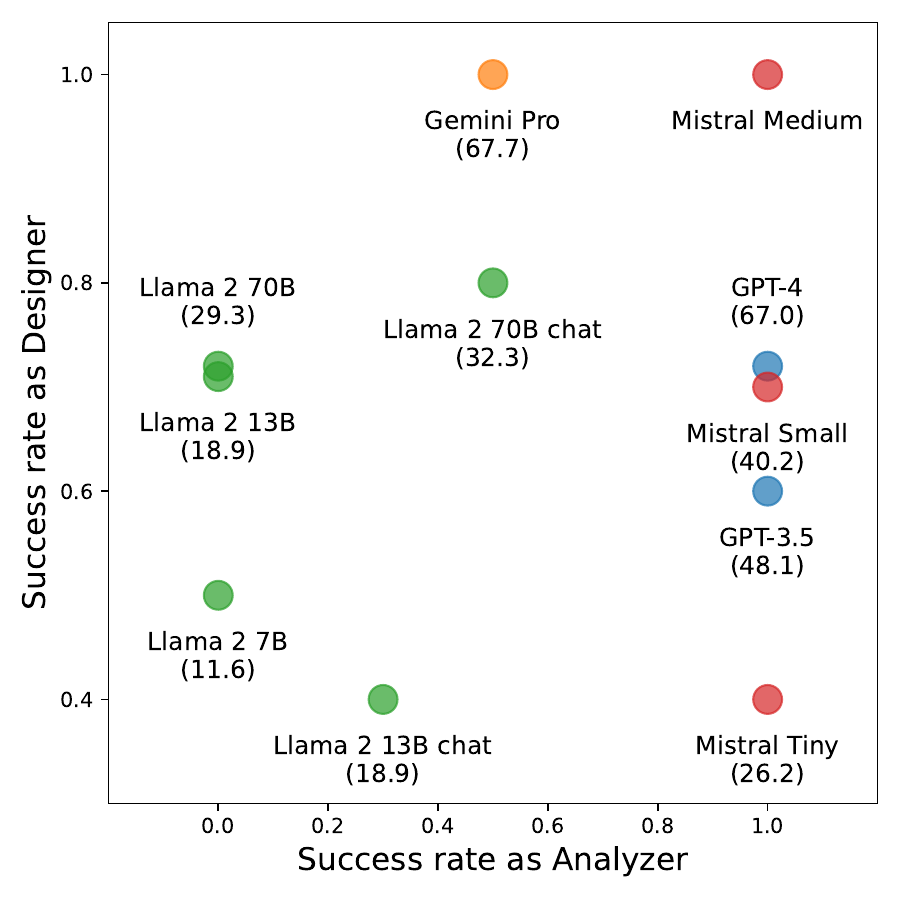}
\caption{Effect of different LLMs. Numbers in the parentheses are benchmark scores of HumanEval on code generation.}
\label{fig:llm}
\end{figure}

\section{Foundation Model Imitation Learning} 
\subsection{Pretraining Data}
When large labeled datasets do not exist, the canonical strategy for training capable agents is RL, which is inefficient and expensive to sample for hard-exploration problems~\cite{berner2019dota,jaderberg2019human,baker2022video}, \emph{e.g.} mining diamond in Minecraft. Therefore, in order to more efficiently explore the effectiveness of the LLM-based reward function design mechanism proposed in this paper, we pre-trained some foundation models through imitation learning as done by VPT~\cite{baker2022video}. 
Specifically, we use GITM~\cite{zhu2023ghost} to continuously perform Diamond Mining task and record important observation data of each frame, such as RGB, action, inventory, compass (\emph{e.g.} y$\_$level), structured action (action plans), etc. The following is the format of the collected data:

\begin{tcolorbox}[breakable=true, boxrule={0.5pt}, sharp corners={all}]
\setlength{\parskip}{1ex}
\footnotesize
\{\\
\hbox{\ \ \ \ }"action\_plans": [\\
\hbox{\ \ \ \ \ \ \ \ }\{\\
\hbox{\ \ \ \ \ \ \ \ \ \ \ \ }"name": "mine",\\
\hbox{\ \ \ \ \ \ \ \ \ \ \ \ }"args": \{\\
\hbox{\ \ \ \ \ \ \ \ \ \ \ \ \ \ \ \ }"object": \{"log": 1\},\\
\hbox{\ \ \ \ \ \ \ \ \ \ \ \ \ \ \ \ }"tool": None\\
\hbox{\ \ \ \ \ \ \ \ \ \ \ \ }\},\\
\hbox{\ \ \ \ \ \ \ \ \ \ \ \ }"expectation": "Obtain 1 log"\\
\hbox{\ \ \ \ \ \ \ \ }\}, \\
\hbox{\ \ \ \ \ \ \ \ }...\\
\hbox{\ \ \ \ }],\\
\\
\hbox{\ \ \ \ }"actions": [\\
\hbox{\ \ \ \ \ \ \ \ }[0, 0, 0, 12, 12, 3, 0, 0], ...\\
\hbox{\ \ \ \ }],\\
\\
\hbox{\ \ \ \ }"y\_level": [66, ...],\\
\\
\hbox{\ \ \ \ }"action\_indices": [100, ...],\\
\\
\hbox{\ \ \ \ }"biomes": ["Forest", ...],\\
\\
\hbox{\ \ \ \ }"deaths": [0, ...],\\
\\
\hbox{\ \ \ \ }"visible\_blocks": [\\
\hbox{\ \ \ \ \ \ \ \ }[\\
\hbox{\ \ \ \ \ \ \ \ \ \ \ \ }\{"name": "wood", "nearest\_distance": 0.71\}, ...\\
\hbox{\ \ \ \ \ \ \ \ }],\\
\hbox{\ \ \ \ \ \ \ \ }...\\
\hbox{\ \ \ \ }],\\
\\
\hbox{\ \ \ \ }"visible\_entities": [\\
\hbox{\ \ \ \ \ \ \ \ }[\\
\hbox{\ \ \ \ \ \ \ \ \ \ \ \ }\{"name": "cow", "nearest\_distance": 4.57\}, ...\\
\hbox{\ \ \ \ \ \ \ \ }],\\
\hbox{\ \ \ \ \ \ \ \ }...\\
\hbox{\ \ \ \ }],\\
\\
\hbox{\ \ \ \ }"inventory": [\{"log": 1\}, ...],\\
\\
\hbox{\ \ \ \ }"task\_names": ["Obtain 1 log", ...],\\
\\
\hbox{\ \ \ \ }"task\_indices": [88, ...],\\
\}
  
\end{tcolorbox}

Finally, we collect approximately 20.5/5.3 million image data in forest/plains biome, as listed in Table~\ref{table:gitm_data}. To reduce training costs, we only use 9 million and 2 million image data for training and validation in forest and plains biomes, respectively, totaling about 153 hours of game video (the control frequency is 20 Hz).

\begin{table}[h]
    \centering
    \caption{Detailed statistics of generated trajectories.}
    \fontsize{8.0pt}{10.0pt}\selectfont
    \setlength{\tabcolsep}{1.5mm} 
    \setlength{\abovecaptionskip}{2mm} 
        \renewcommand\arraystretch{1.2}
    \resizebox{0.98\textwidth}{!}{
        \begin{tabular}{c|cccc|cccc}
         \toprule
        \multirow{2}{*}{\textbf{Item}} & \multicolumn{4}{c|}{\textbf{Forest Biome}} & \multicolumn{4}{c}{\textbf{Plains Biome}} \\ 
        \cline{2-9}
        & trajectories & frames & max frames / traj.  & min frames / traj. & trajectories & frames & max frames / traj.  & min frames / traj. \\ \hline
        log \includegraphics[width=0.015\textwidth]{figs/icon/oak_log.png} & 14,790 & 5,209,246 & 13,806 & 8 & 3,254 & 2,425,602 & 14,373 & 10 \\
        planks \includegraphics[width=0.015\textwidth]{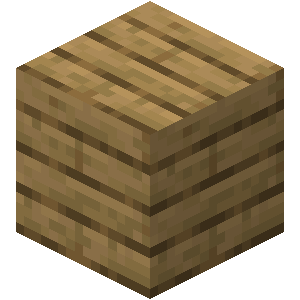} & 16,452 & 16,452 & 1 & 1 & 3,647 & 3,647 & 1 & 1 \\
        stick \includegraphics[width=0.015\textwidth]{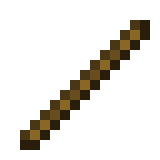} & 7,198 & 9,076 & 2 & 1 & 1,550 & 1,908 & 2 & 1 \\
        crafting table \includegraphics[width=0.015\textwidth]{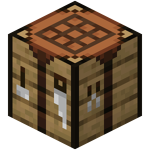} & 3,452 & 3,452 & 1 & 1 & 837 & 837 & 1 & 1  \\
        wooden pickaxe \includegraphics[width=0.015\textwidth]{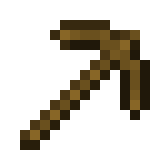} & 2,972 & 408,056 & 2,263 & 72 & 685 & 97,529 & 2,328 & 106 \\
        cobblestone \includegraphics[width=0.015\textwidth]{figs/icon/cobblestone.png} & 3,047 & 1,704,098 & 13,107 & 180 & 681 & 341,737 & 6,681 & 184 \\
        stone pickaxe \includegraphics[width=0.015\textwidth]{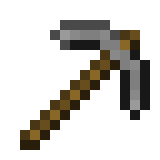} & 2,729 & 414,750 & 4,473 & 57 & 607 & 86,806 & 2,486 & 108 \\
        iron ore \includegraphics[width=0.015\textwidth]{figs/icon/iron_ore.png} & 2,177 & 6,498,252 & 14,833 & 200 & 461 & 1,394,251 & 12,783 & 235 \\
        furnace \includegraphics[width=0.015\textwidth]{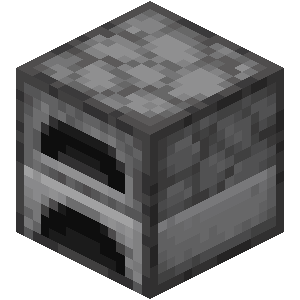} & 1,879 & 280,604 & 7,912 & 86 & 362 & 55,101 & 2,885 & 28 \\
        iron ingot \includegraphics[width=0.015\textwidth]{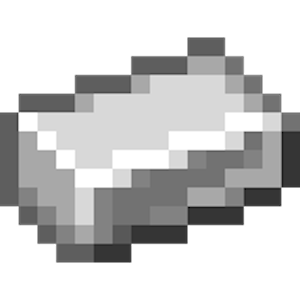} & 1,632 & 497,513 & 7,334 & 17 & 281 & 91,585 & 4,916 & 254 \\
        iron pickaxe \includegraphics[width=0.015\textwidth]{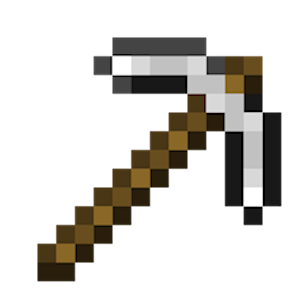} & 1,588 & 189,718 & 877 & 38 & 274 & 32,450 & 656 & 106 \\
        diamond \includegraphics[width=0.015\textwidth]{figs/icon/diamond.png} & 756 & 5,285,204 & 14,989 & 1,154 & 119 & 766,785 & 14,634 & 1,226 \\
        total & 58,672 & 20,516,421 & - & - & 12,758 & 5,298,238 & - & - \\
        \bottomrule
\end{tabular}}
    \label{table:gitm_data}
\end{table}

\subsection{Action Space}
We select 14 actions to train imitation learning for four tasks (obtain log/cobblestone/iron ore/diamond), as shown in Table~\ref{tab:action_space}.

\begin{table}[h]
\renewcommand\arraystretch{1.4}
\setlength\tabcolsep{4pt}
\footnotesize
\centering		
\caption{Action space used in our method.}
\label{tab:action_space}
\resizebox{0.75\textwidth}{!}
{
    \begin{tabular}{ccc|ccc}
    \toprule
     \textbf{Index} & \textbf{Action Name} & \textbf{Action Box} & \textbf{Index} & \textbf{Action Name} & \textbf{Action Box} \\
     \hline
     0 & no$\_$op & [0, 0, 0, 12, 12, 0, 0, 0] & 7 & jump  & [0, 0, 1, 12, 12, 0, 0, 0] \\
     1 & turn$\_$up & [0, 0, 0, 11, 12, 0, 0, 0] & 8 & back & [2, 0, 0, 12, 12, 0, 0, 0] \\
     2 & turn$\_$down & [0, 0, 0, 13, 12, 0, 0, 0] & 9 & move$\_$left & [0, 1, 0, 12, 12, 0, 0, 0] \\
     3 & turn$\_$left & [0, 0, 0, 12, 11, 0, 0, 0] & 10 & move$\_$right & [0, 2, 0, 12, 12, 0, 0, 0] \\
     4 & turn$\_$right & [0, 0, 0, 12, 13, 0, 0, 0] & 11 & attack & [0, 0, 0, 12, 12, 3, 0, 0] \\
     5 & forward & [1, 0, 0, 12, 12, 0, 0, 0] & 12 & equip & [0, 0, 0, 12, 12, 5, 0, 0]  \\
     6 & forward$\_$jump & [1, 0, 1, 12, 12, 0, 0, 0] & 13 & jump$\_$place & [0, 0, 1, 12, 12, 6, 0, 0]\\
    
    \bottomrule
    \end{tabular}
} 
\end{table}

\subsection{Training Details}
Subsequently, we conduct fully supervised training on the four main tasks of the Diamond mining Task, \emph{i.e.} log, cobblestone, iron ore, diamond, by using Impala CNN~\cite{espeholt2018impala} and Transformer~\cite{vaswani2017attention} as backbone, and obtained several foundation models. The main differences between the foundation models are different biomes (forest and plains), temporal frames (16 and 128), and whether goal and y$\_$level embeddings are used as conditions. We use 32 A800 GPUs for foundation model training, and initialize the network using the foundation model of VPT to speed up the training of imitation learning. The hyperparameters are listed in Table~\ref{tab:hyperparameter_il}.

\begin{table}[h]
\renewcommand\arraystretch{1.4}
\setlength\tabcolsep{4pt}
\footnotesize
\centering		
\caption{Hyperparameters for imitation learning.}
\label{tab:hyperparameter_il}
\resizebox{0.22\textwidth}{!}
{
    \begin{tabular}{cc}
    \toprule
     \textbf{Hyperparameter} & \textbf{Value} \\
     \hline
     Learning rate & 1e-4 \\
     Batch size & 128 \\
     Epochs & 25 \\
     Optimizer & Adam \\
    
    \bottomrule
    \end{tabular}
} 
\end{table}

The obtained foundation models, especially the goal-conditioned one, already have satisfactory basic behavioral capabilities. Based on these models, this paper uses the proposed methods to further improve existing behaviors (\emph{e.g.} avoid lava while exploring diamond) and complete new tasks (\emph{e.g.} approaching / killing animals).

\clearpage
\section{Reinforcement Learning Training} 
We use proximal policy optimization (PPO) algorithm~\cite{schulman2017proximal} with generalized advantage estimation (GAE)~\cite{schulman2015high} implemented in TorchRL~\cite{bou2023torchrl} library to train our RL model. 
The main hyperparameters used for RL training are listed in Table~\ref{tab:hyperparameter_rl}.

\begin{table}[h]
\renewcommand\arraystretch{1.4}
\setlength\tabcolsep{4pt}
\footnotesize
\centering		
\caption{Hyperparameters for reinforcement learning.}
\label{tab:hyperparameter_rl}
\resizebox{0.3\textwidth}{!}
{
    \begin{tabular}{cc}
    \toprule
     \textbf{Hyperparameter} & \textbf{Value} \\
     \hline
     Learning rate & 5e-5 \\
     Total training frames & 256,000 \\
     Batches per iteration & 64 \\
     Discount factor ($\gamma$) & 0.99 \\
     GAE $\lambda$ & 0.95 \\
     Value pre-training iteration & 2 \\
     PPO clip & 0.1 \\
     Entropy loss weight & 0.1 \\
     Critic loss type & $L_{2}$ \\
     Critic coefficient & 1 \\
     Optimizer & Adam \\
    \bottomrule
    \end{tabular}
} 
\end{table}

In addition to the above hyperparameters, several training techniques are applied to all of our experiments:
\begin{itemize}
    \item We follow VPT~\cite{baker2022video} to apply an auxiliary Kullback-Leibler (KL) divergence loss between the RL model and the frozen pre-trained policy to prevent catastrophically forgetting or overly aggressive policy update during RL training, \eg, maintaining the ability to dig horizontal tunnels. This loss is defined as:
    \begin{equation}
        \begin{aligned}
            L_{kl} = \alpha \text{KL}(\pi_{pt}, \pi_{\theta})
        \end{aligned}
    \end{equation}
    where $\pi_{\theta}$ is the the policy being trained, $\pi_{pt}$ is the frozen pretrained policy, $\text{KL}(\pi_{pt}, \pi_{\theta})$is the Kullback-Leibler divergence between the policy being trained and the pretrained policy, and $\alpha$ is the loss weight that is set to 0.125 by default.
    \item We also normalize the reward based on the trajectory returns to constrain the return scales of different tasks. The formula is as follows:
        \begin{equation}
            \begin{aligned}
                R' &= \text{mean}(\text{abs}(R)) \\
                R'_{cur} &= \beta R'_{cur} + (1-\beta) R'_{pre} \\
                r_{norm} &= \frac{r}{\text{max}(R'_{cur}, 1)}
            \end{aligned}
        \end{equation}
        where $r$ and $R$ are reward and trajectory returns. $\beta$ is a momentum parameter, set as 0.7 by default.
\end{itemize}

\clearpage
\section{Full Text Prompts}

We provide all the LLM input prompts of our method in this section. 

\subsection{Task Description}

The task description contains task objective, initial status, success criteria and procedure of each task. It describe basic information of a task without 
prior knowledge like game strategies.
\vspace{.5em}

\noindent\textbf{Exploring Diamond Ore \includegraphics[width=0.02\textwidth]{figs/icon/diamond_ore.png}}
\begin{tcolorbox}[breakable=true, boxrule={0.5pt}, sharp corners={all}]
\setlength{\parskip}{1ex}
\small

\#\# Task description

- Objective: Find and approach diamond with higher success rate and avoid death.

- Initial Status: 
\\
\hbox{\ \ }1. The agent is under ground at y level 11, the most common level to find diamonds.\\
\hbox{\ \ }2. The agent already has an iron pickaxe.\\
- Success criteria: The distance to the nearest diamond block is less than 2 meters. 

- Procedure: Explore horizontally to find a diamond, face it and approach it.

\end{tcolorbox}

\noindent\textbf{Approaching a Tree \includegraphics[width=0.02\textwidth]{figs/icon/oak_log.png} on Plains \includegraphics[width=0.02\textwidth]{figs/icon/plains.png}}
\begin{tcolorbox}[breakable=true, boxrule={0.5pt}, sharp corners={all}]
\setlength{\parskip}{1ex}
\small

\#\# Task description

- Objective: Find and approach wood in the plains biome.

- Initial Status: The agent is on the ground.

- Success Criteria: The distance to the nearest wood block is less than 1 meters.

- Procedure: Find a wood, face it and approach it.

- Additional Information: Wood is sparse in plains biome. The agent may need to walk a long distance to find and approach a wood.

\end{tcolorbox}

\noindent\textbf{Approaching a Cow \includegraphics[width=0.02\textwidth]{figs/icon/cow.jpg} on Plains \includegraphics[width=0.02\textwidth]{figs/icon/plains.png}}
\begin{tcolorbox}[breakable=true, boxrule={0.5pt}, sharp corners={all}]
\setlength{\parskip}{1ex}
\small

\#\# Task description

- Objective: Find and approach cow in the plains biome.

- Initial Status: The agent is on the ground.

- Success Criteria: The distance to the nearest cow is less than 2 meters.

- Procedure: Find a cow, face it and approach it.

\end{tcolorbox}

\noindent\textbf{Attacking a Cow \includegraphics[width=0.02\textwidth]{figs/icon/diamond_sword.png} \includegraphics[width=0.02\textwidth]{figs/icon/cow.jpg} on Plains \includegraphics[width=0.02\textwidth]{figs/icon/plains.png}}
\begin{tcolorbox}[breakable=true, boxrule={0.5pt}, sharp corners={all}]
\setlength{\parskip}{1ex}
\small

\#\# Task description

- Objective: Find, approach and kill cow in the plains biome.

- Initial Status: The agent is on the ground.

- Success Criteria: The agent successfully kills a cow.

- Procedure: Find a cow, face it, approach it and kill it.

\end{tcolorbox}

\clearpage
\subsection{Reward Designer}

Prompts for Reward Designer include prompts for initializing reward function, updating reward function, and handling critic failure and execution error.

\vspace{.5em}
\noindent\textbf{Prompt for Initializing Reward Function}
\begin{tcolorbox}[breakable=true, boxrule={0.5pt}, sharp corners={all}]
\setlength{\parskip}{1ex}
\small

You are now a proficient Minecraft player. You should help me write proper reward functions to train a Minecraft agent with reinforcement learning to complete the described task.

\texttt{\{task\_description\}}

\#\# Input parameters of the reward function

- current\_nearest\_blocks: nearby block types and nearest distance of each block type at the **current step**. It is a dictionary, consisting of the names, relative distances with the agent, and relative angles (yaw and pitch) with the agent, in the form of: \{NAME\_1: (DIST\_1, YAW\_1, PITCH\_1), NAME\_2: (DIST\_2, YAW\_2, PITCH\_2)\}. For example, \{"wood": (24.7, 1.48, -1.57), "cobblestone": (1.3, -0.17, 1.05), "iron\_ore": (4.5, 0.61, -0.17)\}. If a certrain block type does not exist in the field of view, it is not present in the dictionary. The yaw and pitch here is relative to the agent's front, i.e., pitch = 0 and yaw = 0 is the front of the agent.

- previou\_nearest\_blocks: nearby block types and nearest distance of each block type at the **previous step**, with the same format as the current\_nearest\_blocks.

- inventory\_change: the change of the agent's inventory from the previous step to current step, in the form of a dictionary: \{NAME\_1: CHANGE\_NUM\_1, NAME\_2: CHANGE\_NUM\_2\}. Positive values mean increase and negative values mean decrease. For example, \{"wood": 2, "dirt": 3, "stone\_pickaxe": -1\}.

- health: an integer value in range 0 to 10 indicating the health level of the agent. 0 means death and 10 means full health.

- past\_agent\_positions: the history of location of agent, in the form of a list: [[x1, y1, z1, yaw1, pitch1], [x2, y2, z2, yaw2, pitch2], ...]. The yaw and pitch here are relative to the agent's initial forward direction, i.e., pitch = 0 and yaw = 0 is the front of the agent when it was born. The length of the list is the number of steps the agent has taken. The last element of the list is the current location of the agent.

- GLOBAL\_DATA: a global variable. It is initialized as a dictionary. You can save necessary information between different steps with it.

\#\# General game information

- The version of Minecraft is 1.11.

- Common block names : dirt, cobblestone, iron, diamond, wood, coal, water, air, lava, leaves, ... Collected item names are the same.

- FOV is set to -35 to 35 degrees for yaw and -30 to 30 degrees for pitch. The max visible distance of blocks is 64.

- Length of each block is 1 meter.

- The y coordinate of the agent is the agent's height. The larger the y, the higher the agent. The ground level is around y = 63, but is not fixed.

- At each step, the agent selects one operation in is action space. The action space includes doing nothing (i.e. staying still), moving forward and backward, jumping and attacking. The action space also includes turning left, right, up and down. 

- The attack range is 2 meters in front of the agent. The agent can move 0.2 meters in each step. The agent can turn 15 degrees in each step. Typically, the agent needs to stay still and attack for **60 successive steps** to break a block.

- The hunger value is always at the max level.

\#\# Reward function requirements

- You should write a dense reward function \textasciigrave dense\textasciigrave  and a sparse reward function \textasciigrave sparse\textasciigrave . The sparse reward indicates achieving the goal or receiving heavy punishment. The dense reward provides intermediate signal to guide the agent in the process of achieving the goal. The magnitude of the return value does not matter, but the sign (positive or negative) is important. The final reward will be \textasciigrave np.sign(sparse(...)) * 1 + np.sign(dense(...)) * 0.1\textasciigrave .

\#\# Output Requirements

- The reward function should be written in Python 3.9.

- Output the code block only. **Do not output anything else outside the code block**.

- You should include **sufficient comments** in your reward function to explain your thoughts, the objective and **implementation details**. The implementation can be specified to a specific line of code.

- If you need to import packages (e.g. math, numpy) or define helper functions, define them at the beginning of the function. Do not use unimported packages and undefined functions.

\#\# Output format

Strictly follow the following format. **Do not output anything else outside the code block**.

def reward\_function(current\_nearest\_blocks, previous\_nearest\_blocks, inventory\_change, health, past\_agent\_positions, GLOBAL\_DATA):
\\
\hbox{\ \ \ \ }\# Thoughts:\\
\hbox{\ \ \ \ }\# ...\\
\hbox{\ \ \ \ }\# (import packages and define helper functions)\\
\hbox{\ \ \ \ }import numpy as np\\
\hbox{\ \ \ \ }...\\
\hbox{\ \ \ \ }def dense(current\_nearest\_blocks, previous\_nearest\_blocks, inventory\_change, health, past\_agent\_positions, GLOBAL\_DATA):\\
\hbox{\ \ \ \ \ \ \ \ }...\\
\hbox{\ \ \ \ }def sparse(current\_nearest\_blocks, previous\_nearest\_blocks, inventory\_change, health, past\_agent\_positions, GLOBAL\_DATA):\\
\hbox{\ \ \ \ \ \ \ \ }...\\
\hbox{\ \ \ \ }dense\_reward = dense(current\_nearest\_blocks, previous\_nearest\_blocks, inventory\_change, health, past\_agent\_positions, GLOBAL\_DATA)\\
\hbox{\ \ \ \ }sparse\_reward = sparse(current\_nearest\_blocks, previous\_nearest\_blocks, inventory\_change, health, past\_agent\_positions, GLOBAL\_DATA)\\
\hbox{\ \ \ \ }return np.sign(sparse\_reward) * 1 + np.sign(dense\_reward) * 0.1\\
\hbox{\ \ \ \ }...\\
Now write a reward function. Then in each iteration, I will use the reward function to train an RL agent, and test it in the environment. I will give you possible reasons of the failure found during the testing, and you should modify the reward function accordingly.

\end{tcolorbox}

\noindent\textbf{Prompt for Updating Reward Function}
\begin{tcolorbox}[breakable=true, boxrule={0.5pt}, sharp corners={all}]
\setlength{\parskip}{1ex}
\small

You are now a proficient Minecraft player. I now have a reward function to train a Minecraft agent with reinforcement learning to complete the described task. The reward function is used to train the reinforcement learning agent for several times. I will provide you the analysis of failure and inefficiency and suggestions. You should help me modify the reward function.

\texttt{\{task\_description\}}

\#\# Input parameters of the reward function

- current\_nearest\_blocks: nearby block types and nearest distance of each block type at the **current step**. It is a dictionary, consisting of the names, relative distances with the agent, and relative angles (yaw and pitch) with the agent, in the form of: \{NAME\_1: (DIST\_1, YAW\_1, PITCH\_1), NAME\_2: (DIST\_2, YAW\_2, PITCH\_2)\}. For example, \{"wood": (24.7, 1.48, -1.57), "cobblestone": (1.3, -0.17, 1.05), "iron\_ore": (4.5, 0.61, -0.17)\}. If a certrain block type does not exist in the field of view, it is not present in the dictionary. The yaw and pitch here is relative to the agent's front, i.e., pitch = 0 and yaw = 0 is the front of the agent.

- previou\_nearest\_blocks: nearby block types and nearest distance of each block type at the **previous step**, with the same format as the current\_nearest\_blocks.

- inventory\_change: the change of the agent's inventory from the previous step to current step, in the form of a dictionary: \{NAME\_1: CHANGE\_NUM\_1, NAME\_2: CHANGE\_NUM\_2\}. Positive values mean increase and negative values mean decrease. For example, \{"wood": 2, "dirt": 3, "stone\_pickaxe": -1\}.

- health: an integer value in range 0 to 10 indicating the health level of the agent. 0 means death and 10 means full health.

- past\_agent\_positions: the history of location of agent, in the form of a list: [[x1, y1, z1, yaw1, pitch1], [x2, y2, z2, yaw2, pitch2], ...]. The yaw and pitch here are relative to the agent's initial forward direction, i.e., pitch = 0 and yaw = 0 is the front of the agent when it was born. The length of the list is the number of steps the agent has taken. The last element of the list is the current location of the agent.

- GLOBAL\_DATA: a global variable. It is initialized as a dictionary. You can save necessary information between different steps with it.

\#\# General game information

- The version of Minecraft is 1.11.

- Common block names : dirt, cobblestone, iron, diamond, wood, coal, water, air, lava, leaves, ... Collected item names are the same.

- FOV is set to -35 to 35 degrees for yaw and -30 to 30 degrees for pitch. The max visible distance of blocks is 64.

- Length of each block is 1 meter.

- The y coordinate of the agent is the agent's height. The larger the y, the higher the agent. The ground level is around y = 63, but is not fixed.

- At each step, the agent selects one operation in is action space. The action space includes doing nothing (i.e. staying still), moving forward and backward, jumping and attacking. The action space also includes turning left, right, up and down. 

- The attack range is 2 meters in front of the agent. The agent can move 0.2 meters in each step. The agent can turn 15 degrees in each step. Typically, the agent needs to stay still and attack for **60 successive steps** to break a block.

- The hunger value is always at the max level.

\#\# Reward function requirements

- You should write a dense reward function \textasciigrave dense\textasciigrave  and a sparse reward function \textasciigrave sparse\textasciigrave . The sparse reward indicates achieving the goal or receiving heavy punishment. The dense reward provides intermediate signal to guide the agent in the process of achieving the goal. The magnitude of the return value does not matter, but the sign (positive or negative) is important. The final reward will be \textasciigrave np.sign(sparse(...)) * 1 + np.sign(dense(...)) * 0.1\textasciigrave .

\#\# Current reward function

\texttt{\{current\_reward\_function\}}

\#\# Reward function test result

The reward function is used to train the reinforcement learning agent for several times. Here is some analysis of failure and inefficiency and suggestions:

\textasciigrave \textasciigrave \textasciigrave 

\texttt{\{analysis\}}

\textasciigrave \textasciigrave \textasciigrave

\#\# Requirements

Please consider the analysis and suggestions above and modify the reward function.

1. You can both modify the current lines and add new lines.

2. If necessary, you can write a **totally different** reward function than the current one.

3. Consider modifing the reward and penalty values in the current reward function to balance them.

4. In the first part of the reward function, you should provide your thoughts of modifying the reward function. **The thoughts should be concise.**

\#\# Output format

Strictly follow the following format. **Do not output anything else outside the code block**.

def reward\_function(current\_nearest\_blocks, previous\_nearest\_blocks, inventory\_change, health, past\_agent\_positions, GLOBAL\_DATA):
\\
\hbox{\ \ \ \ }\# Thoughts:\\
\hbox{\ \ \ \ }\# ...\\
\hbox{\ \ \ \ }\# (import packages and define helper functions)\\
\hbox{\ \ \ \ }import numpy as np\\
\hbox{\ \ \ \ }...\\
\hbox{\ \ \ \ }def dense(current\_nearest\_blocks, previous\_nearest\_blocks, inventory\_change, health, past\_agent\_positions, GLOBAL\_DATA):\\
\hbox{\ \ \ \ \ \ \ \ }...\\
\hbox{\ \ \ \ }def sparse(current\_nearest\_blocks, previous\_nearest\_blocks, inventory\_change, health, past\_agent\_positions, GLOBAL\_DATA):\\
\hbox{\ \ \ \ \ \ \ \ }...\\
\hbox{\ \ \ \ }dense\_reward = dense(current\_nearest\_blocks, previous\_nearest\_blocks, inventory\_change, health, past\_agent\_positions, GLOBAL\_DATA)\\
\hbox{\ \ \ \ }sparse\_reward = sparse(current\_nearest\_blocks, previous\_nearest\_blocks, inventory\_change, health, past\_agent\_positions, GLOBAL\_DATA)\\
\hbox{\ \ \ \ }return np.sign(sparse\_reward) * 1 + np.sign(dense\_reward) * 0.1\\
\hbox{\ \ \ \ }...\\
   
\end{tcolorbox}

\noindent\textbf{Prompt for Critic Failure}
\begin{tcolorbox}[breakable=true, boxrule={0.5pt}, sharp corners={all}]
\setlength{\parskip}{1ex}
\small

The reward function fails to pass the verification. The reasoning is: 

\textasciigrave \textasciigrave \textasciigrave 

\texttt{\{reasoning\}}

\textasciigrave \textasciigrave \textasciigrave 

The critique is:

\textasciigrave \textasciigrave \textasciigrave 

\texttt{\{critique\}}

\textasciigrave \textasciigrave \textasciigrave

\#\# Requirements

Please consider the reasoning and critique, and modify the reward function. 

1. If necessary, you can write a totally different reward function than the current one.

2. In the first part of the reward function, you should provide your thoughts of modifying the reward function. The thoughts should not directly copy the given reasoning or critique. **The thoughts should be concise.**

\#\# Reward function requirements

- You should write a dense reward function \textasciigrave dense\textasciigrave  and a sparse reward function \textasciigrave sparse\textasciigrave . The sparse reward indicates achieving the goal or receiving heavy punishment. The dense reward provides intermediate signal to guide the agent in the process of achieving the goal. The magnitude of the return value does not matter, but the sign (positive or negative) is important. The final reward will be \textasciigrave np.sign(sparse(...)) * 1 + np.sign(dense(...)) * 0.1\textasciigrave .

\#\# Output format

Strictly follow the following format. **Do not output anything else outside the code block**.

def reward\_function(current\_nearest\_blocks, previous\_nearest\_blocks, inventory\_change, health, past\_agent\_positions, GLOBAL\_DATA):
\\
\hbox{\ \ \ \ }\# Thoughts:\\
\hbox{\ \ \ \ }\# ...\\
\hbox{\ \ \ \ }\# (import packages and define helper functions)\\
\hbox{\ \ \ \ }import numpy as np\\
\hbox{\ \ \ \ }...\\
\hbox{\ \ \ \ }def dense(current\_nearest\_blocks, previous\_nearest\_blocks, inventory\_change, health, past\_agent\_positions, GLOBAL\_DATA):\\
\hbox{\ \ \ \ \ \ \ \ }...\\
\hbox{\ \ \ \ }def sparse(current\_nearest\_blocks, previous\_nearest\_blocks, inventory\_change, health, past\_agent\_positions, GLOBAL\_DATA):\\
\hbox{\ \ \ \ \ \ \ \ }...\\
\hbox{\ \ \ \ }dense\_reward = dense(current\_nearest\_blocks, previous\_nearest\_blocks, inventory\_change, health, past\_agent\_positions, GLOBAL\_DATA)\\
\hbox{\ \ \ \ }sparse\_reward = sparse(current\_nearest\_blocks, previous\_nearest\_blocks, inventory\_change, health, past\_agent\_positions, GLOBAL\_DATA)\\
\hbox{\ \ \ \ }return np.sign(sparse\_reward) * 1 + np.sign(dense\_reward) * 0.1\\
\hbox{\ \ \ \ }...\\
   
\end{tcolorbox}

\noindent\textbf{Prompt for Code Execution Error}
\begin{tcolorbox}[breakable=true, boxrule={0.5pt}, sharp corners={all}]
\setlength{\parskip}{1ex}
\small

You are now a proficient Minecraft player. I now have a reward function to train a Minecraft agent with reinforcement learning to complete the described task. The reward function now encounters an execution error. I will provide you the execution error and the reward function. You should help me modify the reward function.

\texttt{\{task\_description\}}

\#\# Input parameters of the reward function

- current\_nearest\_blocks: nearby block types and nearest distance of each block type at the **current step**. It is a dictionary, consisting of the names, relative distances with the agent, and relative angles (yaw and pitch) with the agent, in the form of: \{NAME\_1: (DIST\_1, YAW\_1, PITCH\_1), NAME\_2: (DIST\_2, YAW\_2, PITCH\_2)\}. For example, \{"wood": (24.7, 1.48, -1.57), "cobblestone": (1.3, -0.17, 1.05), "iron\_ore": (4.5, 0.61, -0.17)\}. If a certrain block type does not exist in the field of view, it is not present in the dictionary. The yaw and pitch here is relative to the agent's front, i.e., pitch = 0 and yaw = 0 is the front of the agent.

- previou\_nearest\_blocks: nearby block types and nearest distance of each block type at the **previous step**, with the same format as the current\_nearest\_blocks.

- inventory\_change: the change of the agent's inventory from the previous step to current step, in the form of a dictionary: \{NAME\_1: CHANGE\_NUM\_1, NAME\_2: CHANGE\_NUM\_2\}. Positive values mean increase and negative values mean decrease. For example, \{"wood": 2, "dirt": 3, "stone\_pickaxe": -1\}.

- health: an integer value in range 0 to 10 indicating the health level of the agent. 0 means death and 10 means full health.

- past\_agent\_positions: the history of location of agent, in the form of a list: [[x1, y1, z1, yaw1, pitch1], [x2, y2, z2, yaw2, pitch2], ...]. The yaw and pitch here are relative to the agent's initial forward direction, i.e., pitch = 0 and yaw = 0 is the front of the agent when it was born. The length of the list is the number of steps the agent has taken. The last element of the list is the current location of the agent.

- GLOBAL\_DATA: a global variable. It is initialized as a dictionary. You can save necessary information between different steps with it.

\#\# General game information

- The version of Minecraft is 1.11.

- Common block names : dirt, cobblestone, iron, diamond, wood, coal, water, air, lava, leaves, ... Collected item names are the same.

- FOV is set to -35 to 35 degrees for yaw and -30 to 30 degrees for pitch. The max visible distance of blocks is 64.

- Length of each block is 1 meter.

- The y coordinate of the agent is the agent's height. The larger the y, the higher the agent. The ground level is around y = 63, but is not fixed.

- At each step, the agent selects one operation in is action space. The action space includes doing nothing (i.e. staying still), moving forward and backward, jumping and attacking. The action space also includes turning left, right, up and down. 

- The attack range is 2 meters in front of the agent. The agent can move 0.2 meters in each step. The agent can turn 15 degrees in each step. Typically, the agent needs to stay still and attack for **60 successive steps** to break a block.

- The hunger value is always at the max level.

\#\# Reward function requirements

- You should write a dense reward function \textasciigrave dense\textasciigrave  and a sparse reward function \textasciigrave sparse\textasciigrave . The sparse reward indicates achieving the goal or receiving heavy punishment. The dense reward provides intermediate signal to guide the agent in the process of achieving the goal. The magnitude of the return value does not matter, but the sign (positive or negative) is important. The final reward will be \textasciigrave np.sign(sparse(...)) * 1 + np.sign(dense(...)) * 0.1\textasciigrave .

\#\# Current reward function

\texttt{\{current\_reward\_function\}}

\#\# Execution error traceback

\textasciigrave \textasciigrave \textasciigrave 

\texttt{\{execution\_error\}}

\textasciigrave \textasciigrave \textasciigrave

\#\# Requirements

Please consider the reason of error and modify the reward function. 

In the first part of the reward function, you should write your analysis of the error in comments, and provide your thoughts of modifying the reward function. **The analysis and thoughts should be concise.**

\#\# Output format

Strictly follow the following format. **Do not output anything else outside the code block**.

def reward\_function(current\_nearest\_blocks, previous\_nearest\_blocks, inventory\_change, health, past\_agent\_positions, GLOBAL\_DATA):
\\
\hbox{\ \ \ \ }\# Thoughts:\\
\hbox{\ \ \ \ }\# ...\\
\hbox{\ \ \ \ }\# (import packages and define helper functions)\\
\hbox{\ \ \ \ }import numpy as np\\
\hbox{\ \ \ \ }...\\
\hbox{\ \ \ \ }def dense(current\_nearest\_blocks, previous\_nearest\_blocks, inventory\_change, health, past\_agent\_positions, GLOBAL\_DATA):\\
\hbox{\ \ \ \ \ \ \ \ }...\\
\hbox{\ \ \ \ }def sparse(current\_nearest\_blocks, previous\_nearest\_blocks, inventory\_change, health, past\_agent\_positions, GLOBAL\_DATA):\\
\hbox{\ \ \ \ \ \ \ \ }...\\
\hbox{\ \ \ \ }dense\_reward = dense(current\_nearest\_blocks, previous\_nearest\_blocks, inventory\_change, health, past\_agent\_positions, GLOBAL\_DATA)\\
\hbox{\ \ \ \ }sparse\_reward = sparse(current\_nearest\_blocks, previous\_nearest\_blocks, inventory\_change, health, past\_agent\_positions, GLOBAL\_DATA)\\
\hbox{\ \ \ \ }return np.sign(sparse\_reward) * 1 + np.sign(dense\_reward) * 0.1\\
\hbox{\ \ \ \ }...\\
    
\end{tcolorbox}

\clearpage
\subsection{Reward Critic}
The prompt for Reward Critic judges whether the current reward function meets the requirements, and provides critique when the judgement fails.
\begin{tcolorbox}[breakable=true, boxrule={0.5pt}, sharp corners={all}]
\setlength{\parskip}{1ex}
\small

You are now a proficient Minecraft player. I will give you a reward function written in Python, which is used to train an RL agent to play Minecraft. 

You should help me to evaluate whether the reward function has met the requirements.

\texttt{\{task\_description\}}

\#\# Input parameters of the reward function

- current\_nearest\_blocks: nearby block types and nearest distance of each block type at the **current step**. It is a dictionary, consisting of the names, relative distances with the agent, and relative angles (yaw and pitch) with the agent, in the form of: \{NAME\_1: (DIST\_1, YAW\_1, PITCH\_1), NAME\_2: (DIST\_2, YAW\_2, PITCH\_2)\}. For example, \{"wood": (24.7, 1.48, -1.57), "cobblestone": (1.3, -0.17, 1.05), "iron\_ore": (4.5, 0.61, -0.17)\}. If a certrain block type does not exist in the field of view, it is not present in the dictionary. The yaw and pitch here is relative to the agent's front, i.e., pitch = 0 and yaw = 0 is the front of the agent.

- previou\_nearest\_blocks: nearby block types and nearest distance of each block type at the **previous step**, with the same format as the current\_nearest\_blocks.

- inventory\_change: the change of the agent's inventory from the previous step to current step, in the form of a dictionary: \{NAME\_1: CHANGE\_NUM\_1, NAME\_2: CHANGE\_NUM\_2\}. Positive values mean increase and negative values mean decrease. For example, \{"wood": 2, "dirt": 3, "stone\_pickaxe": -1\}.

- health: an integer value in range 0 to 10 indicating the health level of the agent. 0 means death and 10 means full health.

- past\_agent\_positions: the history of location of agent, in the form of a list: [[x1, y1, z1, yaw1, pitch1], [x2, y2, z2, yaw2, pitch2], ...]. The yaw and pitch here are relative to the agent's initial forward direction, i.e., pitch = 0 and yaw = 0 is the front of the agent when it was born. The length of the list is the number of steps the agent has taken. The last element of the list is the current location of the agent.

- GLOBAL\_DATA: a global variable. It is initialized as a dictionary. You can save necessary information between different steps with it.

\#\# General game information

- The version of Minecraft is 1.11.

- Common block names : dirt, cobblestone, iron, diamond, wood, coal, water, air, lava, leaves, ... Collected item names are the same.

- FOV is set to -35 to 35 degrees for yaw and -30 to 30 degrees for pitch. The max visible distance of blocks is 64.

- Length of each block is 1 meter.

- The y coordinate of the agent is the agent's height. The larger the y, the higher the agent. The ground level is around y = 63, but is not fixed.

- At each step, the agent selects one operation in is action space. The action space includes doing nothing (i.e. staying still), moving forward and backward, jumping and attacking. The action space also includes turning left, right, up and down. 

- The attack range is 2 meters in front of the agent. The agent can move 0.2 meters in each step. The agent can turn 15 degrees in each step. Typically, the agent needs to stay still and attack for **60 successive steps** to break a block.

- The hunger value is always at the max level.

\#\# Reward function requirements

- You should write a dense reward function \textasciigrave dense\textasciigrave  and a sparse reward function \textasciigrave sparse\textasciigrave . The sparse reward indicates achieving the goal or receiving heavy punishment. The dense reward provides intermediate signal to guide the agent in the process of achieving the goal. The magnitude of the return value does not matter, but the sign (positive or negative) is important. The final reward will be \textasciigrave np.sign(sparse(...)) * 1 + np.sign(dense(...)) * 0.1\textasciigrave .

\#\# Requirements

1. You should check whether the reward function meets the **reward function requirements** above. 

2. Your judgement should consider whether the comments and the detailed implementation code are consistent. You can judge whether a line is correct by its comment, but do only rely on the comments.

3. You should also check the format of the reward function. It should be like:

def reward\_function(current\_nearest\_blocks, previous\_nearest\_blocks, inventory\_change, health, past\_agent\_positions, GLOBAL\_DATA):
\\
\hbox{\ \ \ \ }\# Thoughts:\\
\hbox{\ \ \ \ }\# ...\\
\hbox{\ \ \ \ }\# (import packages and define helper functions)\\
\hbox{\ \ \ \ }import numpy as np\\
\hbox{\ \ \ \ }...\\
\hbox{\ \ \ \ }def dense(current\_nearest\_blocks, previous\_nearest\_blocks, inventory\_change, health, past\_agent\_positions, GLOBAL\_DATA):\\
\hbox{\ \ \ \ \ \ \ \ }...\\
\hbox{\ \ \ \ }def sparse(current\_nearest\_blocks, previous\_nearest\_blocks, inventory\_change, health, past\_agent\_positions, GLOBAL\_DATA):\\
\hbox{\ \ \ \ \ \ \ \ }...\\
\hbox{\ \ \ \ }dense\_reward = dense(current\_nearest\_blocks, previous\_nearest\_blocks, inventory\_change, health, past\_agent\_positions, GLOBAL\_DATA)\\
\hbox{\ \ \ \ }sparse\_reward = sparse(current\_nearest\_blocks, previous\_nearest\_blocks, inventory\_change, health, past\_agent\_positions, GLOBAL\_DATA)\\
\hbox{\ \ \ \ }return np.sign(sparse\_reward) * 1 + np.sign(dense\_reward) * 0.1\\
\hbox{\ \ \ \ }...\\
4. Consider the **Information about the game state** above.

5. **You don't need to check whether the reward function meets the task description**.

6. You should first provide a reasoning and decision. If the reward function fails, provide a critique to help to improve the reward function.

\#\# Output format

You should only respond in JSON format as described below:

\{
\\
\hbox{\ \ \ \ }"reasoning": "reasoning",\\
\hbox{\ \ \ \ }"success": boolean,\\
\hbox{\ \ \ \ }"critique": critique,\\
\}

Ensure the response can be parsed by Python \textasciigrave json.loads\textasciigrave  , e.g. no trailing commas, no single quotes, etc. Do not output anything else like explanations.

\#\# Reward function to be evaluated

\texttt{\{reward\_function\}}

\end{tcolorbox}

\clearpage
\subsection{Trajectory Analyzer}
The prompt for Trajectory Analyzer analyzes the trajectory information and statistics, and provides possible reasons of failure and suggestions on game strategy.
\begin{tcolorbox}[breakable=true, boxrule={0.5pt}, sharp corners={all}]
\setlength{\parskip}{1ex}
\small

You are now a proficient Minecraft player. I have trained an RL agent and tested for several times in the Minecraft game environment. 

I will give you the information of the failed test results, i.e. trajectories of actions, rewards and observations. You should help me write proper analysis of possible reasons of failure and inefficiency, and your suggestion on the game strategy.

\texttt{\{task\_description\}}

\#\# General game information

- The version of Minecraft is 1.11.

- Common block names : dirt, cobblestone, iron, diamond, wood, coal, water, air, lava, leaves, ... Collected item names are the same.

- FOV is set to -35 to 35 degrees for yaw and -30 to 30 degrees for pitch. The max visible distance of blocks is 64.

- Length of each block is 1 meter.

- The y coordinate of the agent is the agent's height. The larger the y, the higher the agent. The ground level is around y = 63, but is not fixed.

- At each step, the agent selects one operation in is action space. The action space includes doing nothing (i.e. staying still), moving forward and backward, jumping and attacking. The action space also includes turning left, right, up and down. 

- The attack range is 2 meters in front of the agent. The agent can move 0.2 meters in each step. The agent can turn 15 degrees in each step. Typically, the agent needs to stay still and attack for **60 successive steps** to break a block.

- The hunger value is always at the max level.

\#\# Input format

\#\#\# Failed trajectories

The failed trajectories are shown as a list, where each entry is a dictionary representing a trajectory. Note that **only the failed cases are provided.**

The each item of "history" is a list representing statistics of each single step.  When the history is too long, it will be truncated to the last 32 steps, indicated by the value of "truncated".

"dead" indicates whether the agent has died.

The format is:

[
\\
\hbox{\ \ \ \ }\{\\
\hbox{\ \ \ \ \ \ \ \ }"history": \{\\
\hbox{\ \ \ \ \ \ \ \ \ \ \ \ }"rewards": [reward1, ...],\\
\hbox{\ \ \ \ \ \ \ \ \ \ \ \ }"actions": [action1, action2, ...],\\
\hbox{\ \ \ \ \ \ \ \ \ \ \ \ }"locations": [[x1, y1, z1, pitch1, yaw1], ...],\\
\hbox{\ \ \ \ \ \ \ \ \ \ \ \ }"inventory\_change": \{item1: delta\_num1, ...\},\\
\hbox{\ \ \ \ \ \ \ \ \ \ \ \ }"truncated": True or False,\\
\hbox{\ \ \ \ \ \ \ \ }\},\\
\hbox{\ \ \ \ \ \ \ \ }"final\_health": health,\\
\hbox{\ \ \ \ \ \ \ \ }"final\_inventory": \{item1: num1, ...\},\\
\hbox{\ \ \ \ \ \ \ \ }"final\_nearest\_blocks": \{block\_name1: nearest\_distance1, block\_name2: nearest\_distance2, ...\},\\
\hbox{\ \ \ \ \ \ \ \ }"block\_under\_foot": block\_name,\\
\hbox{\ \ \ \ \ \ \ \ }"dead": True or False,\\
\hbox{\ \ \ \ }\},\\
\hbox{\ \ \ \ }...\\
]

where "final\_health" is in range 0 and 10. "block\_under\_foot" is the block right under the agent.

\#\#\# Statistics

The statistics is a dictionary of all the test results, including **both successful and failed trajectories**.

The format is:

\{
\\
\hbox{\ \ \ \ }"success\_rate" : success\_rate,\\
\}

\#\# Requirements

Please write proper analysis of possible reasons of failure and inefficiency, and your suggestion on the game strategy. You should not not be limited to the task description above, but also come up with other failure cases based on the test results.

**The analysis and suggestion should be concise.**

\#\# Test results

\#\#\# Failed trajectories

\texttt{\{trajectories\}}

\#\#\# Statistics

\texttt{\{statistics\}}

\end{tcolorbox} 

We also try to replace game-specific feedback information (i.e. trajectory information and statistics) with captions from multi-modal large language model (e.g. GPT-4V) to enhance the generality of the agent. First, we input the video sequence into GPT-4V and let it describe the sequence in as much detail as possible. The prompt is as follows:
\begin{tcolorbox}[breakable=true, boxrule={0.5pt}, sharp corners={all}]
\setlength{\parskip}{1ex}
\small
This is a Minecraft game video clip. Please describe the scene and the player's actions in detail. Note that describe the entire sequence of images rather than describing each image individually.   
\end{tcolorbox}

Then, the output of GPT-4V will be further combined with the prompt of the above trajectory analysis for another question and answer, allowing LLM to analyze the possible reasons for the failure. See Section \ref{sec:sup-gpt4v} for detailed examples.

\clearpage
\section{Examples of LLM-Designed Reward Functions}

We provide reward function examples for four tasks. For the attacking cow task, we add two extra input parameters \emph{damage\_dealt} and \emph{kill\_entity}, which records the amount of damage the agent caused and the amounts of each entity the agent killed, respectively.

\begin{algorithm}[h]
\caption{Reward Function Example of Exploring Diamond Ore \includegraphics[width=0.02\textwidth]{figs/icon/diamond_ore.png}}
\definecolor{codeblue}{rgb}{0.25,0.5,0.5}
\lstset{
  backgroundcolor=\color{white},
  basicstyle=\fontsize{7.2pt}{7.2pt}\ttfamily\selectfont,
  columns=fullflexible,
  breaklines=true,
  captionpos=b,
  commentstyle=\fontsize{7.2pt}{7.2pt}\color{codeblue},
  keywordstyle=\fontsize{7.2pt}{7.2pt},
}
\begin{lstlisting}[language=python]
def reward_function(current_nearest_blocks, previous_nearest_blocks, inventory_change, health, past_agent_positions, GLOBAL_DATA):
    """
    Thoughts:
    In order to make the agent approach the diamond, we need to give positive rewards for actions that
    make the agent get closer to the diamond and negative rewards for actions that make the agent get further from the diamond.
    We also need to give a negative reward if the agent's health decreases to encourage the agent to avoid dangerous situations.
    The dense reward function will be based on the distance to the diamond and health.
    The sparse reward function will give a large positive reward when the agent reaches the diamond and a large negative reward when the agent dies.
    In addition, we should implement a threat detection mechanism to avoid unnecessary deaths.
    """
    import numpy as np

    def dense(current_nearest_blocks, previous_nearest_blocks, inventory_change, health, past_agent_positions, GLOBAL_DATA):
        # Initialize reward
        reward = 0

        # If the agent is closer to diamond than before, give a positive reward
        if "diamond" in current_nearest_blocks and "diamond" in previous_nearest_blocks:
            if current_nearest_blocks["diamond"][0] < previous_nearest_blocks["diamond"][0]:
                reward += 1
            elif current_nearest_blocks["diamond"][0] > previous_nearest_blocks["diamond"][0]:
                reward -= 1
        
        # If the agent is close to lava, give a negative reward
        if "lava" in current_nearest_blocks:
            reward -= 1
        
        # If the agent is not in pitch range 0 to 45, give a negative reward
        cur_pitch = past_agent_positions[-1][-1]
        if cur_pitch < 0 or cur_pitch > 45:
            reward -= 2

        # If the agent's health decreases, give a negative reward
        if "health" in GLOBAL_DATA:
            if health < GLOBAL_DATA["health"]:
                reward -= 1
        GLOBAL_DATA["health"] = health

        return reward

    def sparse(current_nearest_blocks, previous_nearest_blocks, inventory_change, health, past_agent_positions, GLOBAL_DATA):
        # Initialize reward
        reward = 0

        # If the agent reaches the diamond, give a large positive reward
        if "diamond" in current_nearest_blocks and current_nearest_blocks["diamond"][0] < 2:
            reward += 100

        # If the agent dies, give a large negative reward
        if health == 0:
            reward -= 100

        return reward

    dense_reward = dense(current_nearest_blocks, previous_nearest_blocks, inventory_change, health, past_agent_positions, GLOBAL_DATA)
    sparse_reward = sparse(current_nearest_blocks, previous_nearest_blocks, inventory_change, health, past_agent_positions, GLOBAL_DATA)

    return np.sign(sparse_reward) * 1 + np.sign(dense_reward) * 0.1
\end{lstlisting}
\end{algorithm}

\begin{algorithm}[h]
\caption{Reward Function Example of Approaching a Tree \includegraphics[width=0.02\textwidth]{figs/icon/oak_log.png} on Plains \includegraphics[width=0.02\textwidth]{figs/icon/plains.png}.}
\definecolor{codeblue}{rgb}{0.25,0.5,0.5}
\lstset{
  backgroundcolor=\color{white},
  basicstyle=\fontsize{7.2pt}{7.2pt}\ttfamily\selectfont,
  columns=fullflexible,
  breaklines=true,
  captionpos=b,
  commentstyle=\fontsize{7.2pt}{7.2pt}\color{codeblue},
  keywordstyle=\fontsize{7.2pt}{7.2pt},
}
\begin{lstlisting}[language=python]
def reward_function(current_nearest_blocks, previous_nearest_blocks, inventory_change, health, past_agent_positions, GLOBAL_DATA):
    # Thoughts:
    # Objective: The agent should approach the nearest wood block in the plains biome.
    # We want to encourage the agent to:
    # - Make progress towards wood
    # - Keep health high
    # - Avoid circling around or staying still
    # We use a dense reward to encourage the above behaviors, and a sparse reward to give a big bonus when the agent achieves the objective.
    import numpy as np
    import math
    
    # Helper function to calculate the Euclidean distance between two points
    def calculate_distance(point1, point2):
        return math.sqrt(sum([(a - b) ** 2 for a, b in zip(point1, point2)]))
    
    def dense(current_nearest_blocks, previous_nearest_blocks, inventory_change, health, past_agent_positions, GLOBAL_DATA):
        reward = 0
        
        # Encourage the agent to approach wood
        if "wood" in current_nearest_blocks:
            current_distance = current_nearest_blocks["wood"][0]
            if "wood" in previous_nearest_blocks:
                previous_distance = previous_nearest_blocks["wood"][0]
                # Encourage the agent to decrease the distance to the nearest wood
                if current_distance < previous_distance:
                    reward += 1
                # Discourage the agent to increase the distance to the nearest wood
                else:
                    reward -= 1
            # Encourage the agent to find wood
            else:
                reward += 1
        
        # Encourage the agent to keep health high
        if health <= 3:
            reward -= 1
        
        # Discourage the agent from circling around or staying still
        if len(past_agent_positions) > 1:
            current_position = past_agent_positions[-1][:3]
            previous_position = past_agent_positions[-2][:3]
            # Encourage the agent to move
            if calculate_distance(current_position, previous_position) < 0.1:
                reward -= 1
        
        return reward
    
    def sparse(current_nearest_blocks, previous_nearest_blocks, inventory_change, health, past_agent_positions, GLOBAL_DATA):
        reward = 0
        
        # Give a big bonus when the agent achieves the objective
        if "wood" in current_nearest_blocks and current_nearest_blocks["wood"][0] < 1:
            reward += 10
        
        return reward
    
    dense_reward = dense(current_nearest_blocks, previous_nearest_blocks, inventory_change, health, past_agent_positions, GLOBAL_DATA)
    sparse_reward = sparse(current_nearest_blocks, previous_nearest_blocks, inventory_change, health, past_agent_positions, GLOBAL_DATA)
    
    return np.sign(sparse_reward) * 1 + np.sign(dense_reward) * 0.1
\end{lstlisting}
\end{algorithm}

\begin{algorithm}[h]
\caption{Reward Function Example of Approaching a Cow \includegraphics[width=0.02\textwidth]{figs/icon/cow.jpg} on Plains \includegraphics[width=0.02\textwidth]{figs/icon/plains.png}}
\definecolor{codeblue}{rgb}{0.25,0.5,0.5}
\lstset{
  backgroundcolor=\color{white},
  basicstyle=\fontsize{7.2pt}{7.2pt}\ttfamily\selectfont,
  columns=fullflexible,
  breaklines=true,
  captionpos=b,
  commentstyle=\fontsize{7.2pt}{7.2pt}\color{codeblue},
  keywordstyle=\fontsize{7.2pt}{7.2pt},
}
\begin{lstlisting}[language=python]
def reward_function(current_nearest_blocks, previous_nearest_blocks, inventory_change, health, past_agent_positions, GLOBAL_DATA):
    """
    Thoughts:
    - The agent needs to find a cow and approach it. So, the primary factor affecting the reward is the distance to the nearest cow.
    - The agent should avoid getting hurt (losing health).
    - We do not need to consider inventory change in this task.
    - We need to encourage the agent to keep exploring, so we also consider the agent's movement.
    """

    import numpy as np

    # Helper function to calculate the Euclidean distance between two positions
    def calculate_distance(pos1, pos2):
        return np.sqrt((pos1[0] - pos2[0]) ** 2 + (pos1[1] - pos2[1]) ** 2 + (pos1[2] - pos2[2]) ** 2)

    def dense(current_nearest_blocks, previous_nearest_blocks, health, past_agent_positions):
        # Reward for approaching the cow
        cow_reward = 0
        if "cow" in current_nearest_blocks:
            if "cow" in previous_nearest_blocks:
                # If the agent is getting closer to the cow, give a positive reward
                if current_nearest_blocks["cow"][0] < previous_nearest_blocks["cow"][0]:
                    cow_reward = 1
                # If the agent is getting farther from the cow, give a negative reward
                elif current_nearest_blocks["cow"][0] > previous_nearest_blocks["cow"][0]:
                    cow_reward = -1
        
        # Reward for moving (exploring)
        move_reward = 0
        if len(past_agent_positions) > 1:
            # If the agent is moving, give a positive reward
            if calculate_distance(past_agent_positions[-1], past_agent_positions[-2]) > 0:
                move_reward = 1
            # If the agent is not moving, give a negative reward
            else:
                move_reward = -1
        
        return cow_reward + move_reward

    def sparse(current_nearest_blocks, health):
        # If the agent finds a cow and approaches it successfully, give a large positive reward
        if "cow" in current_nearest_blocks and current_nearest_blocks["cow"][0] < 2:
            return 100
        # If the agent gets hurt, give a large negative reward
        elif health < 10:
            return -100
        else:
            return 0
    
    dense_reward = dense(current_nearest_blocks, previous_nearest_blocks, health, past_agent_positions)
    sparse_reward = sparse(current_nearest_blocks, health)
    
    return np.sign(sparse_reward) * 1 + np.sign(dense_reward) * 0.1

\end{lstlisting}
\end{algorithm}

\begin{algorithm}[h]
\caption{Reward Function Example of Attacking a Cow \includegraphics[width=0.02\textwidth]{figs/icon/diamond_sword.png} \includegraphics[width=0.02\textwidth]{figs/icon/cow.jpg} on Plains \includegraphics[width=0.02\textwidth]{figs/icon/plains.png}}
\definecolor{codeblue}{rgb}{0.25,0.5,0.5}
\lstset{
  backgroundcolor=\color{white},
  basicstyle=\fontsize{7.2pt}{7.2pt}\ttfamily\selectfont,
  columns=fullflexible,
  breaklines=true,
  captionpos=b,
  commentstyle=\fontsize{7.2pt}{7.2pt}\color{codeblue},
  keywordstyle=\fontsize{7.2pt}{7.2pt},
}
\begin{lstlisting}[language=python]
def reward_function(current_nearest_blocks, previous_nearest_blocks, inventory_change, health, past_agent_positions, GLOBAL_DATA, damage_dealt, kill_entity):
    """
    Thoughts:
    - The agent needs to find a cow and approach and kill it. We can give a positive dense reward for approaching or attacking a cow.
    - We can give a sparse reward for killing a cow.
    """

    import numpy as np

    def dense(current_nearest_blocks, previous_nearest_blocks, damage_dealt):
        # Reward for approaching the cow
        approach_reward = 0
        if "cow" in current_nearest_blocks and "cow" in previous_nearest_blocks:
            if current_nearest_blocks["cow"][0] < previous_nearest_blocks["cow"][0]:
                approach_reward = 1
            elif current_nearest_blocks["cow"][0] > previous_nearest_blocks["cow"][0]:
                approach_reward = -1
        
        # Reward for attacking the cow
        if damage_dealt > 0:
            attack_reward = 5
        else:
            attack_reward = 0
        
        return approach_reward + attack_reward

    def sparse(kill_entity):
        # Reward for killing the cow
        if kill_entity["cow"] > 0:
            return 100
        else:
            return 0
    
    dense_reward = dense(current_nearest_blocks, previous_nearest_blocks, damage_dealt)
    sparse_reward = sparse(kill_entity)
    
    return np.sign(sparse_reward) * 1 + np.sign(dense_reward) * 0.1
\end{lstlisting}
\end{algorithm}

\clearpage
\section{Example of Full Interactions with LLM}\label{sec:sup-interaction}

We show an example of interactions with the LLM for the diamond exploration task, including initializing reward function and one iteration of updating reward function. \texttt{\textbf{INPUT}} is the input prompt and \texttt{\textbf{OUTPUT}} is the response from GPT-4. 
We highlight some lines that are related to critic revising and reward updating in \hl{yellow}.

\begin{tcolorbox}[breakable=true, boxrule={0.5pt}, sharp corners={all}]
\setlength{\parskip}{1ex}
\small
\texttt{\textbf{DESIGNER INPUT:}}

You are now a proficient Minecraft player. You should help me write proper reward functions to train a Minecraft agent with reinforcement learning to complete the described task.

\#\# Task description

- Objective: Find and approach diamond with higher success rate and avoid death.

- Initial Status: 
\\
\hbox{\ \ }1. The agent is under ground at y level 11, the most common level to find diamonds.\\
\hbox{\ \ }2. The agent already has an iron pickaxe.\\
- Success criteria: The distance to the nearest diamond block is less than 2 meter. 

- Procedure: Explore horizontally to find a diamond, face it and approach it.

\#\# Input parameters of the reward function

- current\_nearest\_blocks: nearby block types and nearest distance of each block type at the **current step**. It is a dictionary, consisting of the names, relative distances with the agent, and relative angles (yaw and pitch) with the agent, in the form of: \{NAME\_1: (DIST\_1, YAW\_1, PITCH\_1), NAME\_2: (DIST\_2, YAW\_2, PITCH\_2)\}. For example, \{"wood": (24.7, 1.48, -1.57), "cobblestone": (1.3, -0.17, 1.05), "iron\_ore": (4.5, 0.61, -0.17)\}. If a certrain block type does not exist in the field of view, it is not present in the dictionary. The yaw and pitch here is relative to the agent's front, i.e., pitch = 0 and yaw = 0 is the front of the agent.

- previou\_nearest\_blocks: nearby block types and nearest distance of each block type at the **previous step**, with the same format as the current\_nearest\_blocks.

- inventory\_change: the change of the agent's inventory from the previous step to current step, in the form of a dictionary: \{NAME\_1: CHANGE\_NUM\_1, NAME\_2: CHANGE\_NUM\_2\}. Positive values mean increase and negative values mean decrease. For example, \{"wood": 2, "dirt": 3, "stone\_pickaxe": -1\}.

- health: an integer value in range 0 to 10 indicating the health level of the agent. 0 means death and 10 means full health.

- past\_agent\_positions: the history of location of agent, in the form of a list: [[x1, y1, z1, yaw1, pitch1], [x2, y2, z2, yaw2, pitch2], ...]. The yaw and pitch here are relative to the agent's initial forward direction, i.e., pitch = 0 and yaw = 0 is the front of the agent when it was born. The length of the list is the number of steps the agent has taken. The last element of the list is the current location of the agent.

- GLOBAL\_DATA: a global variable. It is initialized as a dictionary. You can save necessary information between different steps with it.

\#\# General game information

- The version of Minecraft is 1.11.

- Common block names : dirt, cobblestone, iron, diamond, wood, coal, water, air, lava, leaves, ... Collected item names are the same.

- FOV is set to -35 to 35 degrees for yaw and -30 to 30 degrees for pitch. The max visible distance of blocks is 64.

- Length of each block is 1 meter.

- The y coordinate of the agent is the agent's height. The larger the y, the higher the agent. The ground level is around y = 63, but is not fixed.

- At each step, the agent selects one operation in is action space. The action space includes doing nothing (i.e. staying still), moving forward and backward, jumping and attacking. The action space also includes turning left, right, up and down. 

- The attack range is 2 meters in front of the agent. The agent can move 0.2 meters in each step. The agent can turn 15 degrees in each step. Typically, the agent needs to stay still and attack for **60 successive steps** to break a block.

- The hunger value is always at the max level.

\#\# Reward function requirements

- You should write a dense reward function \textasciigrave dense\textasciigrave  and a sparse reward function \textasciigrave sparse\textasciigrave . The sparse reward indicates achieving the goal or receiving heavy punishment. The dense reward provides intermediate signal to guide the agent in the process of achieving the goal. The magnitude of the return value does not matter, but the sign (positive or negative) is important. The final reward will be \textasciigrave np.sign(sparse(...)) * 1 + np.sign(dense(...)) * 0.1\textasciigrave .

\#\# Output Requirements

- The reward function should be written in Python 3.9.

- Output the code block only. **Do not output anything else outside the code block**.

- You should include **sufficient comments** in your reward function to explain your thoughts, the objective and **implementation details**. The implementation can be specified to a specific line of code.

- If you need to import packages (e.g. math, numpy) or define helper functions, define them at the beginning of the function. Do not use unimported packages and undefined functions.

\#\# Output format

Strictly follow the following format. **Do not output anything else outside the code block**.

def reward\_function(current\_nearest\_blocks, previous\_nearest\_blocks, inventory\_change, health, past\_agent\_positions, GLOBAL\_DATA):
\\
\hbox{\ \ \ \ }\# Thoughts:\\
\hbox{\ \ \ \ }\# ...\\
\hbox{\ \ \ \ }\# (import packages and define helper functions)\\
\hbox{\ \ \ \ }import numpy as np\\
\hbox{\ \ \ \ }...\\
\hbox{\ \ \ \ }def dense(current\_nearest\_blocks, previous\_nearest\_blocks, inventory\_change, health, past\_agent\_positions, GLOBAL\_DATA):\\
\hbox{\ \ \ \ \ \ \ \ }...\\
\hbox{\ \ \ \ }def sparse(current\_nearest\_blocks, previous\_nearest\_blocks, inventory\_change, health, past\_agent\_positions, GLOBAL\_DATA):\\
\hbox{\ \ \ \ \ \ \ \ }...\\
\hbox{\ \ \ \ }dense\_reward = dense(current\_nearest\_blocks, previous\_nearest\_blocks, inventory\_change, health, past\_agent\_positions, GLOBAL\_DATA)\\
\hbox{\ \ \ \ }sparse\_reward = sparse(current\_nearest\_blocks, previous\_nearest\_blocks, inventory\_change, health, past\_agent\_positions, GLOBAL\_DATA)\\
\hbox{\ \ \ \ }return np.sign(sparse\_reward) * 1 + np.sign(dense\_reward) * 0.1\\
\hbox{\ \ \ \ }...\\
Now write a reward function. Then in each iteration, I will use the reward function to train an RL agent, and test it in the environment. I will give you possible reasons of the failure found during the testing, and you should modify the reward function accordingly.

--------------------------------------------------------------------------------------------------------------------------------------------

\texttt{\textbf{DESIGNER OUTPUT:}}

def reward\_function(current\_nearest\_blocks, previous\_nearest\_blocks, inventory\_change, health, past\_agent\_positions, GLOBAL\_DATA):
\\
\hbox{\ \ \ \ }"""\\
\hbox{\ \ \ \ }Thoughts:\\
\hbox{\ \ \ \ }In order to make the agent approach the diamond, we need to give positive rewards for actions that\\
\hbox{\ \ \ \ }make the agent get closer to the diamond and negative rewards for actions that make the agent get further from the diamond.\\
\hbox{\ \ \ \ }We also need to give a negative reward if the agent's health decreases to encourage the agent to avoid dangerous situations.\\
\hbox{\ \ \ \ }The dense reward function will be based on the distance to the diamond and health.\\
\hbox{\ \ \ \ }The sparse reward function will give a large positive reward when the agent reaches the diamond and a large negative reward when the agent dies.\\
\hbox{\ \ \ \ }"""\\
\hbox{\ \ \ \ }import numpy as np\\
\hbox{\ \ \ \ }def dense(current\_nearest\_blocks, previous\_nearest\_blocks, inventory\_change, health, past\_agent\_positions, GLOBAL\_DATA):\\
\hbox{\ \ \ \ \ \ \ \ }\# Initialize reward\\
\hbox{\ \ \ \ \ \ \ \ }reward = 0\\
\hbox{\ \ \ \ \ \ \ \ }\# If the agent is closer to diamond than before, give a positive reward\\
\hbox{\ \ \ \ \ \ \ \ }if "diamond" in current\_nearest\_blocks and "diamond" in previous\_nearest\_blocks:\\
\hbox{\ \ \ \ \ \ \ \ \ \ \ \ }if current\_nearest\_blocks["diamond"][0] < previous\_nearest\_blocks["diamond"][0]:\\
\hbox{\ \ \ \ \ \ \ \ \ \ \ \ \ \ \ \ }reward += 1\\
\hbox{\ \ \ \ \ \ \ \ \ \ \ \ }elif current\_nearest\_blocks["diamond"][0] > previous\_nearest\_blocks["diamond"][0]:\\
\hbox{\ \ \ \ \ \ \ \ \ \ \ \ \ \ \ \ }reward -= 1\\
\hbox{\ \ \ \ \ \ \ \ }\# If the agent's health decreases, give a negative reward\\
\hbox{\ \ \ \ \ \ \ \ }\hl{if health < GLOBAL\_DATA["health"]:}\\
\hbox{\ \ \ \ \ \ \ \ \ \ \ \ }reward -= 1\\
\hbox{\ \ \ \ \ \ \ \ }GLOBAL\_DATA["health"] = health\\
\hbox{\ \ \ \ \ \ \ \ }return reward\\
\hbox{\ \ \ \ }def sparse(current\_nearest\_blocks, previous\_nearest\_blocks, inventory\_change, health, past\_agent\_positions, GLOBAL\_DATA):\\
\hbox{\ \ \ \ \ \ \ \ }\# Initialize reward\\
\hbox{\ \ \ \ \ \ \ \ }reward = 0\\
\hbox{\ \ \ \ \ \ \ \ }\# If the agent reaches the diamond, give a large positive reward\\
\hbox{\ \ \ \ \ \ \ \ }if "diamond" in current\_nearest\_blocks and current\_nearest\_blocks["diamond"][0] < 1:\\
\hbox{\ \ \ \ \ \ \ \ \ \ \ \ }reward += 100\\
\hbox{\ \ \ \ \ \ \ \ }\# If the agent dies, give a large negative reward\\
\hbox{\ \ \ \ \ \ \ \ }if health == 0:\\
\hbox{\ \ \ \ \ \ \ \ \ \ \ \ }reward -= 100\\
\hbox{\ \ \ \ \ \ \ \ }return reward\\
\hbox{\ \ \ \ }dense\_reward = dense(current\_nearest\_blocks, previous\_nearest\_blocks, inventory\_change, health, past\_agent\_positions, GLOBAL\_DATA)\\
\hbox{\ \ \ \ }sparse\_reward = sparse(current\_nearest\_blocks, previous\_nearest\_blocks, inventory\_change, health, past\_agent\_positions, GLOBAL\_DATA)\\
\hbox{\ \ \ \ }return np.sign(dense\_reward) * 1 + np.sign(sparse\_reward) * 0.1\\
--------------------------------------------------------------------------------------------------------------------------------------------

\texttt{\textbf{CRITIC INPUT:}}

You are now a proficient Minecraft player. I will give you a reward function written in Python, which is used to train an RL agent to play Minecraft. 

You should help me to evaluate whether the reward function has met the requirements.

\#\# Task description

- Objective: Find and approach diamond with higher success rate and avoid death.

- Initial Status: 
\\
\hbox{\ \ }1. The agent is under ground at y level 11, the most common level to find diamonds.\\
\hbox{\ \ }2. The agent already has an iron pickaxe.\\
- Success criteria: The distance to the nearest diamond block is less than 2 meter. 

- Procedure: Explore horizontally to find a diamond, face it and approach it.

\#\# Input parameters of the reward function

- current\_nearest\_blocks: nearby block types and nearest distance of each block type at the **current step**. It is a dictionary, consisting of the names, relative distances with the agent, and relative angles (yaw and pitch) with the agent, in the form of: \{NAME\_1: (DIST\_1, YAW\_1, PITCH\_1), NAME\_2: (DIST\_2, YAW\_2, PITCH\_2)\}. For example, \{"wood": (24.7, 1.48, -1.57), "cobblestone": (1.3, -0.17, 1.05), "iron\_ore": (4.5, 0.61, -0.17)\}. If a certrain block type does not exist in the field of view, it is not present in the dictionary. The yaw and pitch here is relative to the agent's front, i.e., pitch = 0 and yaw = 0 is the front of the agent.

- previou\_nearest\_blocks: nearby block types and nearest distance of each block type at the **previous step**, with the same format as the current\_nearest\_blocks.

- inventory\_change: the change of the agent's inventory from the previous step to current step, in the form of a dictionary: \{NAME\_1: CHANGE\_NUM\_1, NAME\_2: CHANGE\_NUM\_2\}. Positive values mean increase and negative values mean decrease. For example, \{"wood": 2, "dirt": 3, "stone\_pickaxe": -1\}.

- health: an integer value in range 0 to 10 indicating the health level of the agent. 0 means death and 10 means full health.

- past\_agent\_positions: the history of location of agent, in the form of a list: [[x1, y1, z1, yaw1, pitch1], [x2, y2, z2, yaw2, pitch2], ...]. The yaw and pitch here are relative to the agent's initial forward direction, i.e., pitch = 0 and yaw = 0 is the front of the agent when it was born. The length of the list is the number of steps the agent has taken. The last element of the list is the current location of the agent.

- GLOBAL\_DATA: a global variable. It is initialized as a dictionary. You can save necessary information between different steps with it.

\#\# General game information

- The version of Minecraft is 1.11.

- Common block names : dirt, cobblestone, iron, diamond, wood, coal, water, air, lava, leaves, ... Collected item names are the same.

- FOV is set to -35 to 35 degrees for yaw and -30 to 30 degrees for pitch. The max visible distance of blocks is 64.

- Length of each block is 1 meter.

- The y coordinate of the agent is the agent's height. The larger the y, the higher the agent. The ground level is around y = 63, but is not fixed.

- At each step, the agent selects one operation in is action space. The action space includes doing nothing (i.e. staying still), moving forward and backward, jumping and attacking. The action space also includes turning left, right, up and down. 

- The attack range is 2 meters in front of the agent. The agent can move 0.2 meters in each step. The agent can turn 15 degrees in each step. Typically, the agent needs to stay still and attack for **60 successive steps** to break a block.

- The hunger value is always at the max level.

\#\# Reward function requirements

- You should write a dense reward function \textasciigrave dense\textasciigrave  and a sparse reward function \textasciigrave sparse\textasciigrave . The sparse reward indicates achieving the goal or receiving heavy punishment. The dense reward provides intermediate signal to guide the agent in the process of achieving the goal. The magnitude of the return value does not matter, but the sign (positive or negative) is important. The final reward will be \textasciigrave np.sign(sparse(...)) * 1 + np.sign(dense(...)) * 0.1\textasciigrave .

\#\# Requirements

1. You should check whether the reward function meets the **reward function requirements** above. 

2. Your judgement should consider whether the comments and the detailed implementation code are consistent. You can judge whether a line is correct by its comment, but do only rely on the comments.

3. You should also check the format of the reward function. It should be like:

def reward\_function(current\_nearest\_blocks, previous\_nearest\_blocks, inventory\_change, health, past\_agent\_positions, GLOBAL\_DATA):
\\
\hbox{\ \ \ \ }\# Thoughts:\\
\hbox{\ \ \ \ }\# ...\\
\hbox{\ \ \ \ }\# (import packages and define helper functions)\\
\hbox{\ \ \ \ }import numpy as np\\
\hbox{\ \ \ \ }...\\
\hbox{\ \ \ \ }def dense(current\_nearest\_blocks, previous\_nearest\_blocks, inventory\_change, health, past\_agent\_positions, GLOBAL\_DATA):\\
\hbox{\ \ \ \ \ \ \ \ }...\\
\hbox{\ \ \ \ }def sparse(current\_nearest\_blocks, previous\_nearest\_blocks, inventory\_change, health, past\_agent\_positions, GLOBAL\_DATA):\\
\hbox{\ \ \ \ \ \ \ \ }...\\
\hbox{\ \ \ \ }dense\_reward = dense(current\_nearest\_blocks, previous\_nearest\_blocks, inventory\_change, health, past\_agent\_positions, GLOBAL\_DATA)\\
\hbox{\ \ \ \ }sparse\_reward = sparse(current\_nearest\_blocks, previous\_nearest\_blocks, inventory\_change, health, past\_agent\_positions, GLOBAL\_DATA)\\
\hbox{\ \ \ \ }return np.sign(sparse\_reward) * 1 + np.sign(dense\_reward) * 0.1\\
\hbox{\ \ \ \ }...\\
4. Consider the **Information about the game state** above.

5. **You don't need to check whether the reward function meets the task description**.

6. You should first provide a reasoning and decision. If the reward function fails, provide a critique to help to improve the reward function.

\#\# Output format

You should only respond in JSON format as described below:

\{
\\
\hbox{\ \ \ \ }"reasoning": "reasoning",\\
\hbox{\ \ \ \ }"success": boolean,\\
\hbox{\ \ \ \ }"critique": critique,\\
\}

Ensure the response can be parsed by Python \textasciigrave json.loads\textasciigrave  , e.g. no trailing commas, no single quotes, etc. Do not output anything else like explanations.

\#\# Reward function to be evaluated

def reward\_function(current\_nearest\_blocks, previous\_nearest\_blocks, inventory\_change, health, past\_agent\_positions, GLOBAL\_DATA):
\\
\hbox{\ \ \ \ }"""\\
\hbox{\ \ \ \ }Thoughts:\\
\hbox{\ \ \ \ }In order to make the agent approach the diamond, we need to give positive rewards for actions that\\
\hbox{\ \ \ \ }make the agent get closer to the diamond and negative rewards for actions that make the agent get further from the diamond.\\
\hbox{\ \ \ \ }We also need to give a negative reward if the agent's health decreases to encourage the agent to avoid dangerous situations.\\
\hbox{\ \ \ \ }The dense reward function will be based on the distance to the diamond and health.\\
\hbox{\ \ \ \ }The sparse reward function will give a large positive reward when the agent reaches the diamond and a large negative reward when the agent dies.\\
\hbox{\ \ \ \ }"""\\
\hbox{\ \ \ \ }import numpy as np\\
\hbox{\ \ \ \ }def dense(current\_nearest\_blocks, previous\_nearest\_blocks, inventory\_change, health, past\_agent\_positions, GLOBAL\_DATA):\\
\hbox{\ \ \ \ \ \ \ \ }\# Initialize reward\\
\hbox{\ \ \ \ \ \ \ \ }reward = 0\\
\hbox{\ \ \ \ \ \ \ \ }\# If the agent is closer to diamond than before, give a positive reward\\
\hbox{\ \ \ \ \ \ \ \ }if "diamond" in current\_nearest\_blocks and "diamond" in previous\_nearest\_blocks:\\
\hbox{\ \ \ \ \ \ \ \ \ \ \ \ }if current\_nearest\_blocks["diamond"][0] < previous\_nearest\_blocks["diamond"][0]:\\
\hbox{\ \ \ \ \ \ \ \ \ \ \ \ \ \ \ \ }reward += 1\\
\hbox{\ \ \ \ \ \ \ \ \ \ \ \ }elif current\_nearest\_blocks["diamond"][0] > previous\_nearest\_blocks["diamond"][0]:\\
\hbox{\ \ \ \ \ \ \ \ \ \ \ \ \ \ \ \ }reward -= 1\\
\hbox{\ \ \ \ \ \ \ \ }\# If the agent's health decreases, give a negative reward\\
\hbox{\ \ \ \ \ \ \ \ }if health < GLOBAL\_DATA["health"]:\\
\hbox{\ \ \ \ \ \ \ \ \ \ \ \ }reward -= 1\\
\hbox{\ \ \ \ \ \ \ \ }GLOBAL\_DATA["health"] = health\\
\hbox{\ \ \ \ \ \ \ \ }return reward\\
\hbox{\ \ \ \ }def sparse(current\_nearest\_blocks, previous\_nearest\_blocks, inventory\_change, health, past\_agent\_positions, GLOBAL\_DATA):\\
\hbox{\ \ \ \ \ \ \ \ }\# Initialize reward\\
\hbox{\ \ \ \ \ \ \ \ }reward = 0\\
\hbox{\ \ \ \ \ \ \ \ }\# If the agent reaches the diamond, give a large positive reward\\
\hbox{\ \ \ \ \ \ \ \ }if "diamond" in current\_nearest\_blocks and current\_nearest\_blocks["diamond"][0] < 1:\\
\hbox{\ \ \ \ \ \ \ \ \ \ \ \ }reward += 100\\
\hbox{\ \ \ \ \ \ \ \ }\# If the agent dies, give a large negative reward\\
\hbox{\ \ \ \ \ \ \ \ }if health == 0:\\
\hbox{\ \ \ \ \ \ \ \ \ \ \ \ }reward -= 100\\
\hbox{\ \ \ \ \ \ \ \ }return reward\\
\hbox{\ \ \ \ }dense\_reward = dense(current\_nearest\_blocks, previous\_nearest\_blocks, inventory\_change, health, past\_agent\_positions, GLOBAL\_DATA)\\
\hbox{\ \ \ \ }sparse\_reward = sparse(current\_nearest\_blocks, previous\_nearest\_blocks, inventory\_change, health, past\_agent\_positions, GLOBAL\_DATA)\\
\hbox{\ \ \ \ }return np.sign(dense\_reward) * 1 + np.sign(sparse\_reward) * 0.1\\
--------------------------------------------------------------------------------------------------------------------------------------------

\texttt{\textbf{CRITIC OUTPUT:}}

\{
\\
\hbox{\ \ \ \ }"reasoning": "The reward function does not consider that \hl{\textasciigrave health\textasciigrave  is not initialized in GLOBAL\_DATA}. It will raise a error because the key \textasciigrave health\textasciigrave  is not found.",\\
\hbox{\ \ \ \ }\hl{"success": false,}\\
\hbox{\ \ \ \ }"critique": "The reward function needs to be revised to initialize \textasciigrave health\textasciigrave  in GLOBAL\_DATA."\\
\}

--------------------------------------------------------------------------------------------------------------------------------------------

\texttt{\textbf{DESIGNER INPUT:}}

The reward function fails to pass the verification. The reasoning is: 

\textasciigrave \textasciigrave \textasciigrave 

The reward function does not consider that \textasciigrave health\textasciigrave  is not initialized in GLOBAL\_DATA. It will raise a error because the key \textasciigrave health\textasciigrave  is not found.

\textasciigrave \textasciigrave \textasciigrave 

The critique is:

\textasciigrave \textasciigrave \textasciigrave 

The reward function needs to be revised to initialize \textasciigrave health\textasciigrave  in GLOBAL\_DATA.

\textasciigrave \textasciigrave \textasciigrave

\#\# Requirements

Please consider the reasoning and critique, and modify the reward function. 

1. If necessary, you can write a totally different reward function than the current one.

2. In the first part of the reward function, you should provide your thoughts of modifying the reward function. The thoughts should not directly copy the given reasoning or critique. **The thoughts should be concise.**

\#\# Reward function requirements

- You should write a dense reward function \textasciigrave dense\textasciigrave  and a sparse reward function \textasciigrave sparse\textasciigrave . The sparse reward indicates achieving the goal or receiving heavy punishment. The dense reward provides intermediate signal to guide the agent in the process of achieving the goal. The magnitude of the return value does not matter, but the sign (positive or negative) is important. The final reward will be \textasciigrave np.sign(sparse(...)) * 1 + np.sign(dense(...)) * 0.1\textasciigrave .

\#\# Output format

Strictly follow the following format. **Do not output anything else outside the code block**.

def reward\_function(current\_nearest\_blocks, previous\_nearest\_blocks, inventory\_change, health, past\_agent\_positions, GLOBAL\_DATA):
\\
\hbox{\ \ \ \ }\# Thoughts:\\
\hbox{\ \ \ \ }\# ...\\
\hbox{\ \ \ \ }\# (import packages and define helper functions)\\
\hbox{\ \ \ \ }import numpy as np\\
\hbox{\ \ \ \ }...\\
\hbox{\ \ \ \ }def dense(current\_nearest\_blocks, previous\_nearest\_blocks, inventory\_change, health, past\_agent\_positions, GLOBAL\_DATA):\\
\hbox{\ \ \ \ \ \ \ \ }...\\
\hbox{\ \ \ \ }def sparse(current\_nearest\_blocks, previous\_nearest\_blocks, inventory\_change, health, past\_agent\_positions, GLOBAL\_DATA):\\
\hbox{\ \ \ \ \ \ \ \ }...\\
\hbox{\ \ \ \ }dense\_reward = dense(current\_nearest\_blocks, previous\_nearest\_blocks, inventory\_change, health, past\_agent\_positions, GLOBAL\_DATA)\\
\hbox{\ \ \ \ }sparse\_reward = sparse(current\_nearest\_blocks, previous\_nearest\_blocks, inventory\_change, health, past\_agent\_positions, GLOBAL\_DATA)\\
\hbox{\ \ \ \ }return np.sign(sparse\_reward) * 1 + np.sign(dense\_reward) * 0.1\\
\hbox{\ \ \ \ }...\\
--------------------------------------------------------------------------------------------------------------------------------------------

\texttt{\textbf{DESIGNER OUTPUT:}}

def reward\_function(current\_nearest\_blocks, previous\_nearest\_blocks, inventory\_change, health, past\_agent\_positions, GLOBAL\_DATA):
\\
\hbox{\ \ \ \ }"""\\
\hbox{\ \ \ \ }Thoughts:\\
\hbox{\ \ \ \ }In order to make the agent approach the diamond, we need to give positive rewards for actions that\\
\hbox{\ \ \ \ }make the agent get closer to the diamond and negative rewards for actions that make the agent get further from the diamond.\\
\hbox{\ \ \ \ }We also need to give a negative reward if the agent's health decreases to encourage the agent to avoid dangerous situations.\\
\hbox{\ \ \ \ }The dense reward function will be based on the distance to the diamond and health.\\
\hbox{\ \ \ \ }The sparse reward function will give a large positive reward when the agent reaches the diamond and a large negative reward when the agent dies.\\
\hbox{\ \ \ \ }"""\\
\hbox{\ \ \ \ }import numpy as np\\
\hbox{\ \ \ \ }def dense(current\_nearest\_blocks, previous\_nearest\_blocks, inventory\_change, health, past\_agent\_positions, GLOBAL\_DATA):\\
\hbox{\ \ \ \ \ \ \ \ }\# Initialize reward\\
\hbox{\ \ \ \ \ \ \ \ }reward = 0\\
\hbox{\ \ \ \ \ \ \ \ }\# If the agent is closer to diamond than before, give a positive reward\\
\hbox{\ \ \ \ \ \ \ \ }if "diamond" in current\_nearest\_blocks and "diamond" in previous\_nearest\_blocks:\\
\hbox{\ \ \ \ \ \ \ \ \ \ \ \ }if current\_nearest\_blocks["diamond"][0] < previous\_nearest\_blocks["diamond"][0]:\\
\hbox{\ \ \ \ \ \ \ \ \ \ \ \ \ \ \ \ }reward += 1\\
\hbox{\ \ \ \ \ \ \ \ \ \ \ \ }elif current\_nearest\_blocks["diamond"][0] > previous\_nearest\_blocks["diamond"][0]:\\
\hbox{\ \ \ \ \ \ \ \ \ \ \ \ \ \ \ \ }reward -= 1\\
\hbox{\ \ \ \ \ \ \ \ }\# If the agent's health decreases, give a negative reward\\
\hbox{\ \ \ \ \ \ \ \ }\hl{if "health" in GLOBAL\_DATA:}\\
\hbox{\ \ \ \ \ \ \ \ \ \ \ \ }if health < GLOBAL\_DATA["health"]:\\
\hbox{\ \ \ \ \ \ \ \ \ \ \ \ \ \ \ \ }reward -= 1\\
\hbox{\ \ \ \ \ \ \ \ }GLOBAL\_DATA["health"] = health\\
\hbox{\ \ \ \ \ \ \ \ }return reward\\
\hbox{\ \ \ \ }def sparse(current\_nearest\_blocks, previous\_nearest\_blocks, inventory\_change, health, past\_agent\_positions, GLOBAL\_DATA):\\
\hbox{\ \ \ \ \ \ \ \ }\# Initialize reward\\
\hbox{\ \ \ \ \ \ \ \ }reward = 0\\
\hbox{\ \ \ \ \ \ \ \ }\# If the agent reaches the diamond, give a large positive reward\\
\hbox{\ \ \ \ \ \ \ \ }if "diamond" in current\_nearest\_blocks and current\_nearest\_blocks["diamond"][0] < 1:\\
\hbox{\ \ \ \ \ \ \ \ \ \ \ \ }reward += 100\\
\hbox{\ \ \ \ \ \ \ \ }\# If the agent dies, give a large negative reward\\
\hbox{\ \ \ \ \ \ \ \ }if health == 0:\\
\hbox{\ \ \ \ \ \ \ \ \ \ \ \ }reward -= 100\\
\hbox{\ \ \ \ \ \ \ \ }return reward\\
\hbox{\ \ \ \ }dense\_reward = dense(current\_nearest\_blocks, previous\_nearest\_blocks, inventory\_change, health, past\_agent\_positions, GLOBAL\_DATA)\\
\hbox{\ \ \ \ }sparse\_reward = sparse(current\_nearest\_blocks, previous\_nearest\_blocks, inventory\_change, health, past\_agent\_positions, GLOBAL\_DATA)\\
\hbox{\ \ \ \ }return np.sign(dense\_reward) * 1 + np.sign(sparse\_reward) * 0.1\\
--------------------------------------------------------------------------------------------------------------------------------------------

\texttt{\textbf{CRITIC INPUT:}}

You are now a proficient Minecraft player. I will give you a reward function written in Python, which is used to train an RL agent to play Minecraft. 

You should help me to evaluate whether the reward function has met the requirements.

\#\# Task description

- Objective: Find and approach diamond with higher success rate and avoid death.

- Initial Status: 
\\
\hbox{\ \ }1. The agent is under ground at y level 11, the most common level to find diamonds.\\
\hbox{\ \ }2. The agent already has an iron pickaxe.\\
- Success criteria: The distance to the nearest diamond block is less than 2 meter. 

- Procedure: Explore horizontally to find a diamond, face it and approach it.

\#\# Input parameters of the reward function

- current\_nearest\_blocks: nearby block types and nearest distance of each block type at the **current step**. It is a dictionary, consisting of the names, relative distances with the agent, and relative angles (yaw and pitch) with the agent, in the form of: \{NAME\_1: (DIST\_1, YAW\_1, PITCH\_1), NAME\_2: (DIST\_2, YAW\_2, PITCH\_2)\}. For example, \{"wood": (24.7, 1.48, -1.57), "cobblestone": (1.3, -0.17, 1.05), "iron\_ore": (4.5, 0.61, -0.17)\}. If a certrain block type does not exist in the field of view, it is not present in the dictionary. The yaw and pitch here is relative to the agent's front, i.e., pitch = 0 and yaw = 0 is the front of the agent.

- previou\_nearest\_blocks: nearby block types and nearest distance of each block type at the **previous step**, with the same format as the current\_nearest\_blocks.

- inventory\_change: the change of the agent's inventory from the previous step to current step, in the form of a dictionary: \{NAME\_1: CHANGE\_NUM\_1, NAME\_2: CHANGE\_NUM\_2\}. Positive values mean increase and negative values mean decrease. For example, \{"wood": 2, "dirt": 3, "stone\_pickaxe": -1\}.

- health: an integer value in range 0 to 10 indicating the health level of the agent. 0 means death and 10 means full health.

- past\_agent\_positions: the history of location of agent, in the form of a list: [[x1, y1, z1, yaw1, pitch1], [x2, y2, z2, yaw2, pitch2], ...]. The yaw and pitch here are relative to the agent's initial forward direction, i.e., pitch = 0 and yaw = 0 is the front of the agent when it was born. The length of the list is the number of steps the agent has taken. The last element of the list is the current location of the agent.

- GLOBAL\_DATA: a global variable. It is initialized as a dictionary. You can save necessary information between different steps with it.

\#\# General game information

- The version of Minecraft is 1.11.

- Common block names : dirt, cobblestone, iron, diamond, wood, coal, water, air, lava, leaves, ... Collected item names are the same.

- FOV is set to -35 to 35 degrees for yaw and -30 to 30 degrees for pitch. The max visible distance of blocks is 64.

- Length of each block is 1 meter.

- The y coordinate of the agent is the agent's height. The larger the y, the higher the agent. The ground level is around y = 63, but is not fixed.

- At each step, the agent selects one operation in is action space. The action space includes doing nothing (i.e. staying still), moving forward and backward, jumping and attacking. The action space also includes turning left, right, up and down. 

- The attack range is 2 meters in front of the agent. The agent can move 0.2 meters in each step. The agent can turn 15 degrees in each step. Typically, the agent needs to stay still and attack for **60 successive steps** to break a block.

- The hunger value is always at the max level.

\#\# Reward function requirements

- You should write a dense reward function \textasciigrave dense\textasciigrave  and a sparse reward function \textasciigrave sparse\textasciigrave . The sparse reward indicates achieving the goal or receiving heavy punishment. The dense reward provides intermediate signal to guide the agent in the process of achieving the goal. The magnitude of the return value does not matter, but the sign (positive or negative) is important. The final reward will be \textasciigrave np.sign(sparse(...)) * 1 + np.sign(dense(...)) * 0.1\textasciigrave .

\#\# Requirements

1. You should check whether the reward function meets the **reward function requirements** above. 

2. Your judgement should consider whether the comments and the detailed implementation code are consistent. You can judge whether a line is correct by its comment, but do only rely on the comments.

3. You should also check the format of the reward function. It should be like:

def reward\_function(current\_nearest\_blocks, previous\_nearest\_blocks, inventory\_change, health, past\_agent\_positions, GLOBAL\_DATA):
\\
\hbox{\ \ \ \ }\# Thoughts:\\
\hbox{\ \ \ \ }\# ...\\
\hbox{\ \ \ \ }\# (import packages and define helper functions)\\
\hbox{\ \ \ \ }import numpy as np\\
\hbox{\ \ \ \ }...\\
\hbox{\ \ \ \ }def dense(current\_nearest\_blocks, previous\_nearest\_blocks, inventory\_change, health, past\_agent\_positions, GLOBAL\_DATA):\\
\hbox{\ \ \ \ \ \ \ \ }...\\
\hbox{\ \ \ \ }def sparse(current\_nearest\_blocks, previous\_nearest\_blocks, inventory\_change, health, past\_agent\_positions, GLOBAL\_DATA):\\
\hbox{\ \ \ \ \ \ \ \ }...\\
\hbox{\ \ \ \ }dense\_reward = dense(current\_nearest\_blocks, previous\_nearest\_blocks, inventory\_change, health, past\_agent\_positions, GLOBAL\_DATA)\\
\hbox{\ \ \ \ }sparse\_reward = sparse(current\_nearest\_blocks, previous\_nearest\_blocks, inventory\_change, health, past\_agent\_positions, GLOBAL\_DATA)\\
\hbox{\ \ \ \ }return np.sign(sparse\_reward) * 1 + np.sign(dense\_reward) * 0.1\\
\hbox{\ \ \ \ }...\\
4. Consider the **Information about the game state** above.

5. **You don't need to check whether the reward function meets the task description**.

6. You should first provide a reasoning and decision. If the reward function fails, provide a critique to help to improve the reward function.

\#\# Output format

You should only respond in JSON format as described below:

\{
\\
\hbox{\ \ \ \ }"reasoning": "reasoning",\\
\hbox{\ \ \ \ }"success": boolean,\\
\hbox{\ \ \ \ }"critique": critique,\\
\}

Ensure the response can be parsed by Python \textasciigrave json.loads\textasciigrave  , e.g. no trailing commas, no single quotes, etc. Do not output anything else like explanations.

\#\# Reward function to be evaluated

def reward\_function(current\_nearest\_blocks, previous\_nearest\_blocks, inventory\_change, health, past\_agent\_positions, GLOBAL\_DATA):
\\
\hbox{\ \ \ \ }"""\\
\hbox{\ \ \ \ }Thoughts:\\
\hbox{\ \ \ \ }In order to make the agent approach the diamond, we need to give positive rewards for actions that\\
\hbox{\ \ \ \ }make the agent get closer to the diamond and negative rewards for actions that make the agent get further from the diamond.\\
\hbox{\ \ \ \ }We also need to give a negative reward if the agent's health decreases to encourage the agent to avoid dangerous situations.\\
\hbox{\ \ \ \ }The dense reward function will be based on the distance to the diamond and health.\\
\hbox{\ \ \ \ }The sparse reward function will give a large positive reward when the agent reaches the diamond and a large negative reward when the agent dies.\\
\hbox{\ \ \ \ }"""\\
\hbox{\ \ \ \ }import numpy as np\\
\hbox{\ \ \ \ }def dense(current\_nearest\_blocks, previous\_nearest\_blocks, inventory\_change, health, past\_agent\_positions, GLOBAL\_DATA):\\
\hbox{\ \ \ \ \ \ \ \ }\# Initialize reward\\
\hbox{\ \ \ \ \ \ \ \ }reward = 0\\
\hbox{\ \ \ \ \ \ \ \ }\# If the agent is closer to diamond than before, give a positive reward\\
\hbox{\ \ \ \ \ \ \ \ }if "diamond" in current\_nearest\_blocks and "diamond" in previous\_nearest\_blocks:\\
\hbox{\ \ \ \ \ \ \ \ \ \ \ \ }if current\_nearest\_blocks["diamond"][0] < previous\_nearest\_blocks["diamond"][0]:\\
\hbox{\ \ \ \ \ \ \ \ \ \ \ \ \ \ \ \ }reward += 1\\
\hbox{\ \ \ \ \ \ \ \ \ \ \ \ }elif current\_nearest\_blocks["diamond"][0] > previous\_nearest\_blocks["diamond"][0]:\\
\hbox{\ \ \ \ \ \ \ \ \ \ \ \ \ \ \ \ }reward -= 1\\
\hbox{\ \ \ \ \ \ \ \ }\# If the agent's health decreases, give a negative reward\\
\hbox{\ \ \ \ \ \ \ \ }if "health" in GLOBAL\_DATA:\\
\hbox{\ \ \ \ \ \ \ \ \ \ \ \ }if health < GLOBAL\_DATA["health"]:\\
\hbox{\ \ \ \ \ \ \ \ \ \ \ \ \ \ \ \ }reward -= 1\\
\hbox{\ \ \ \ \ \ \ \ }GLOBAL\_DATA["health"] = health\\
\hbox{\ \ \ \ \ \ \ \ }return reward\\
\hbox{\ \ \ \ }def sparse(current\_nearest\_blocks, previous\_nearest\_blocks, inventory\_change, health, past\_agent\_positions, GLOBAL\_DATA):\\
\hbox{\ \ \ \ \ \ \ \ }\# Initialize reward\\
\hbox{\ \ \ \ \ \ \ \ }reward = 0\\
\hbox{\ \ \ \ \ \ \ \ }\# If the agent reaches the diamond, give a large positive reward\\
\hbox{\ \ \ \ \ \ \ \ }if "diamond" in current\_nearest\_blocks and current\_nearest\_blocks["diamond"][0] < 1:\\
\hbox{\ \ \ \ \ \ \ \ \ \ \ \ }reward += 100\\
\hbox{\ \ \ \ \ \ \ \ }\# If the agent dies, give a large negative reward\\
\hbox{\ \ \ \ \ \ \ \ }if health == 0:\\
\hbox{\ \ \ \ \ \ \ \ \ \ \ \ }reward -= 100\\
\hbox{\ \ \ \ \ \ \ \ }return reward\\
\hbox{\ \ \ \ }dense\_reward = dense(current\_nearest\_blocks, previous\_nearest\_blocks, inventory\_change, health, past\_agent\_positions, GLOBAL\_DATA)\\
\hbox{\ \ \ \ }sparse\_reward = sparse(current\_nearest\_blocks, previous\_nearest\_blocks, inventory\_change, health, past\_agent\_positions, GLOBAL\_DATA)\\
\hbox{\ \ \ \ }return np.sign(dense\_reward) * 1 + np.sign(sparse\_reward) * 0.1\\
--------------------------------------------------------------------------------------------------------------------------------------------

\texttt{\textbf{CRITIC OUTPUT:}}

\{
\\
\hbox{\ \ \ \ }"reasoning": "The reward function meets the requirements. It considers sparse reward of completing the objective and death, and dense reward of getting closer to diamond and health decrease.",\\
\hbox{\ \ \ \ }\hl{"success": true,}\\
\hbox{\ \ \ \ }"critique": null\\
\}

--------------------------------------------------------------------------------------------------------------------------------------------

(TRAIN AN AGENT WITH THE REWARD FUNCTION)

--------------------------------------------------------------------------------------------------------------------------------------------

\texttt{\textbf{ANALYZER INPUT:}}

You are now a proficient Minecraft player. I have trained an RL agent and tested for several times in the Minecraft game environment. 

I will give you the information of the failed test results, i.e. trajectories of actions, rewards and observations. You should help me write proper analysis of possible reasons of failure and inefficiency, and your suggestion on the game strategy.

\#\# Task description

- Objective: Find and approach diamond with higher success rate and avoid death.

- Initial Status: 
\\
\hbox{\ \ }1. The agent is under ground at y level 11, the most common level to find diamonds.\\
\hbox{\ \ }2. The agent already has an iron pickaxe.\\
- Success criteria: The distance to the nearest diamond block is less than 2 meter. 

- Procedure: Explore horizontally to find a diamond, face it and approach it.

\#\# General game information

- The version of Minecraft is 1.11.

- Common block names : dirt, cobblestone, iron, diamond, wood, coal, water, air, lava, leaves, ... Collected item names are the same.

- FOV is set to -35 to 35 degrees for yaw and -30 to 30 degrees for pitch. The max visible distance of blocks is 64.

- Length of each block is 1 meter.

- The y coordinate of the agent is the agent's height. The larger the y, the higher the agent. The ground level is around y = 63, but is not fixed.

- At each step, the agent selects one operation in is action space. The action space includes doing nothing (i.e. staying still), moving forward and backward, jumping and attacking. The action space also includes turning left, right, up and down. 

- The attack range is 2 meters in front of the agent. The agent can move 0.2 meters in each step. The agent can turn 15 degrees in each step. Typically, the agent needs to stay still and attack for **60 successive steps** to break a block.

- The hunger value is always at the max level.

\#\# Input format

\#\#\# Failed trajectories

The failed trajectories are shown as a list, where each entry is a dictionary representing a trajectory. Note that **only the failed cases are provided.**

The each item of "history" is a list representing statistics of each single step.  When the history is too long, it will be truncated to the last 32 steps, indicated by the value of "truncated".

"dead" indicates whether the agent has died.

The format is:

[
\\
\hbox{\ \ \ \ }\{\\
\hbox{\ \ \ \ \ \ \ \ }"history": \{\\
\hbox{\ \ \ \ \ \ \ \ \ \ \ \ }"rewards": [reward1, ...],\\
\hbox{\ \ \ \ \ \ \ \ \ \ \ \ }"actions": [action1, action2, ...],\\
\hbox{\ \ \ \ \ \ \ \ \ \ \ \ }"locations": [[x1, y1, z1, pitch1, yaw1], ...],\\
\hbox{\ \ \ \ \ \ \ \ \ \ \ \ }"inventory\_change": \{item1: delta\_num1, ...\},\\
\hbox{\ \ \ \ \ \ \ \ \ \ \ \ }"truncated": True or False,\\
\hbox{\ \ \ \ \ \ \ \ }\},\\
\hbox{\ \ \ \ \ \ \ \ }"final\_health": health,\\
\hbox{\ \ \ \ \ \ \ \ }"final\_inventory": \{item1: num1, ...\},\\
\hbox{\ \ \ \ \ \ \ \ }"final\_nearest\_blocks": \{block\_name1: nearest\_distance1, block\_name2: nearest\_distance2, ...\},\\
\hbox{\ \ \ \ \ \ \ \ }"block\_under\_foot": block\_name,\\
\hbox{\ \ \ \ \ \ \ \ }"dead": True or False,\\
\hbox{\ \ \ \ }\},\\
\hbox{\ \ \ \ }...\\
]

where "final\_health" is in range 0 and 10. "block\_under\_foot" is the block right under the agent.

\#\#\# Statistics

The statistics is a dictionary of all the test results, including **both successful and failed trajectories**.

The format is:

\{
\\
\hbox{\ \ \ \ }"success\_rate" : success\_rate,\\
\}

\#\# Requirements

Please write proper analysis of possible reasons of failure and inefficiency, and your suggestion on the game strategy. You should not not be limited to the task description above, but also come up with other failure cases based on the test results.

**The analysis and suggestion should be concise.**

\#\# Test results

\#\#\# Failed trajectories

[\{'block\_under\_foot': 'lava',
\\
\hbox{\ \ }'dead': True,\\
\hbox{\ \ }'final\_health': 0.5,\\
\hbox{\ \ }'final\_inventory': \{'coal': 3,\\
\hbox{\ \ \ \ \ \ \ \ \ \ \ \ \ \ \ \ \ \ \ \ \ \ }'cobblestone': 84,\\
\hbox{\ \ \ \ \ \ \ \ \ \ \ \ \ \ \ \ \ \ \ \ \ \ }'crafting\_table': 1,\\
\hbox{\ \ \ \ \ \ \ \ \ \ \ \ \ \ \ \ \ \ \ \ \ \ }'furnace': 1,\\
\hbox{\ \ \ \ \ \ \ \ \ \ \ \ \ \ \ \ \ \ \ \ \ \ }'gravel': 3,\\
\hbox{\ \ \ \ \ \ \ \ \ \ \ \ \ \ \ \ \ \ \ \ \ \ }'iron\_pickaxe': 1,\\
\hbox{\ \ \ \ \ \ \ \ \ \ \ \ \ \ \ \ \ \ \ \ \ \ }'planks': 2,\\
\hbox{\ \ \ \ \ \ \ \ \ \ \ \ \ \ \ \ \ \ \ \ \ \ }'stone': 17,\\
\hbox{\ \ \ \ \ \ \ \ \ \ \ \ \ \ \ \ \ \ \ \ \ \ }'stone\_pickaxe': 10,\\
\hbox{\ \ \ \ \ \ \ \ \ \ \ \ \ \ \ \ \ \ \ \ \ \ }'wooden\_pickaxe': 1\},\\
\hbox{\ \ }'final\_nearest\_blocks': \hl{\{'lava': [0.01, -0.87, -0.63]\}},\\
\hbox{\ \ }'history': \{'actions': ['noop', 'attack', 'noop', 'attack', 'noop', 'attack',\\
\hbox{\ \ \ \ \ \ \ \ \ \ \ \ \ \ \ \ \ \ \ \ \ \ \ \ \ \ }'attack', 'turn\_down', 'attack', 'attack', 'noop',\\
\hbox{\ \ \ \ \ \ \ \ \ \ \ \ \ \ \ \ \ \ \ \ \ \ \ \ \ \ }'attack', 'forward', 'forward', 'noop', 'noop',\\
\hbox{\ \ \ \ \ \ \ \ \ \ \ \ \ \ \ \ \ \ \ \ \ \ \ \ \ \ }'forward\_jump', 'forward\_jump', 'attack', 'attack',\\
\hbox{\ \ \ \ \ \ \ \ \ \ \ \ \ \ \ \ \ \ \ \ \ \ \ \ \ \ }'attack', 'noop', 'jump', 'noop', 'noop', 'jump',\\
\hbox{\ \ \ \ \ \ \ \ \ \ \ \ \ \ \ \ \ \ \ \ \ \ \ \ \ \ }'attack', 'turn\_down', 'turn\_up', 'forward\_jump',\\
\hbox{\ \ \ \ \ \ \ \ \ \ \ \ \ \ \ \ \ \ \ \ \ \ \ \ \ \ }'turn\_down', 'turn\_up'],\\
\hbox{\ \ \ \ \ \ \ \ \ \ \ \ \ \ }'inventory\_change': [\{\}, \{\}, \{\}, \{\}, \{\}, \{\}, \{\}, \{\}, \{\}, \{\}, \{\},\\
\hbox{\ \ \ \ \ \ \ \ \ \ \ \ \ \ \ \ \ \ \ \ \ \ \ \ \ \ \ \ \ \ \ \ \ \ \ }\{\}, \{\}, \{\}, \{\}, \{\}, \{\}, \{\}, \{\}, \{\}, \{\}, \{\},\\
\hbox{\ \ \ \ \ \ \ \ \ \ \ \ \ \ \ \ \ \ \ \ \ \ \ \ \ \ \ \ \ \ \ \ \ \ \ }\{\}, \{\}, \{\}, \{\}, \{\}, \{\}, \{\}, \{\}, \{\}, \{\}],\\
\hbox{\ \ \ \ \ \ \ \ \ \ \ \ \ \ }'locations': [[-205.42, 9.27, 474.4, 15.0, -90.0],\\
\hbox{\ \ \ \ \ \ \ \ \ \ \ \ \ \ \ \ \ \ \ \ \ \ \ \ \ \ \ \ }[-205.41, 9.24, 474.4, 15.0, -90.0],\\
\hbox{\ \ \ \ \ \ \ \ \ \ \ \ \ \ \ \ \ \ \ \ \ \ \ \ \ \ \ \ }[-205.41, 9.2, 474.4, 15.0, -90.0],\\
\hbox{\ \ \ \ \ \ \ \ \ \ \ \ \ \ \ \ \ \ \ \ \ \ \ \ \ \ \ \ }[-205.41, 9.16, 474.4, 15.0, -90.0],\\
\hbox{\ \ \ \ \ \ \ \ \ \ \ \ \ \ \ \ \ \ \ \ \ \ \ \ \ \ \ \ }[-205.41, 9.12, 474.4, 15.0, -90.0],\\
\hbox{\ \ \ \ \ \ \ \ \ \ \ \ \ \ \ \ \ \ \ \ \ \ \ \ \ \ \ \ }[-205.41, 9.08, 474.4, 15.0, -90.0],\\
\hbox{\ \ \ \ \ \ \ \ \ \ \ \ \ \ \ \ \ \ \ \ \ \ \ \ \ \ \ \ }[-205.41, 9.04, 474.4, 15.0, -90.0],\\
\hbox{\ \ \ \ \ \ \ \ \ \ \ \ \ \ \ \ \ \ \ \ \ \ \ \ \ \ \ \ }[-205.41, 9.0, 474.4, 15.0, -90.0],\\
\hbox{\ \ \ \ \ \ \ \ \ \ \ \ \ \ \ \ \ \ \ \ \ \ \ \ \ \ \ \ }[-205.41, 8.96, 474.4, 30.0, -90.0],\\
\hbox{\ \ \ \ \ \ \ \ \ \ \ \ \ \ \ \ \ \ \ \ \ \ \ \ \ \ \ \ }[-205.41, 8.92, 474.4, 30.0, -90.0],\\
\hbox{\ \ \ \ \ \ \ \ \ \ \ \ \ \ \ \ \ \ \ \ \ \ \ \ \ \ \ \ }[-205.41, 8.88, 474.4, 30.0, -90.0],\\
\hbox{\ \ \ \ \ \ \ \ \ \ \ \ \ \ \ \ \ \ \ \ \ \ \ \ \ \ \ \ }[-205.41, 8.84, 474.4, 30.0, -90.0],\\
\hbox{\ \ \ \ \ \ \ \ \ \ \ \ \ \ \ \ \ \ \ \ \ \ \ \ \ \ \ \ }[-205.41, 8.8, 474.4, 30.0, -90.0],\\
\hbox{\ \ \ \ \ \ \ \ \ \ \ \ \ \ \ \ \ \ \ \ \ \ \ \ \ \ \ \ }[-205.39, 8.76, 474.4, 30.0, -90.0],\\
\hbox{\ \ \ \ \ \ \ \ \ \ \ \ \ \ \ \ \ \ \ \ \ \ \ \ \ \ \ \ }[-205.36, 8.72, 474.4, 30.0, -90.0],\\
\hbox{\ \ \ \ \ \ \ \ \ \ \ \ \ \ \ \ \ \ \ \ \ \ \ \ \ \ \ \ }[-205.34, 8.68, 474.4, 30.0, -90.0],\\
\hbox{\ \ \ \ \ \ \ \ \ \ \ \ \ \ \ \ \ \ \ \ \ \ \ \ \ \ \ \ }[-205.34, 8.64, 474.4, 30.0, -90.0],\\
\hbox{\ \ \ \ \ \ \ \ \ \ \ \ \ \ \ \ \ \ \ \ \ \ \ \ \ \ \ \ }[-205.31, 8.64, 474.4, 30.0, -90.0],\\
\hbox{\ \ \ \ \ \ \ \ \ \ \ \ \ \ \ \ \ \ \ \ \ \ \ \ \ \ \ \ }[-205.3, 8.66, 474.4, 30.0, -90.0],\\
\hbox{\ \ \ \ \ \ \ \ \ \ \ \ \ \ \ \ \ \ \ \ \ \ \ \ \ \ \ \ }[-205.3, 8.65, 474.4, 30.0, -90.0],\\
\hbox{\ \ \ \ \ \ \ \ \ \ \ \ \ \ \ \ \ \ \ \ \ \ \ \ \ \ \ \ }[-205.3, 8.63, 474.4, 30.0, -90.0],\\
\hbox{\ \ \ \ \ \ \ \ \ \ \ \ \ \ \ \ \ \ \ \ \ \ \ \ \ \ \ \ }[-205.3, 8.6, 474.4, 30.0, -90.0],\\
\hbox{\ \ \ \ \ \ \ \ \ \ \ \ \ \ \ \ \ \ \ \ \ \ \ \ \ \ \ \ }[-205.3, 8.56, 474.4, 30.0, -90.0],\\
\hbox{\ \ \ \ \ \ \ \ \ \ \ \ \ \ \ \ \ \ \ \ \ \ \ \ \ \ \ \ }[-205.3, 8.56, 474.4, 30.0, -90.0],\\
\hbox{\ \ \ \ \ \ \ \ \ \ \ \ \ \ \ \ \ \ \ \ \ \ \ \ \ \ \ \ }[-205.3, 8.54, 474.4, 30.0, -90.0],\\
\hbox{\ \ \ \ \ \ \ \ \ \ \ \ \ \ \ \ \ \ \ \ \ \ \ \ \ \ \ \ }[-205.3, 8.51, 474.4, 30.0, -90.0],\\
\hbox{\ \ \ \ \ \ \ \ \ \ \ \ \ \ \ \ \ \ \ \ \ \ \ \ \ \ \ \ }[-205.3, 8.52, 474.4, 30.0, -90.0],\\
\hbox{\ \ \ \ \ \ \ \ \ \ \ \ \ \ \ \ \ \ \ \ \ \ \ \ \ \ \ \ }[-205.3, 8.5, 474.4, 30.0, -90.0],\\
\hbox{\ \ \ \ \ \ \ \ \ \ \ \ \ \ \ \ \ \ \ \ \ \ \ \ \ \ \ \ }[-205.3, 8.47, 474.4, 45.0, -90.0],\\
\hbox{\ \ \ \ \ \ \ \ \ \ \ \ \ \ \ \ \ \ \ \ \ \ \ \ \ \ \ \ }[-205.3, 8.43, 474.4, 30.0, -90.0],\\
\hbox{\ \ \ \ \ \ \ \ \ \ \ \ \ \ \ \ \ \ \ \ \ \ \ \ \ \ \ \ }[-205.3, 8.44, 474.4, 30.0, -90.0],\\
\hbox{\ \ \ \ \ \ \ \ \ \ \ \ \ \ \ \ \ \ \ \ \ \ \ \ \ \ \ \ }[-205.3, 8.42, 474.4, 45.0, -90.0]],\\
\hbox{\ \ \ \ \ \ \ \ \ \ \ \ \ \ }'nearest\_blocks': [\{'lava': [0.25, 0.87, -0.63]\},\\
\hbox{\ \ \ \ \ \ \ \ \ \ \ \ \ \ \ \ \ \ \ \ \ \ \ \ \ \ \ \ \ \ \ \ \ }\{'lava': [0.31, 0.87, -0.63]\},\\
\hbox{\ \ \ \ \ \ \ \ \ \ \ \ \ \ \ \ \ \ \ \ \ \ \ \ \ \ \ \ \ \ \ \ \ }\{'lava': [0.38, 0.87, -0.63]\},\\
\hbox{\ \ \ \ \ \ \ \ \ \ \ \ \ \ \ \ \ \ \ \ \ \ \ \ \ \ \ \ \ \ \ \ \ }\{'lava': [0.41, 0.07, -0.03]\},\\
\hbox{\ \ \ \ \ \ \ \ \ \ \ \ \ \ \ \ \ \ \ \ \ \ \ \ \ \ \ \ \ \ \ \ \ }\{'lava': [0.41, 0.07, -0.03]\},\\
\hbox{\ \ \ \ \ \ \ \ \ \ \ \ \ \ \ \ \ \ \ \ \ \ \ \ \ \ \ \ \ \ \ \ \ }\{'lava': [0.41, 0.07, -0.03]\},\\
\hbox{\ \ \ \ \ \ \ \ \ \ \ \ \ \ \ \ \ \ \ \ \ \ \ \ \ \ \ \ \ \ \ \ \ }\{'lava': [0.41, 0.07, -0.03]\},\\
\hbox{\ \ \ \ \ \ \ \ \ \ \ \ \ \ \ \ \ \ \ \ \ \ \ \ \ \ \ \ \ \ \ \ \ }\{'lava': [0.41, 0.35, -0.1]\},\\
\hbox{\ \ \ \ \ \ \ \ \ \ \ \ \ \ \ \ \ \ \ \ \ \ \ \ \ \ \ \ \ \ \ \ \ }\{'lava': [0.41, 0.35, -0.1]\},\\
\hbox{\ \ \ \ \ \ \ \ \ \ \ \ \ \ \ \ \ \ \ \ \ \ \ \ \ \ \ \ \ \ \ \ \ }\{'lava': [0.41, 0.35, -0.1]\},\\
\hbox{\ \ \ \ \ \ \ \ \ \ \ \ \ \ \ \ \ \ \ \ \ \ \ \ \ \ \ \ \ \ \ \ \ }\{'lava': [0.41, 0.35, -0.1]\},\\
\hbox{\ \ \ \ \ \ \ \ \ \ \ \ \ \ \ \ \ \ \ \ \ \ \ \ \ \ \ \ \ \ \ \ \ }\{'lava': [0.41, 0.35, -0.1]\},\\
\hbox{\ \ \ \ \ \ \ \ \ \ \ \ \ \ \ \ \ \ \ \ \ \ \ \ \ \ \ \ \ \ \ \ \ }\{'lava': [0.38, -1.57, 0.0]\},\\
\hbox{\ \ \ \ \ \ \ \ \ \ \ \ \ \ \ \ \ \ \ \ \ \ \ \ \ \ \ \ \ \ \ \ \ }\{'lava': [0.34, -1.57, 0.0]\},\\
\hbox{\ \ \ \ \ \ \ \ \ \ \ \ \ \ \ \ \ \ \ \ \ \ \ \ \ \ \ \ \ \ \ \ \ }\{'lava': [0.3, -1.57, 0.0]\},\\
\hbox{\ \ \ \ \ \ \ \ \ \ \ \ \ \ \ \ \ \ \ \ \ \ \ \ \ \ \ \ \ \ \ \ \ }\{'lava': [0.26, -1.57, 0.0]\},\\
\hbox{\ \ \ \ \ \ \ \ \ \ \ \ \ \ \ \ \ \ \ \ \ \ \ \ \ \ \ \ \ \ \ \ \ }\{'lava': [0.26, -1.57, 0.0]\},\\
\hbox{\ \ \ \ \ \ \ \ \ \ \ \ \ \ \ \ \ \ \ \ \ \ \ \ \ \ \ \ \ \ \ \ \ }\{'lava': [0.28, -1.57, 0.0]\},\\
\hbox{\ \ \ \ \ \ \ \ \ \ \ \ \ \ \ \ \ \ \ \ \ \ \ \ \ \ \ \ \ \ \ \ \ }\{'lava': [0.27, -1.57, 0.0]\},\\
\hbox{\ \ \ \ \ \ \ \ \ \ \ \ \ \ \ \ \ \ \ \ \ \ \ \ \ \ \ \ \ \ \ \ \ }\{'lava': [0.25, -0.87, -0.63]\},\\
\hbox{\ \ \ \ \ \ \ \ \ \ \ \ \ \ \ \ \ \ \ \ \ \ \ \ \ \ \ \ \ \ \ \ \ }\{'lava': [0.22, -0.87, -0.63]\},\\
\hbox{\ \ \ \ \ \ \ \ \ \ \ \ \ \ \ \ \ \ \ \ \ \ \ \ \ \ \ \ \ \ \ \ \ }\{'lava': [0.18, -0.87, -0.63]\},\\
\hbox{\ \ \ \ \ \ \ \ \ \ \ \ \ \ \ \ \ \ \ \ \ \ \ \ \ \ \ \ \ \ \ \ \ }\{'lava': [0.18, -0.87, -0.63]\},\\
\hbox{\ \ \ \ \ \ \ \ \ \ \ \ \ \ \ \ \ \ \ \ \ \ \ \ \ \ \ \ \ \ \ \ \ }\{'lava': [0.16, -0.87, -0.63]\},\\
\hbox{\ \ \ \ \ \ \ \ \ \ \ \ \ \ \ \ \ \ \ \ \ \ \ \ \ \ \ \ \ \ \ \ \ }\{'lava': [0.13, -0.87, -0.63]\},\\
\hbox{\ \ \ \ \ \ \ \ \ \ \ \ \ \ \ \ \ \ \ \ \ \ \ \ \ \ \ \ \ \ \ \ \ }\{'lava': [0.14, -0.87, -0.63]\},\\
\hbox{\ \ \ \ \ \ \ \ \ \ \ \ \ \ \ \ \ \ \ \ \ \ \ \ \ \ \ \ \ \ \ \ \ }\{'lava': [0.12, -0.87, -0.63]\},\\
\hbox{\ \ \ \ \ \ \ \ \ \ \ \ \ \ \ \ \ \ \ \ \ \ \ \ \ \ \ \ \ \ \ \ \ }\{'lava': [0.09, -0.87, -0.63]\},\\
\hbox{\ \ \ \ \ \ \ \ \ \ \ \ \ \ \ \ \ \ \ \ \ \ \ \ \ \ \ \ \ \ \ \ \ }\{'lava': [0.05, -1.57, 0.0]\},\\
\hbox{\ \ \ \ \ \ \ \ \ \ \ \ \ \ \ \ \ \ \ \ \ \ \ \ \ \ \ \ \ \ \ \ \ }\{'lava': [0.06, -0.87, -0.63]\},\\
\hbox{\ \ \ \ \ \ \ \ \ \ \ \ \ \ \ \ \ \ \ \ \ \ \ \ \ \ \ \ \ \ \ \ \ }\{'lava': [0.04, -0.87, -0.63]\},\\
\hbox{\ \ \ \ \ \ \ \ \ \ \ \ \ \ \ \ \ \ \ \ \ \ \ \ \ \ \ \ \ \ \ \ \ }\{'lava': [0.01, -0.87, -0.63]\}],\\
\hbox{\ \ \ \ \ \ \ \ \ \ \ \ \ \ }'rewards': [],\\
\hbox{\ \ \ \ \ \ \ \ \ \ \ \ \ \ }'truncated': True\}\},\\
\{'block\_under\_foot': 'stone',
\\
\hbox{\ \ }'dead': False,\\
\hbox{\ \ }'final\_health': 10.0,\\
\hbox{\ \ }'final\_inventory': \{'coal': 5,\\
\hbox{\ \ \ \ \ \ \ \ \ \ \ \ \ \ \ \ \ \ \ \ \ \ }'cobblestone': 133,\\
\hbox{\ \ \ \ \ \ \ \ \ \ \ \ \ \ \ \ \ \ \ \ \ \ }'crafting\_table': 1,\\
\hbox{\ \ \ \ \ \ \ \ \ \ \ \ \ \ \ \ \ \ \ \ \ \ }'dirt': 18,\\
\hbox{\ \ \ \ \ \ \ \ \ \ \ \ \ \ \ \ \ \ \ \ \ \ }'furnace': 1,\\
\hbox{\ \ \ \ \ \ \ \ \ \ \ \ \ \ \ \ \ \ \ \ \ \ }'iron\_pickaxe': 1,\\
\hbox{\ \ \ \ \ \ \ \ \ \ \ \ \ \ \ \ \ \ \ \ \ \ }'planks': 2,\\
\hbox{\ \ \ \ \ \ \ \ \ \ \ \ \ \ \ \ \ \ \ \ \ \ }'stone': 37,\\
\hbox{\ \ \ \ \ \ \ \ \ \ \ \ \ \ \ \ \ \ \ \ \ \ }'stone\_pickaxe': 9,\\
\hbox{\ \ \ \ \ \ \ \ \ \ \ \ \ \ \ \ \ \ \ \ \ \ }'wooden\_pickaxe': 1\},\\
\hbox{\ \ }'final\_nearest\_blocks': \{'stone': [0.5, 0.87, -0.63]\},\\
\hbox{\ \ }'history': \{'actions': ['attack', 'attack', 'attack', 'attack', 'attack',\\
\hbox{\ \ \ \ \ \ \ \ \ \ \ \ \ \ \ \ \ \ \ \ \ \ \ \ \ \ }'attack', 'attack', 'attack', 'attack', 'attack',\\
\hbox{\ \ \ \ \ \ \ \ \ \ \ \ \ \ \ \ \ \ \ \ \ \ \ \ \ \ }'attack', 'attack', 'attack', 'attack', 'attack',\\
\hbox{\ \ \ \ \ \ \ \ \ \ \ \ \ \ \ \ \ \ \ \ \ \ \ \ \ \ }'noop', 'turn\_up', 'turn\_up', 'turn\_up', 'noop',\\
\hbox{\ \ \ \ \ \ \ \ \ \ \ \ \ \ \ \ \ \ \ \ \ \ \ \ \ \ }'forward', 'forward', 'forward', 'forward', 'forward',\\
\hbox{\ \ \ \ \ \ \ \ \ \ \ \ \ \ \ \ \ \ \ \ \ \ \ \ \ \ }'forward', 'back', 'noop', 'noop', 'attack', 'attack',\\
\hbox{\ \ \ \ \ \ \ \ \ \ \ \ \ \ \ \ \ \ \ \ \ \ \ \ \ \ }'attack'],\\
\hbox{\ \ \ \ \ \ \ \ \ \ \ \ \ \ }'inventory\_change': [\{\}, \{\}, \{\}, \{'cobblestone': 1\}, \{\}, \{\}, \{\},\\
\hbox{\ \ \ \ \ \ \ \ \ \ \ \ \ \ \ \ \ \ \ \ \ \ \ \ \ \ \ \ \ \ \ \ \ \ \ }\{\}, \{\}, \{\}, \{\}, \{\}, \{\}, \{\}, \{\}, \{\}, \{\}, \{\},\\
\hbox{\ \ \ \ \ \ \ \ \ \ \ \ \ \ \ \ \ \ \ \ \ \ \ \ \ \ \ \ \ \ \ \ \ \ \ }\{\}, \{\}, \{\}, \{\}, \{\}, \{\}, \{'cobblestone': 1\},\\
\hbox{\ \ \ \ \ \ \ \ \ \ \ \ \ \ \ \ \ \ \ \ \ \ \ \ \ \ \ \ \ \ \ \ \ \ \ }\{\}, \{\}, \{\}, \{\}, \{\}, \{\}, \{\}],\\
\hbox{\ \ \ \ \ \ \ \ \ \ \ \ \ \ }'locations': [[611.54, 11.0, 212.67, 45.0, -180.0],\\
\hbox{\ \ \ \ \ \ \ \ \ \ \ \ \ \ \ \ \ \ \ \ \ \ \ \ \ \ \ \ }[611.54, 11.0, 212.67, 45.0, -180.0],\\
\hbox{\ \ \ \ \ \ \ \ \ \ \ \ \ \ \ \ \ \ \ \ \ \ \ \ \ \ \ \ }[611.54, 11.0, 212.67, 45.0, -180.0],\\
\hbox{\ \ \ \ \ \ \ \ \ \ \ \ \ \ \ \ \ \ \ \ \ \ \ \ \ \ \ \ }[611.54, 11.0, 212.67, 45.0, -180.0],\\
\hbox{\ \ \ \ \ \ \ \ \ \ \ \ \ \ \ \ \ \ \ \ \ \ \ \ \ \ \ \ }[611.54, 11.0, 212.67, 45.0, -180.0],\\
\hbox{\ \ \ \ \ \ \ \ \ \ \ \ \ \ \ \ \ \ \ \ \ \ \ \ \ \ \ \ }[611.54, 11.0, 212.67, 45.0, -180.0],\\
\hbox{\ \ \ \ \ \ \ \ \ \ \ \ \ \ \ \ \ \ \ \ \ \ \ \ \ \ \ \ }[611.54, 11.0, 212.67, 45.0, -180.0],\\
\hbox{\ \ \ \ \ \ \ \ \ \ \ \ \ \ \ \ \ \ \ \ \ \ \ \ \ \ \ \ }[611.54, 11.0, 212.67, 45.0, -180.0],\\
\hbox{\ \ \ \ \ \ \ \ \ \ \ \ \ \ \ \ \ \ \ \ \ \ \ \ \ \ \ \ }[611.54, 11.0, 212.67, 45.0, -180.0],\\
\hbox{\ \ \ \ \ \ \ \ \ \ \ \ \ \ \ \ \ \ \ \ \ \ \ \ \ \ \ \ }[611.54, 11.0, 212.67, 45.0, -180.0],\\
\hbox{\ \ \ \ \ \ \ \ \ \ \ \ \ \ \ \ \ \ \ \ \ \ \ \ \ \ \ \ }[611.54, 11.0, 212.67, 45.0, -180.0],\\
\hbox{\ \ \ \ \ \ \ \ \ \ \ \ \ \ \ \ \ \ \ \ \ \ \ \ \ \ \ \ }[611.54, 11.0, 212.67, 45.0, -180.0],\\
\hbox{\ \ \ \ \ \ \ \ \ \ \ \ \ \ \ \ \ \ \ \ \ \ \ \ \ \ \ \ }[611.54, 11.0, 212.67, 45.0, -180.0],\\
\hbox{\ \ \ \ \ \ \ \ \ \ \ \ \ \ \ \ \ \ \ \ \ \ \ \ \ \ \ \ }[611.54, 11.0, 212.67, 45.0, -180.0],\\
\hbox{\ \ \ \ \ \ \ \ \ \ \ \ \ \ \ \ \ \ \ \ \ \ \ \ \ \ \ \ }[611.54, 11.0, 212.67, 45.0, -180.0],\\
\hbox{\ \ \ \ \ \ \ \ \ \ \ \ \ \ \ \ \ \ \ \ \ \ \ \ \ \ \ \ }[611.54, 11.0, 212.67, 45.0, -180.0],\\
\hbox{\ \ \ \ \ \ \ \ \ \ \ \ \ \ \ \ \ \ \ \ \ \ \ \ \ \ \ \ }[611.54, 11.0, 212.67, 45.0, -180.0],\\
\hbox{\ \ \ \ \ \ \ \ \ \ \ \ \ \ \ \ \ \ \ \ \ \ \ \ \ \ \ \ }[611.54, 11.0, 212.67, 30.0, -180.0],\\
\hbox{\ \ \ \ \ \ \ \ \ \ \ \ \ \ \ \ \ \ \ \ \ \ \ \ \ \ \ \ }[611.54, 11.0, 212.67, 15.0, -180.0],\\
\hbox{\ \ \ \ \ \ \ \ \ \ \ \ \ \ \ \ \ \ \ \ \ \ \ \ \ \ \ \ }[611.54, 11.0, 212.67, 0.0, -180.0],\\
\hbox{\ \ \ \ \ \ \ \ \ \ \ \ \ \ \ \ \ \ \ \ \ \ \ \ \ \ \ \ }[611.54, 11.0, 212.67, 0.0, -180.0],\\
\hbox{\ \ \ \ \ \ \ \ \ \ \ \ \ \ \ \ \ \ \ \ \ \ \ \ \ \ \ \ }[611.54, 11.0, 212.57, 0.0, -180.0],\\
\hbox{\ \ \ \ \ \ \ \ \ \ \ \ \ \ \ \ \ \ \ \ \ \ \ \ \ \ \ \ }[611.54, 11.0, 212.42, 0.0, -180.0],\\
\hbox{\ \ \ \ \ \ \ \ \ \ \ \ \ \ \ \ \ \ \ \ \ \ \ \ \ \ \ \ }[611.54, 11.0, 212.24, 0.0, -180.0],\\
\hbox{\ \ \ \ \ \ \ \ \ \ \ \ \ \ \ \ \ \ \ \ \ \ \ \ \ \ \ \ }[611.54, 11.0, 212.04, 0.0, -180.0],\\
\hbox{\ \ \ \ \ \ \ \ \ \ \ \ \ \ \ \ \ \ \ \ \ \ \ \ \ \ \ \ }[611.54, 11.0, 211.84, 0.0, -180.0],\\
\hbox{\ \ \ \ \ \ \ \ \ \ \ \ \ \ \ \ \ \ \ \ \ \ \ \ \ \ \ \ }[611.54, 11.0, 211.63, 0.0, -180.0],\\
\hbox{\ \ \ \ \ \ \ \ \ \ \ \ \ \ \ \ \ \ \ \ \ \ \ \ \ \ \ \ }[611.54, 11.0, 211.61, 0.0, -180.0],\\
\hbox{\ \ \ \ \ \ \ \ \ \ \ \ \ \ \ \ \ \ \ \ \ \ \ \ \ \ \ \ }[611.54, 11.0, 211.6, 0.0, -180.0],\\
\hbox{\ \ \ \ \ \ \ \ \ \ \ \ \ \ \ \ \ \ \ \ \ \ \ \ \ \ \ \ }[611.54, 11.0, 211.6, 0.0, -180.0],\\
\hbox{\ \ \ \ \ \ \ \ \ \ \ \ \ \ \ \ \ \ \ \ \ \ \ \ \ \ \ \ }[611.54, 11.0, 211.6, 0.0, -180.0],\\
\hbox{\ \ \ \ \ \ \ \ \ \ \ \ \ \ \ \ \ \ \ \ \ \ \ \ \ \ \ \ }[611.54, 11.0, 211.6, 0.0, -180.0]],\\
\hbox{\ \ \ \ \ \ \ \ \ \ \ \ \ \ }'nearest\_blocks': [\{'stone': [0.78, 0.73, -0.63]\},\\
\hbox{\ \ \ \ \ \ \ \ \ \ \ \ \ \ \ \ \ \ \ \ \ \ \ \ \ \ \ \ \ \ \ \ \ }\{'stone': [0.78, 0.73, -0.63]\},\\
\hbox{\ \ \ \ \ \ \ \ \ \ \ \ \ \ \ \ \ \ \ \ \ \ \ \ \ \ \ \ \ \ \ \ \ }\{'stone': [0.78, 0.73, -0.63]\},\\
\hbox{\ \ \ \ \ \ \ \ \ \ \ \ \ \ \ \ \ \ \ \ \ \ \ \ \ \ \ \ \ \ \ \ \ }\{'stone': [0.78, 0.73, -0.63]\},\\
\hbox{\ \ \ \ \ \ \ \ \ \ \ \ \ \ \ \ \ \ \ \ \ \ \ \ \ \ \ \ \ \ \ \ \ }\{'stone': [0.78, 0.73, -0.63]\},\\
\hbox{\ \ \ \ \ \ \ \ \ \ \ \ \ \ \ \ \ \ \ \ \ \ \ \ \ \ \ \ \ \ \ \ \ }\{'stone': [0.78, 0.73, -0.63]\},\\
\hbox{\ \ \ \ \ \ \ \ \ \ \ \ \ \ \ \ \ \ \ \ \ \ \ \ \ \ \ \ \ \ \ \ \ }\{'stone': [0.78, 0.73, -0.63]\},\\
\hbox{\ \ \ \ \ \ \ \ \ \ \ \ \ \ \ \ \ \ \ \ \ \ \ \ \ \ \ \ \ \ \ \ \ }\{'stone': [0.78, 0.73, -0.63]\},\\
\hbox{\ \ \ \ \ \ \ \ \ \ \ \ \ \ \ \ \ \ \ \ \ \ \ \ \ \ \ \ \ \ \ \ \ }\{'stone': [0.78, 0.73, -0.63]\},\\
\hbox{\ \ \ \ \ \ \ \ \ \ \ \ \ \ \ \ \ \ \ \ \ \ \ \ \ \ \ \ \ \ \ \ \ }\{'stone': [0.78, 0.73, -0.63]\},\\
\hbox{\ \ \ \ \ \ \ \ \ \ \ \ \ \ \ \ \ \ \ \ \ \ \ \ \ \ \ \ \ \ \ \ \ }\{'stone': [0.78, 0.73, -0.63]\},\\
\hbox{\ \ \ \ \ \ \ \ \ \ \ \ \ \ \ \ \ \ \ \ \ \ \ \ \ \ \ \ \ \ \ \ \ }\{'stone': [0.78, 0.73, -0.63]\},\\
\hbox{\ \ \ \ \ \ \ \ \ \ \ \ \ \ \ \ \ \ \ \ \ \ \ \ \ \ \ \ \ \ \ \ \ }\{'stone': [0.78, 0.73, -0.63]\},\\
\hbox{\ \ \ \ \ \ \ \ \ \ \ \ \ \ \ \ \ \ \ \ \ \ \ \ \ \ \ \ \ \ \ \ \ }\{'stone': [0.78, 0.73, -0.63]\},\\
\hbox{\ \ \ \ \ \ \ \ \ \ \ \ \ \ \ \ \ \ \ \ \ \ \ \ \ \ \ \ \ \ \ \ \ }\{'stone': [0.78, 0.73, -0.63]\},\\
\hbox{\ \ \ \ \ \ \ \ \ \ \ \ \ \ \ \ \ \ \ \ \ \ \ \ \ \ \ \ \ \ \ \ \ }\{'stone': [0.78, 0.73, -0.63]\},\\
\hbox{\ \ \ \ \ \ \ \ \ \ \ \ \ \ \ \ \ \ \ \ \ \ \ \ \ \ \ \ \ \ \ \ \ }\{'stone': [0.78, 0.49, -0.63]\},\\
\hbox{\ \ \ \ \ \ \ \ \ \ \ \ \ \ \ \ \ \ \ \ \ \ \ \ \ \ \ \ \ \ \ \ \ }\{'stone': [0.66, 0.87, -0.63]\},\\
\hbox{\ \ \ \ \ \ \ \ \ \ \ \ \ \ \ \ \ \ \ \ \ \ \ \ \ \ \ \ \ \ \ \ \ }\{'stone': [0.5, 0.87, -0.63]\},\\
\hbox{\ \ \ \ \ \ \ \ \ \ \ \ \ \ \ \ \ \ \ \ \ \ \ \ \ \ \ \ \ \ \ \ \ }\{'stone': [0.5, 0.87, -0.63]\},\\
\hbox{\ \ \ \ \ \ \ \ \ \ \ \ \ \ \ \ \ \ \ \ \ \ \ \ \ \ \ \ \ \ \ \ \ }\{'stone': [0.5, 0.87, -0.63]\},\\
\hbox{\ \ \ \ \ \ \ \ \ \ \ \ \ \ \ \ \ \ \ \ \ \ \ \ \ \ \ \ \ \ \ \ \ }\{'stone': [0.5, 0.87, -0.63]\},\\
\hbox{\ \ \ \ \ \ \ \ \ \ \ \ \ \ \ \ \ \ \ \ \ \ \ \ \ \ \ \ \ \ \ \ \ }\{'stone': [0.5, 0.87, -0.63]\},\\
\hbox{\ \ \ \ \ \ \ \ \ \ \ \ \ \ \ \ \ \ \ \ \ \ \ \ \ \ \ \ \ \ \ \ \ }\{'stone': [0.5, 0.87, -0.63]\},\\
\hbox{\ \ \ \ \ \ \ \ \ \ \ \ \ \ \ \ \ \ \ \ \ \ \ \ \ \ \ \ \ \ \ \ \ }\{'stone': [0.5, 0.87, -0.63]\},\\
\hbox{\ \ \ \ \ \ \ \ \ \ \ \ \ \ \ \ \ \ \ \ \ \ \ \ \ \ \ \ \ \ \ \ \ }\{'stone': [0.5, 0.87, -0.63]\},\\
\hbox{\ \ \ \ \ \ \ \ \ \ \ \ \ \ \ \ \ \ \ \ \ \ \ \ \ \ \ \ \ \ \ \ \ }\{'stone': [0.5, 0.87, -0.63]\},\\
\hbox{\ \ \ \ \ \ \ \ \ \ \ \ \ \ \ \ \ \ \ \ \ \ \ \ \ \ \ \ \ \ \ \ \ }\{'stone': [0.5, 0.87, -0.63]\},\\
\hbox{\ \ \ \ \ \ \ \ \ \ \ \ \ \ \ \ \ \ \ \ \ \ \ \ \ \ \ \ \ \ \ \ \ }\{'stone': [0.5, 0.87, -0.63]\},\\
\hbox{\ \ \ \ \ \ \ \ \ \ \ \ \ \ \ \ \ \ \ \ \ \ \ \ \ \ \ \ \ \ \ \ \ }\{'stone': [0.5, 0.87, -0.63]\},\\
\hbox{\ \ \ \ \ \ \ \ \ \ \ \ \ \ \ \ \ \ \ \ \ \ \ \ \ \ \ \ \ \ \ \ \ }\{'stone': [0.5, 0.87, -0.63]\},\\
\hbox{\ \ \ \ \ \ \ \ \ \ \ \ \ \ \ \ \ \ \ \ \ \ \ \ \ \ \ \ \ \ \ \ \ }\{'stone': [0.5, 0.87, -0.63]\}],\\
\hbox{\ \ \ \ \ \ \ \ \ \ \ \ \ \ }'rewards': [],\\
\hbox{\ \ \ \ \ \ \ \ \ \ \ \ \ \ }'truncated': True\}\}]\\
\#\#\# Statistics

\{'success\_rate': 0.0\}

--------------------------------------------------------------------------------------------------------------------------------------------

\texttt{\textbf{ANALYZER OUTPUT:}}

Analysis of Possible Reasons of Failure and Inefficiency: \hl{In one of the failed trajectories, the agent died due to lava. The agent did not take any action to avoid the lava or move away from it. The agent should be programmed to detect and avoid dangerous blocks, such as lava, by turning or moving in a different direction.}

Suggestion on Game Strategy: \hl{Train the agent to detect and avoid dangerous blocks, such as lava or deep holes. The agent should be programmed to prioritize safety and avoid situations that could lead to death or loss of health.}

By implementing the suggestion and improving the agent's training, the success rate of finding diamonds can be increased.

--------------------------------------------------------------------------------------------------------------------------------------------

\texttt{\textbf{DESIGNER INPUT:}}

You are now a proficient Minecraft player. I now have a reward function to train a Minecraft agent with reinforcement learning to complete the described task. The reward function is used to train the reinforcement learning agent for several times. I will provide you the analysis of failure and inefficiency and suggestions. You should help me modify the reward function.

\#\# Task description

- Objective: Find and approach diamond with higher success rate and avoid death.

- Initial Status: 
\\
\hbox{\ \ }1. The agent is under ground at y level 11, the most common level to find diamonds.\\
\hbox{\ \ }2. The agent already has an iron pickaxe.\\
- Success criteria: The distance to the nearest diamond block is less than 2 meter. 

- Procedure: Explore horizontally to find a diamond, face it and approach it.

\#\# Input parameters of the reward function

- current\_nearest\_blocks: nearby block types and nearest distance of each block type at the **current step**. It is a dictionary, consisting of the names, relative distances with the agent, and relative angles (yaw and pitch) with the agent, in the form of: \{NAME\_1: (DIST\_1, YAW\_1, PITCH\_1), NAME\_2: (DIST\_2, YAW\_2, PITCH\_2)\}. For example, \{"wood": (24.7, 1.48, -1.57), "cobblestone": (1.3, -0.17, 1.05), "iron\_ore": (4.5, 0.61, -0.17)\}. If a certrain block type does not exist in the field of view, it is not present in the dictionary. The yaw and pitch here is relative to the agent's front, i.e., pitch = 0 and yaw = 0 is the front of the agent.

- previou\_nearest\_blocks: nearby block types and nearest distance of each block type at the **previous step**, with the same format as the current\_nearest\_blocks.

- inventory\_change: the change of the agent's inventory from the previous step to current step, in the form of a dictionary: \{NAME\_1: CHANGE\_NUM\_1, NAME\_2: CHANGE\_NUM\_2\}. Positive values mean increase and negative values mean decrease. For example, \{"wood": 2, "dirt": 3, "stone\_pickaxe": -1\}.

- health: an integer value in range 0 to 10 indicating the health level of the agent. 0 means death and 10 means full health.

- past\_agent\_positions: the history of location of agent, in the form of a list: [[x1, y1, z1, yaw1, pitch1], [x2, y2, z2, yaw2, pitch2], ...]. The yaw and pitch here are relative to the agent's initial forward direction, i.e., pitch = 0 and yaw = 0 is the front of the agent when it was born. The length of the list is the number of steps the agent has taken. The last element of the list is the current location of the agent.

- GLOBAL\_DATA: a global variable. It is initialized as a dictionary. You can save necessary information between different steps with it.

\#\# General game information

- The version of Minecraft is 1.11.

- Common block names : dirt, cobblestone, iron, diamond, wood, coal, water, air, lava, leaves, ... Collected item names are the same.

- FOV is set to -35 to 35 degrees for yaw and -30 to 30 degrees for pitch. The max visible distance of blocks is 64.

- Length of each block is 1 meter.

- The y coordinate of the agent is the agent's height. The larger the y, the higher the agent. The ground level is around y = 63, but is not fixed.

- At each step, the agent selects one operation in is action space. The action space includes doing nothing (i.e. staying still), moving forward and backward, jumping and attacking. The action space also includes turning left, right, up and down. 

- The attack range is 2 meters in front of the agent. The agent can move 0.2 meters in each step. The agent can turn 15 degrees in each step. Typically, the agent needs to stay still and attack for **60 successive steps** to break a block.

- The hunger value is always at the max level.

\#\# Reward function requirements

- You should write a dense reward function \textasciigrave dense\textasciigrave  and a sparse reward function \textasciigrave sparse\textasciigrave . The sparse reward indicates achieving the goal or receiving heavy punishment. The dense reward provides intermediate signal to guide the agent in the process of achieving the goal. The magnitude of the return value does not matter, but the sign (positive or negative) is important. The final reward will be \textasciigrave np.sign(sparse(...)) * 1 + np.sign(dense(...)) * 0.1\textasciigrave .

\#\# Current reward function

def reward\_function(current\_nearest\_blocks, previous\_nearest\_blocks, inventory\_change, health, past\_agent\_positions, GLOBAL\_DATA):
\\
\hbox{\ \ \ \ }"""\\
\hbox{\ \ \ \ }Thoughts:\\
\hbox{\ \ \ \ }In order to make the agent approach the diamond, we need to give positive rewards for actions that\\
\hbox{\ \ \ \ }make the agent get closer to the diamond and negative rewards for actions that make the agent get further from the diamond.\\
\hbox{\ \ \ \ }We also need to give a negative reward if the agent's health decreases to encourage the agent to avoid dangerous situations.\\
\hbox{\ \ \ \ }The dense reward function will be based on the distance to the diamond and health.\\
\hbox{\ \ \ \ }The sparse reward function will give a large positive reward when the agent reaches the diamond and a large negative reward when the agent dies.\\
\hbox{\ \ \ \ }"""\\
\hbox{\ \ \ \ }import numpy as np\\
\hbox{\ \ \ \ }def dense(current\_nearest\_blocks, previous\_nearest\_blocks, inventory\_change, health, past\_agent\_positions, GLOBAL\_DATA):\\
\hbox{\ \ \ \ \ \ \ \ }\# Initialize reward\\
\hbox{\ \ \ \ \ \ \ \ }reward = 0\\
\hbox{\ \ \ \ \ \ \ \ }\# If the agent is closer to diamond than before, give a positive reward\\
\hbox{\ \ \ \ \ \ \ \ }if "diamond" in current\_nearest\_blocks and "diamond" in previous\_nearest\_blocks:\\
\hbox{\ \ \ \ \ \ \ \ \ \ \ \ }if current\_nearest\_blocks["diamond"][0] < previous\_nearest\_blocks["diamond"][0]:\\
\hbox{\ \ \ \ \ \ \ \ \ \ \ \ \ \ \ \ }reward += 1\\
\hbox{\ \ \ \ \ \ \ \ \ \ \ \ }elif current\_nearest\_blocks["diamond"][0] > previous\_nearest\_blocks["diamond"][0]:\\
\hbox{\ \ \ \ \ \ \ \ \ \ \ \ \ \ \ \ }reward -= 1\\
\hbox{\ \ \ \ \ \ \ \ }\# If the agent's health decreases, give a negative reward\\
\hbox{\ \ \ \ \ \ \ \ }if "health" in GLOBAL\_DATA:\\
\hbox{\ \ \ \ \ \ \ \ \ \ \ \ }if health < GLOBAL\_DATA["health"]:\\
\hbox{\ \ \ \ \ \ \ \ \ \ \ \ \ \ \ \ }reward -= 1\\
\hbox{\ \ \ \ \ \ \ \ }GLOBAL\_DATA["health"] = health\\
\hbox{\ \ \ \ \ \ \ \ }return reward\\
\hbox{\ \ \ \ }def sparse(current\_nearest\_blocks, previous\_nearest\_blocks, inventory\_change, health, past\_agent\_positions, GLOBAL\_DATA):\\
\hbox{\ \ \ \ \ \ \ \ }\# Initialize reward\\
\hbox{\ \ \ \ \ \ \ \ }reward = 0\\
\hbox{\ \ \ \ \ \ \ \ }\# If the agent reaches the diamond, give a large positive reward\\
\hbox{\ \ \ \ \ \ \ \ }if "diamond" in current\_nearest\_blocks and current\_nearest\_blocks["diamond"][0] < 1:\\
\hbox{\ \ \ \ \ \ \ \ \ \ \ \ }reward += 100\\
\hbox{\ \ \ \ \ \ \ \ }\# If the agent dies, give a large negative reward\\
\hbox{\ \ \ \ \ \ \ \ }if health == 0:\\
\hbox{\ \ \ \ \ \ \ \ \ \ \ \ }reward -= 100\\
\hbox{\ \ \ \ \ \ \ \ }return reward\\
\hbox{\ \ \ \ }dense\_reward = dense(current\_nearest\_blocks, previous\_nearest\_blocks, inventory\_change, health, past\_agent\_positions, GLOBAL\_DATA)\\
\hbox{\ \ \ \ }sparse\_reward = sparse(current\_nearest\_blocks, previous\_nearest\_blocks, inventory\_change, health, past\_agent\_positions, GLOBAL\_DATA)\\
\hbox{\ \ \ \ }return np.sign(dense\_reward) * 1 + np.sign(sparse\_reward) * 0.1\\
\#\# Reward function test result

The reward function is used to train the reinforcement learning agent for several times. Here is some analysis of failure and inefficiency and suggestions:

\textasciigrave \textasciigrave \textasciigrave 

Analysis of Possible Reasons of Failure and Inefficiency: In one of the failed trajectories, the agent died due to lava. The agent did not take any action to avoid the lava or move away from it. The agent should be programmed to detect and avoid dangerous blocks, such as lava, by turning or moving in a different direction.

Suggestion on Game Strategy: Train the agent to detect and avoid dangerous blocks, such as lava or deep holes. The agent should be programmed to prioritize safety and avoid situations that could lead to death or loss of health.

By implementing the suggestion and improving the agent's training, the success rate of finding diamonds can be increased.

\textasciigrave \textasciigrave \textasciigrave

\#\# Requirements

Please consider the analysis and suggestions above and modify the reward function.

1. You can both modify the current lines and add new lines.

2. If necessary, you can write a **totally different** reward function than the current one.

3. Consider modifing the reward and penalty values in the current reward function to balance them.

4. In the first part of the reward function, you should provide your thoughts of modifying the reward function. **The thoughts should be concise.**

\#\# Output format

Strictly follow the following format. **Do not output anything else outside the code block**.

def reward\_function(current\_nearest\_blocks, previous\_nearest\_blocks, inventory\_change, health, past\_agent\_positions, GLOBAL\_DATA):
\\
\hbox{\ \ \ \ }\# Thoughts:\\
\hbox{\ \ \ \ }\# ...\\
\hbox{\ \ \ \ }\# (import packages and define helper functions)\\
\hbox{\ \ \ \ }import numpy as np\\
\hbox{\ \ \ \ }...\\
\hbox{\ \ \ \ }def dense(current\_nearest\_blocks, previous\_nearest\_blocks, inventory\_change, health, past\_agent\_positions, GLOBAL\_DATA):\\
\hbox{\ \ \ \ \ \ \ \ }...\\
\hbox{\ \ \ \ }def sparse(current\_nearest\_blocks, previous\_nearest\_blocks, inventory\_change, health, past\_agent\_positions, GLOBAL\_DATA):\\
\hbox{\ \ \ \ \ \ \ \ }...\\
\hbox{\ \ \ \ }dense\_reward = dense(current\_nearest\_blocks, previous\_nearest\_blocks, inventory\_change, health, past\_agent\_positions, GLOBAL\_DATA)\\
\hbox{\ \ \ \ }sparse\_reward = sparse(current\_nearest\_blocks, previous\_nearest\_blocks, inventory\_change, health, past\_agent\_positions, GLOBAL\_DATA)\\
\hbox{\ \ \ \ }return np.sign(sparse\_reward) * 1 + np.sign(dense\_reward) * 0.1\\
\hbox{\ \ \ \ }...\\
--------------------------------------------------------------------------------------------------------------------------------------------

\texttt{\textbf{DESIGNER OUTPUT:}}

def reward\_function(current\_nearest\_blocks, previous\_nearest\_blocks, inventory\_change, health, past\_agent\_positions, GLOBAL\_DATA):
\\
\hbox{\ \ \ \ }"""\\
\hbox{\ \ \ \ }Thoughts:\\
\hbox{\ \ \ \ }In order to make the agent approach the diamond, we need to give positive rewards for actions that\\
\hbox{\ \ \ \ }make the agent get closer to the diamond and negative rewards for actions that make the agent get further from the diamond.\\
\hbox{\ \ \ \ }We also need to give a negative reward if the agent's health decreases to encourage the agent to avoid dangerous situations.\\
\hbox{\ \ \ \ }The dense reward function will be based on the distance to the diamond and health.\\
\hbox{\ \ \ \ }The sparse reward function will give a large positive reward when the agent reaches the diamond and a large negative reward when the agent dies.\\
\hbox{\ \ \ \ }In addition, we should implement a threat detection mechanism to avoid unnecessary deaths.\\
\hbox{\ \ \ \ }"""\\
\hbox{\ \ \ \ }import numpy as np\\
\hbox{\ \ \ \ }def dense(current\_nearest\_blocks, previous\_nearest\_blocks, inventory\_change, health, past\_agent\_positions, GLOBAL\_DATA):\\
\hbox{\ \ \ \ \ \ \ \ }\# Initialize reward\\
\hbox{\ \ \ \ \ \ \ \ }reward = 0\\
\hbox{\ \ \ \ \ \ \ \ }\# If the agent is closer to diamond than before, give a positive reward\\
\hbox{\ \ \ \ \ \ \ \ }if "diamond" in current\_nearest\_blocks and "diamond" in previous\_nearest\_blocks:\\
\hbox{\ \ \ \ \ \ \ \ \ \ \ \ }if current\_nearest\_blocks["diamond"][0] < previous\_nearest\_blocks["diamond"][0]:\\
\hbox{\ \ \ \ \ \ \ \ \ \ \ \ \ \ \ \ }reward += 1\\
\hbox{\ \ \ \ \ \ \ \ \ \ \ \ }elif current\_nearest\_blocks["diamond"][0] > previous\_nearest\_blocks["diamond"][0]:\\
\hbox{\ \ \ \ \ \ \ \ \ \ \ \ \ \ \ \ }reward -= 1\\
\hbox{\ \ \ \ \ \ \ \ }\hl{\# If the agent is close to lava, give a negative reward}\\
\hbox{\ \ \ \ \ \ \ \ }\hl{if "lava" in current\_nearest\_blocks:}\\
\hbox{\ \ \ \ \ \ \ \ \ \ \ \ }\hl{reward -= 1}\\
\hbox{\ \ \ \ \ \ \ \ }\# If the agent's health decreases, give a negative reward\\
\hbox{\ \ \ \ \ \ \ \ }if "health" in GLOBAL\_DATA:\\
\hbox{\ \ \ \ \ \ \ \ \ \ \ \ }if health < GLOBAL\_DATA["health"]:\\
\hbox{\ \ \ \ \ \ \ \ \ \ \ \ \ \ \ \ }reward -= 1\\
\hbox{\ \ \ \ \ \ \ \ }GLOBAL\_DATA["health"] = health\\
\hbox{\ \ \ \ \ \ \ \ }return reward\\
\hbox{\ \ \ \ }def sparse(current\_nearest\_blocks, previous\_nearest\_blocks, inventory\_change, health, past\_agent\_positions, GLOBAL\_DATA):\\
\hbox{\ \ \ \ \ \ \ \ }\# Initialize reward\\
\hbox{\ \ \ \ \ \ \ \ }reward = 0\\
\hbox{\ \ \ \ \ \ \ \ }\# If the agent reaches the diamond, give a large positive reward\\
\hbox{\ \ \ \ \ \ \ \ }if "diamond" in current\_nearest\_blocks and current\_nearest\_blocks["diamond"][0] < 1:\\
\hbox{\ \ \ \ \ \ \ \ \ \ \ \ }reward += 100\\
\hbox{\ \ \ \ \ \ \ \ }\# If the agent dies, give a large negative reward\\
\hbox{\ \ \ \ \ \ \ \ }if health == 0:\\
\hbox{\ \ \ \ \ \ \ \ \ \ \ \ }reward -= 100\\
\hbox{\ \ \ \ \ \ \ \ }return reward\\
\hbox{\ \ \ \ }dense\_reward = dense(current\_nearest\_blocks, previous\_nearest\_blocks, inventory\_change, health, past\_agent\_positions, GLOBAL\_DATA)\\
\hbox{\ \ \ \ }sparse\_reward = sparse(current\_nearest\_blocks, previous\_nearest\_blocks, inventory\_change, health, past\_agent\_positions, GLOBAL\_DATA)\\
\hbox{\ \ \ \ }return np.sign(dense\_reward) * 1 + np.sign(sparse\_reward) * 0.1\\
--------------------------------------------------------------------------------------------------------------------------------------------

\texttt{\textbf{CRITIC INPUT:}}

You are now a proficient Minecraft player. I will give you a reward function written in Python, which is used to train an RL agent to play Minecraft. 

You should help me to evaluate whether the reward function has met the requirements.

\#\# Task description

- Objective: Find and approach diamond with higher success rate and avoid death.

- Initial Status: 
\\
\hbox{\ \ }1. The agent is under ground at y level 11, the most common level to find diamonds.\\
\hbox{\ \ }2. The agent already has an iron pickaxe.\\
- Success criteria: The distance to the nearest diamond block is less than 2 meter. 

- Procedure: Explore horizontally to find a diamond, face it and approach it.

\#\# Input parameters of the reward function

- current\_nearest\_blocks: nearby block types and nearest distance of each block type at the **current step**. It is a dictionary, consisting of the names, relative distances with the agent, and relative angles (yaw and pitch) with the agent, in the form of: \{NAME\_1: (DIST\_1, YAW\_1, PITCH\_1), NAME\_2: (DIST\_2, YAW\_2, PITCH\_2)\}. For example, \{"wood": (24.7, 1.48, -1.57), "cobblestone": (1.3, -0.17, 1.05), "iron\_ore": (4.5, 0.61, -0.17)\}. If a certrain block type does not exist in the field of view, it is not present in the dictionary. The yaw and pitch here is relative to the agent's front, i.e., pitch = 0 and yaw = 0 is the front of the agent.

- previou\_nearest\_blocks: nearby block types and nearest distance of each block type at the **previous step**, with the same format as the current\_nearest\_blocks.

- inventory\_change: the change of the agent's inventory from the previous step to current step, in the form of a dictionary: \{NAME\_1: CHANGE\_NUM\_1, NAME\_2: CHANGE\_NUM\_2\}. Positive values mean increase and negative values mean decrease. For example, \{"wood": 2, "dirt": 3, "stone\_pickaxe": -1\}.

- health: an integer value in range 0 to 10 indicating the health level of the agent. 0 means death and 10 means full health.

- past\_agent\_positions: the history of location of agent, in the form of a list: [[x1, y1, z1, yaw1, pitch1], [x2, y2, z2, yaw2, pitch2], ...]. The yaw and pitch here are relative to the agent's initial forward direction, i.e., pitch = 0 and yaw = 0 is the front of the agent when it was born. The length of the list is the number of steps the agent has taken. The last element of the list is the current location of the agent.

- GLOBAL\_DATA: a global variable. It is initialized as a dictionary. You can save necessary information between different steps with it.

\#\# General game information

- The version of Minecraft is 1.11.

- Common block names : dirt, cobblestone, iron, diamond, wood, coal, water, air, lava, leaves, ... Collected item names are the same.

- FOV is set to -35 to 35 degrees for yaw and -30 to 30 degrees for pitch. The max visible distance of blocks is 64.

- Length of each block is 1 meter.

- The y coordinate of the agent is the agent's height. The larger the y, the higher the agent. The ground level is around y = 63, but is not fixed.

- At each step, the agent selects one operation in is action space. The action space includes doing nothing (i.e. staying still), moving forward and backward, jumping and attacking. The action space also includes turning left, right, up and down. 

- The attack range is 2 meters in front of the agent. The agent can move 0.2 meters in each step. The agent can turn 15 degrees in each step. Typically, the agent needs to stay still and attack for **60 successive steps** to break a block.

- The hunger value is always at the max level.

\#\# Reward function requirements

- You should write a dense reward function \textasciigrave dense\textasciigrave  and a sparse reward function \textasciigrave sparse\textasciigrave . The sparse reward indicates achieving the goal or receiving heavy punishment. The dense reward provides intermediate signal to guide the agent in the process of achieving the goal. The magnitude of the return value does not matter, but the sign (positive or negative) is important. The final reward will be \textasciigrave np.sign(sparse(...)) * 1 + np.sign(dense(...)) * 0.1\textasciigrave .

\#\# Requirements

1. You should check whether the reward function meets the **reward function requirements** above. 

2. Your judgement should consider whether the comments and the detailed implementation code are consistent. You can judge whether a line is correct by its comment, but do only rely on the comments.

3. You should also check the format of the reward function. It should be like:

def reward\_function(current\_nearest\_blocks, previous\_nearest\_blocks, inventory\_change, health, past\_agent\_positions, GLOBAL\_DATA):
\\
\hbox{\ \ \ \ }\# Thoughts:\\
\hbox{\ \ \ \ }\# ...\\
\hbox{\ \ \ \ }\# (import packages and define helper functions)\\
\hbox{\ \ \ \ }import numpy as np\\
\hbox{\ \ \ \ }...\\
\hbox{\ \ \ \ }def dense(current\_nearest\_blocks, previous\_nearest\_blocks, inventory\_change, health, past\_agent\_positions, GLOBAL\_DATA):\\
\hbox{\ \ \ \ \ \ \ \ }...\\
\hbox{\ \ \ \ }def sparse(current\_nearest\_blocks, previous\_nearest\_blocks, inventory\_change, health, past\_agent\_positions, GLOBAL\_DATA):\\
\hbox{\ \ \ \ \ \ \ \ }...\\
\hbox{\ \ \ \ }dense\_reward = dense(current\_nearest\_blocks, previous\_nearest\_blocks, inventory\_change, health, past\_agent\_positions, GLOBAL\_DATA)\\
\hbox{\ \ \ \ }sparse\_reward = sparse(current\_nearest\_blocks, previous\_nearest\_blocks, inventory\_change, health, past\_agent\_positions, GLOBAL\_DATA)\\
\hbox{\ \ \ \ }return np.sign(sparse\_reward) * 1 + np.sign(dense\_reward) * 0.1\\
\hbox{\ \ \ \ }...\\
4. Consider the **Information about the game state** above.

5. **You don't need to check whether the reward function meets the task description**.

6. You should first provide a reasoning and decision. If the reward function fails, provide a critique to help to improve the reward function.

\#\# Output format

You should only respond in JSON format as described below:

\{
\\
\hbox{\ \ \ \ }"reasoning": "reasoning",\\
\hbox{\ \ \ \ }"success": boolean,\\
\hbox{\ \ \ \ }"critique": critique,\\
\}

Ensure the response can be parsed by Python \textasciigrave json.loads\textasciigrave  , e.g. no trailing commas, no single quotes, etc. Do not output anything else like explanations.

\#\# Reward function to be evaluated

def reward\_function(current\_nearest\_blocks, previous\_nearest\_blocks, inventory\_change, health, past\_agent\_positions, GLOBAL\_DATA):
\\
\hbox{\ \ \ \ }"""\\
\hbox{\ \ \ \ }Thoughts:\\
\hbox{\ \ \ \ }In order to make the agent approach the diamond, we need to give positive rewards for actions that\\
\hbox{\ \ \ \ }make the agent get closer to the diamond and negative rewards for actions that make the agent get further from the diamond.\\
\hbox{\ \ \ \ }We also need to give a negative reward if the agent's health decreases to encourage the agent to avoid dangerous situations.\\
\hbox{\ \ \ \ }The dense reward function will be based on the distance to the diamond and health.\\
\hbox{\ \ \ \ }The sparse reward function will give a large positive reward when the agent reaches the diamond and a large negative reward when the agent dies.\\
\hbox{\ \ \ \ }In addition, we should implement a threat detection mechanism to avoid unnecessary deaths.\\
\hbox{\ \ \ \ }"""\\
\hbox{\ \ \ \ }import numpy as np\\
\hbox{\ \ \ \ }def dense(current\_nearest\_blocks, previous\_nearest\_blocks, inventory\_change, health, past\_agent\_positions, GLOBAL\_DATA):\\
\hbox{\ \ \ \ \ \ \ \ }\# Initialize reward\\
\hbox{\ \ \ \ \ \ \ \ }reward = 0\\
\hbox{\ \ \ \ \ \ \ \ }\# If the agent is closer to diamond than before, give a positive reward\\
\hbox{\ \ \ \ \ \ \ \ }if "diamond" in current\_nearest\_blocks and "diamond" in previous\_nearest\_blocks:\\
\hbox{\ \ \ \ \ \ \ \ \ \ \ \ }if current\_nearest\_blocks["diamond"][0] < previous\_nearest\_blocks["diamond"][0]:\\
\hbox{\ \ \ \ \ \ \ \ \ \ \ \ \ \ \ \ }reward += 1\\
\hbox{\ \ \ \ \ \ \ \ \ \ \ \ }elif current\_nearest\_blocks["diamond"][0] > previous\_nearest\_blocks["diamond"][0]:\\
\hbox{\ \ \ \ \ \ \ \ \ \ \ \ \ \ \ \ }reward -= 1\\
\hbox{\ \ \ \ \ \ \ \ }\# If the agent is close to lava, give a negative reward\\
\hbox{\ \ \ \ \ \ \ \ }if "lava" in current\_nearest\_blocks:\\
\hbox{\ \ \ \ \ \ \ \ \ \ \ \ }reward -= 1\\
\hbox{\ \ \ \ \ \ \ \ }\# If the agent's health decreases, give a negative reward\\
\hbox{\ \ \ \ \ \ \ \ }if "health" in GLOBAL\_DATA:\\
\hbox{\ \ \ \ \ \ \ \ \ \ \ \ }if health < GLOBAL\_DATA["health"]:\\
\hbox{\ \ \ \ \ \ \ \ \ \ \ \ \ \ \ \ }reward -= 1\\
\hbox{\ \ \ \ \ \ \ \ }GLOBAL\_DATA["health"] = health\\
\hbox{\ \ \ \ \ \ \ \ }return reward\\
\hbox{\ \ \ \ }def sparse(current\_nearest\_blocks, previous\_nearest\_blocks, inventory\_change, health, past\_agent\_positions, GLOBAL\_DATA):\\
\hbox{\ \ \ \ \ \ \ \ }\# Initialize reward\\
\hbox{\ \ \ \ \ \ \ \ }reward = 0\\
\hbox{\ \ \ \ \ \ \ \ }\# If the agent reaches the diamond, give a large positive reward\\
\hbox{\ \ \ \ \ \ \ \ }if "diamond" in current\_nearest\_blocks and current\_nearest\_blocks["diamond"][0] < 1:\\
\hbox{\ \ \ \ \ \ \ \ \ \ \ \ }reward += 100\\
\hbox{\ \ \ \ \ \ \ \ }\# If the agent dies, give a large negative reward\\
\hbox{\ \ \ \ \ \ \ \ }if health == 0:\\
\hbox{\ \ \ \ \ \ \ \ \ \ \ \ }reward -= 100\\
\hbox{\ \ \ \ \ \ \ \ }return reward\\
\hbox{\ \ \ \ }dense\_reward = dense(current\_nearest\_blocks, previous\_nearest\_blocks, inventory\_change, health, past\_agent\_positions, GLOBAL\_DATA)\\
\hbox{\ \ \ \ }sparse\_reward = sparse(current\_nearest\_blocks, previous\_nearest\_blocks, inventory\_change, health, past\_agent\_positions, GLOBAL\_DATA)\\
\hbox{\ \ \ \ }return np.sign(dense\_reward) * 1 + np.sign(sparse\_reward) * 0.1\\
--------------------------------------------------------------------------------------------------------------------------------------------

\texttt{\textbf{CRITIC OUTPUT:}}

\{
\\
\hbox{\ \ \ \ }"reasoning": "The reward function meets the requirements. It considers sparse reward of completing the objective and death, and dense reward of getting closer to diamond, having lava nearby and health decrease.",\\
\hbox{\ \ \ \ }"success": true,\\
\hbox{\ \ \ \ }"critique": null\\
\}

\end{tcolorbox}

\clearpage
\section{Example of GPT-4V as a Captioner}\label{sec:sup-gpt4v}
We show an example of using GPT-4V as a captioner to assist LLM in trajectory analysis of Figure \ref{fig:gpt4v_input}.

\begin{figure}[h]
    \centering
    \includegraphics[width=0.98\textwidth]{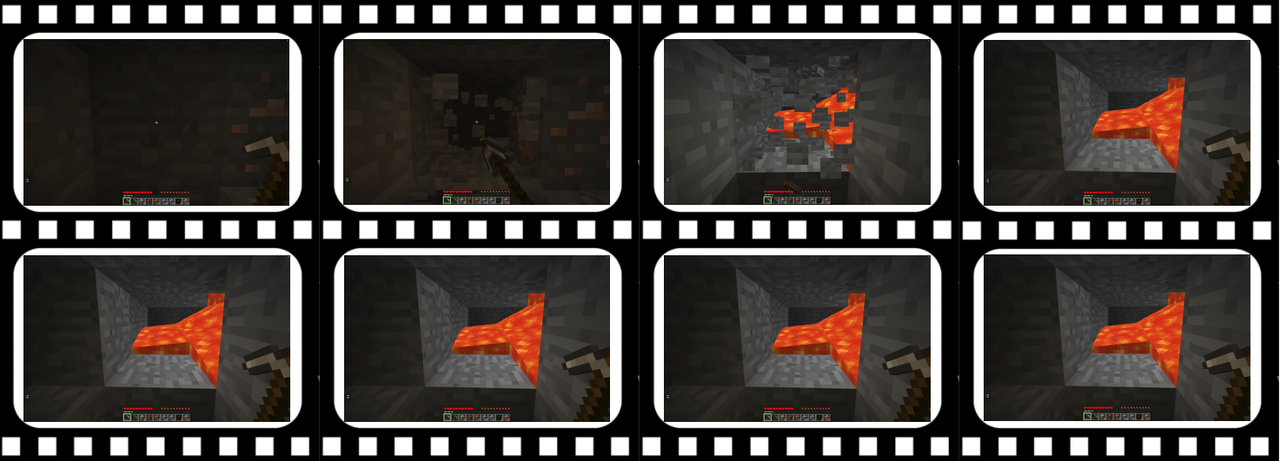}
    \caption{The last few frames of trajectory images of failed diamond mining.}
    \label{fig:gpt4v_input}
\end{figure}

\begin{tcolorbox}[breakable=true, boxrule={0.5pt}, sharp corners={all}]
\setlength{\parskip}{1ex}
\small
\texttt{\textbf{GPT-4V INPUT:}}

This is a Minecraft game video clip. Please describe the scene and the player's actions in detail. Note that describe the entire sequence of images rather than describing each image individually.

--------------------------------------------------------------------------------------------------------------------------------------------

\texttt{\textbf{GPT-4V OUTPUT:}}

The sequence of images depicts a typical scenario in the game Minecraft where a player is mining underground. Initially, the player is in a dimly lit area, facing a wall of stone blocks, indicating they are inside a tunnel or a cave. The player's health bar shows a slight reduction in health, and the inventory is equipped with essential items for mining, such as pickaxes and cobblestone blocks.

As the player begins to mine through the stone with the pickaxe, they make progress with each swing, as indicated by the developing cracks in the stone block. This action is part of the core gameplay in Minecraft, where players gather resources by mining various types of blocks.

Suddenly, the player breaks through to an unexpected hazard — \hl{a flow of bright orange lava appears behind the mined block. Lava is dangerous in Minecraft}; it can cause damage to the player and destroy items. The player quickly steps back to avoid getting burned.

In the next moments, the player cautiously assesses the situation by looking at the lava from different angles, probably searching for a safe way to navigate around or contain it. The subsequent images indicate that the player is trying to deal with the lava flow, which is a common and risky part of mining deep underground in Minecraft.

Throughout the sequence, \hl{the player remains close to the lava}, suggesting they may be attempting to block it with the cobblestone from their inventory or create a barrier to prevent the lava from spreading. This series of images captures the unpredictability and the survival aspects of Minecraft gameplay, where players must quickly adapt to the challenges they encounter while exploring and gathering resources.

--------------------------------------------------------------------------------------------------------------------------------------------

\texttt{\textbf{ANALYZER INPUT:}}

You are now a proficient Minecraft player. I have trained an RL agent and tested for several times in the Minecraft game environment.
I will give you a description of the player's scene and behavior at the end of the failure in text form.

\#\# Task description

- Objective: Find and approach diamond with higher success rate and avoid death.

- Initial Status: 
\\
\hbox{\ \ }1. The agent is under ground at y level 11, the most common level to find diamonds.\\
\hbox{\ \ }2. The agent already has an iron pickaxe.\\
- Success criteria: The distance to the nearest diamond block is less than 2 meter. 

- Procedure: Explore horizontally to find a diamond, face it and approach it.

\#\# General game information

- The version of Minecraft is 1.11.

- Common block names : dirt, cobblestone, iron, diamond, wood, coal, water, air, lava, leaves, ... Collected item names are the same.

- FOV is set to -35 to 35 degrees for yaw and -30 to 30 degrees for pitch. The max visible distance of blocks is 64.

- Length of each block is 1 meter.

- The y coordinate of the agent is the agent's height. The larger the y, the higher the agent. The ground level is around y = 63, but is not fixed.

- At each step, the agent selects one operation in is action space. The action space includes doing nothing (i.e. staying still), moving forward and backward, jumping and attacking. The action space also includes turning left, right, up and down. 

- The attack range is 2 meters in front of the agent. The agent can move 0.2 meters in each step. The agent can turn 15 degrees in each step. Typically, the agent needs to stay still and attack for **60 successive steps** to break a block.

- The hunger value is always at the max level.

\#\# Description of  the player's scene and behavior at the end of the failure

- The sequence of images depicts a typical scenario in the game Minecraft where a player is mining underground. Initially, the player is in a dimly lit area, facing a wall of stone blocks, indicating they are inside a tunnel or a cave. The player's health The bar shows a slight reduction in health, and the inventory is equipped with essential items for mining, such as pickaxes and cobblestone blocks.

- As the player begins to mine through the stone with the pickaxe, they make progress with each swing, as indicated by the developing cracks in the stone block. This action is part of the core gameplay in Minecraft, where players gather resources by mining various types of blocks.

- Suddenly, the player breaks through to an unexpected hazard — a flow of bright orange lava appears behind the mined block. Lava is dangerous in Minecraft; it can cause damage to the player and destroy items. The player quickly steps back to avoid getting burned.
- In the next moments, the player cautiously assesses the situation by looking at the lava from different angles, probably searching for a safe way to navigate around or contain it. The subsequent images indicate that the player is trying to deal with the lava flow, which is a common and risky part of mining deep underground in Minecraft.

- Throughout the sequence, the player remains close to the lava, suggesting they may be attempting to block it with the cobblestone from their inventory or create a barrier to prevent the lava from spreading. This series of images captures the unpredictability and the survival aspects of Minecraft gameplay, where players must quickly adapt to the challenges they encounter while exploring and gathering resources.

Suppose the player dies after going through the above scenarios and actions. You should help me write the most likely reason of failure, and the most feasible suggestion on the game strategy based on the above reason.

--------------------------------------------------------------------------------------------------------------------------------------------

\texttt{\textbf{ANALYZER OUTPUT:}}

Based on the description of the player's scene and behavior at the end of the failure, \hl{the most likely reason for failure appears to be the encounter with lava while mining}. Lava is a significant hazard in Minecraft, especially when mining at deeper levels where diamonds are commonly found. The sudden appearance of lava behind a mined block can catch players off guard, leading to damage from burns or even death if they are unable to escape or protect themselves in time. The player's attempt to manage the lava flow suggests awareness of the danger but may have been insufficiently cautious or effective in containing or avoiding the lava.

\#\#\# Most Likely Reason of Failure:

\hl{The failure was likely due to the player's inability to safely navigate the sudden lava flow encountered during mining}. Despite attempts to assess and manage the situation, the proximity to the lava and possibly inadequate measures to block or divert it might have resulted in sustained damage leading to death.

\#\#\# Suggestion on Game Strategy:

1. **Advance Preparedness**: Always carry water buckets in the inventory. A water bucket can be crucial for quickly converting lava into obsidian or stone, neutralizing the immediate threat. This quick response can save the player from burning and provides a safer path to continue mining.

2. **Strategic Mining Approach**: When mining at levels where lava is common, adopt a cautious approach:

   - **Mine at a Safe Distance**: Instead of directly mining the block in front, mine while standing back or to the side to avoid sudden lava flows.
   
   - **Use Blocks to Block Lava**: Have blocks ready in the hotbar to quickly place in front of a lava source. Cobblestone or dirt blocks can be used to block the flow or create barriers.
   
   - **Listen for Lava Sounds**: Pay attention to audio cues. The sound of bubbling lava can alert the player to its presence behind or adjacent to blocks being mined.

3. **Environmental Awareness**: Before breaking a block, check the surroundings for signs of lava, such as light emanating from cracks or the sound of flowing lava. This can preemptively alert the player to potential dangers.

4. **Escape Plan**: Always have an escape route planned when mining deep underground. This means knowing the way back to safer areas or having a clear path to retreat if faced with hazards like lava.

5. **Practice Lava Management**: Familiarize with techniques to manage lava flows, such as placing blocks to divert the flow away from the mining area or using water to solidify lava. Practicing these techniques in safer conditions can prepare the player for real encounters.

By following these strategies, the player can significantly reduce the risk of death from lava while mining for diamonds and improve their overall success rate in finding valuable resources.   
\end{tcolorbox}

\end{onecolumn}

\end{document}